\documentclass[11pt, a4paper, logo, copyright]{googlecloud}

\makeatletter
\renewenvironment{abstract}{%
    \setlength{\parindent}{0pt}%
    \noindent\bfseries
}{%
    \par\vspace{1.5em}
}
\makeatother

\pdftrailerid{redacted}

\usepackage[utf8]{inputenc}
\usepackage[T1]{fontenc}
\usepackage{microtype}
\usepackage{nicefrac}
\usepackage{textcomp}
\newcommand{\supertiny}{\tiny}

\usepackage[numbers, sort&compress, super]{natbib} 

\usepackage{authblk}
\renewcommand\Affilfont{\itshape\small}

\usepackage{amsmath,amssymb,amsfonts,amsthm}
\usepackage{mathrsfs}
\usepackage{bm}
\usepackage{bbm}

\usepackage{graphicx}
\usepackage{caption}
\usepackage{subcaption}
\usepackage{wrapfig}
\usepackage{pdflscape}
\usepackage{rotating}
\usepackage{float}
\usepackage{multirow}
\usepackage{multicol}
\usepackage{array}
\usepackage{booktabs}
\usepackage{threeparttable}
\usepackage{longtable}
\usepackage{makecell}

\usepackage{algorithm}
\usepackage{algorithmicx}
\usepackage{algpseudocode}
\usepackage{listings}
\usepackage{fancyvrb}
\usepackage{inconsolata}

\usepackage{xcolor}
\usepackage{color}
\usepackage[skins,breakable,most]{tcolorbox}
\usepackage{mdframed}
\usepackage{soul}
\usepackage{tikz, adjustbox}

\usepackage{kantlipsum, lipsum}
\usepackage{dsfont}
\usepackage{enumitem}
\usepackage{flushend}
\usepackage{balance}
\usepackage{url}
\usepackage{pifont}
\usepackage{comment} 
\usepackage{fontawesome}
\usepackage{csquotes}
\usepackage{xspace}
\usepackage{manyfoot}
\usepackage[title]{appendix}
\usepackage[colorinlistoftodos]{todonotes}
\usepackage{placeins}

\definecolor{beigecolor}{RGB}{253, 244, 204}
\definecolor{greencolor}{RGB}{228, 242, 217}
\definecolor{bluecolor}{RGB}{66, 133, 244}
\definecolor{orgcolor}{RGB}{255, 140, 15}
\definecolor{redcolor}{RGB}{234, 67, 53}
\definecolor{ggreen}{RGB}{52, 168, 83}
\definecolor{gyellow}{RGB}{251, 188, 5}
\definecolor{lightblue}{rgb}{0.22,0.45,0.70}

\newcommand{\ourmethod}{{ERA}\xspace}  

\newcolumntype{L}[1]{>{\raggedright\let\newline\\\arraybackslash\hspace{0pt}}m{#1}}
\newcolumntype{C}[1]{>{\centering\let\newline  \\\arraybackslash\hspace{0pt}}m{#1}}
\newcolumntype{R}[1]{>{\raggedleft\let\newline \\\arraybackslash\hspace{0pt}}m{#1}}
\newcolumntype{Y}[1]{>{\centering\arraybackslash}p{#1}}

\definecolor{backcolour}{rgb}{0.95,0.95,0.92}
\definecolor{codegreen}{rgb}{0,0.6,0}
\definecolor{codegray}{rgb}{0.5,0.5,0.5}
\definecolor{codepurple}{rgb}{0.58,0,0.82}

\lstdefinestyle{mystyle}{
    backgroundcolor=\color{backcolour},   
    commentstyle=\color{codegreen},
    keywordstyle=\color{magenta},
    numberstyle=\tiny\color{codegray},
    stringstyle=\color{codepurple},
    basicstyle=\ttfamily\scriptsize,
    breakatwhitespace=false,         
    breaklines=true,                 
    captionpos=b,                    
    keepspaces=true,                 
    numbers=left,                    
    numbersep=5pt,                  
    showspaces=false,                
    showstringspaces=false,
    showtabs=false,                  
    tabsize=2,
    frame=none,
    aboveskip=1pt,
    belowskip=1pt,
}
\lstdefinestyle{plainins}{
    backgroundcolor=\color{white},   
    commentstyle=\color{codegreen},
    keywordstyle=\color{magenta},
    numberstyle=\tiny\color{codegray},
    stringstyle=\color{codepurple},
    basicstyle=\ttfamily\scriptsize,
    breakatwhitespace=false,         
    breaklines=true,                 
    captionpos=b,                    
    keepspaces=true,                 
    numbers=none,                    
    numbersep=5pt,                  
    showspaces=false,                
    showstringspaces=false,
    showtabs=false,                  
    tabsize=2,
    aboveskip=0pt,
    belowskip=0pt,
    frame=single
}
\lstdefinestyle{plainexam}{
    backgroundcolor=\color[HTML]{FFFCF3},   
    commentstyle=\color{codegreen},
    keywordstyle=\color{magenta},
    numberstyle=\tiny\color{codegray},
    stringstyle=\color{codepurple},
    basicstyle=\ttfamily\scriptsize,
    breakatwhitespace=false,         
    breaklines=true,                 
    captionpos=b,                    
    keepspaces=true,                 
    numbers=none,                    
    numbersep=5pt,                  
    showspaces=false,                
    showstringspaces=false,
    showtabs=false,                  
    tabsize=2,
    aboveskip=0pt,
    belowskip=0pt
}
\lstdefinestyle{pythonstyle}{
    language=Python,
    backgroundcolor=\color{black!5},
    commentstyle=\color{green!40!black},
    keywordstyle=\color{blue},
    stringstyle=\color{purple},
    numberstyle=\tiny\color{gray},
    basicstyle=\ttfamily\footnotesize,
    breaklines=true,
    captionpos=b,
    numbers=left,
    showstringspaces=false,
}
\lstdefinestyle{mypythonstyle}{
    language=Python,
    basicstyle=\ttfamily\fontsize{5pt}{5.5pt}\selectfont,
    keywordstyle=\color{blue},
    stringstyle=\color{purple},
    commentstyle=\color{green!40!black},
    breaklines=true,
    showstringspaces=false,
    numbers=left,
    numberstyle=\fontsize{3pt}{4pt}\selectfont\color{gray},
    numbersep=5pt,
    frame=tb, 
    framerule=0.5pt,
}

\tcbset{
  aibox/.style={
    width=\linewidth,
    top=5pt,
    bottom=1pt,
    colback=blue!6!white,
    colframe=black,
    colbacktitle=black,
    enhanced,
    center,
    fontupper=\scriptsize,
    attach boxed title to top left={yshift=-0.1in,xshift=0.15in},
    boxed title style={boxrule=0pt,colframe=white,},
  }
}
\newtcolorbox{AIbox}[2][]{aibox,title=#2,#1}

\renewcommand{\figurename}{Fig.}
\newcommand{\figref}[1]{Fig.~\ref{#1}}
\newcommand{\supfig}[1]{Supplementary Fig.~\ref{#1}}
\newcommand{\supfigs}[1]{Supplementary Figs.~\ref{#1}}
\newcommand{\extfig}[1]{Extended Data Fig.~\ref{#1}}
\newcommand{\extdat}[1]{Extended Data Table~\ref{#1}}
\newcommand{\tabref}[1]{Table~\ref{#1}}
\newcommand{\suptab}[1]{Supplementary Table~\ref{#1}}
\newcommand{\suptabs}[1]{Supplementary Tables~\ref{#1}}

\DeclareCaptionLabelSeparator{pipe}{ $|$ }
\DeclareCaptionLabelFormat{boldsimple}{\textbf{#1 #2}}
\usepackage{hyperref}
\hypersetup{
  colorlinks,
  citecolor=blue,
  linkcolor=red,
  urlcolor=blue
}
\usepackage[capitalize,noabbrev]{cleveref}

\newcommand{\urlprefix}{}

\newcommand{\copyrightext}{\footerfont \textsuperscript{*}Equal contribution in alphabetical order. \\ 
\textsuperscript{**} Carried out as part of a student researchership at Google Research. \\
$\ddagger$ To whom correspondence should be addressed: shibl@google.com, mbrenner@google.com}



\title{An AI system to help scientists write expert-level empirical software 
}
\author[1,*]{Eser Ayg\"un}
\author[2,*]{Anastasiya Belyaeva}
\author[1,*]{Gheorghe Comanici}
\author[2,*]{Marc Coram}
\author[2,*]{Hao Cui}
\author[3,*]{Jake Garrison}
\author[2,*]{Renee Johnston}
\author[2,*]{Anton Kast}
\author[2,*]{Cory Y. McLean}
\author[2,*]{Peter Norgaard}
\author[2,*]{Zahra Shamsi}
\author[1,*]{David Smalling}
\author[2,*]{James Thompson}
\author[2,*]{Subhashini Venugopalan}
\author[2,*]{Brian P. Williams}
\author[2,4,**]{Chujun He}
\author[2,5,**]{Sarah Martinson}
\author[2,6,**]{Martyna Plomecka}
\author[2]{Lai Wei}
\author[2]{Yuchen Zhou}
\author[2,5,**]{Qian-Ze Zhu}
\author[2]{Matthew Abraham}
\author[2]{Erica Brand}
\author[1]{Anna Bulanova}
\author[2,7]{Jeffrey A. Cardille}
\author[2]{Chris Co}
\author[2]{Scott Ellsworth}
\author[2]{Grace Joseph}
\author[2]{Malcolm Kane}
\author[2,5,**]{Ryan Krueger}
\author[2]{Johan Kartiwa}
\author[2]{Dan Liebling}
\author[2]{Jan-Matthis Lueckmann}
\author[2]{Paul Raccuglia}
\author[2,8,**]{Xuefei (Julie) Wang}
\author[2]{Katherine Chou}
\author[2]{James Manyika}
\author[2]{Yossi Matias}
\author[2]{John C. Platt}
\author[2]{Lizzie Dorfman}
\author[1,$\ddagger$]{Shibl Mourad}
\author[2,5,$\ddagger$]{Michael P. Brenner}

\affil[1]{Google DeepMind, Montréal, Quebec H2Z 1W5, Canada}
\affil[2]{Google Research, Cambridge, MA 02142, USA}
\affil[3]{Google Platforms and Devices, Mountain View, CA 94043, USA}
\affil[4]{Massachusetts Institute of Technology, Cambridge, MA 02139, USA}
\affil[5]{School of Engineering and Applied Sciences, Harvard University, Cambridge, MA 02138, USA}
\affil[6]{Google DeepMind, New York, New York 10011, USA}
\affil[7]{Faculty of Agricultural and Environmental Sciences, McGill University, Montréal, Quebec H3A 0G4, Canada}
\affil[8]{California Institute of Technology, Pasadena, CA 91125}

\newcommand{\theabstract}{} 

\date{}
\begin{document}
\maketitle

\noindent\footnotesize \textsuperscript{*}Equal contribution in alphabetical order. \\ 
\textsuperscript{**} Carried out as part of a student researchership at Google Research. \\
\textsuperscript{$\ddagger$} To whom correspondence should be addressed: shibl@google.com, mbrenner@google.com

\normalsize
\vspace{3.5em}
\begin{abstract}
The cycle of scientific discovery is frequently bottlenecked by the slow, manual creation of software to support computational experiments\cite{hannay2009how}. To address this, we present Empirical Research Assistance (ERA), an AI system that creates expert-level scientific software whose goal is to maximize a quality metric. The system uses a Large Language Model (LLM) and Tree Search (TS)\cite{silver2016mastering} to systematically improve the quality metric and intelligently navigate the large space of possible solutions. ERA achieves expert-level results when it explores and integrates complex research ideas from external sources. The effectiveness of tree search is demonstrated across a diverse range of tasks. In bioinformatics, ERA discovered 40 novel methods for single-cell data analysis that outperformed the top human-developed methods on a public leaderboard. In epidemiology, ERA generated 14 models that outperformed the CDC ensemble and all other individual models for forecasting COVID-19 hospitalizations. ERA also produced expert-level software for geospatial analysis, neural activity prediction in zebrafish, and numerical solution of integrals, and a novel rule-based construction for time series forecasting. By devising and implementing novel solutions to diverse tasks, ERA represents a significant step towards accelerating scientific progress.
\end{abstract}

\vspace{1em}
\vspace{1em}

\section*{Introduction}\label{sec1}
Empirical software, designed to maximize a measurable quality score, is ubiquitous and central to many scientific endeavors. Empirical software has recently enabled a number of Nobel Prizes in Chemistry: in 1998 for Density Functional Theory~\cite{hohenberg1964inhomogeneous,kohn1965self}, in 2013 for molecular dynamics simulation~\cite{warshel1976theoretical} and in 2024 for protein structure prediction~\cite{jumper2021highly,baek2021accurate}. Empirical software underlies our ability to create models of complex systems, ranging from parameterizations of a vertical column of the earth's atmosphere for weather modeling \cite{hourdin2017art}, to the parameterization of stress response in a turbulent fluid flow \cite{anderson2009basic}, to the prediction of social systems \cite{silver2012signal,farmer2024making}.

However, {\it empirical software for science is slow and difficult to create}. Domain-specific empirical software requires tedious work\cite{hannay2009how}, often over many years. When empirical software is used to test complex hypotheses, it becomes ever more difficult to write purely from first principles. There usually is no systematic search for alternative approaches. Design choices are often governed by intuition or expediency, rather than exhaustive experimentation. Creating the software is so time-consuming that it severely limits the possibilities that can be productively
explored\cite{sculley2015hidden}.

\begin{figure}[htbp]
    \centering
    \includegraphics[width=\textwidth]{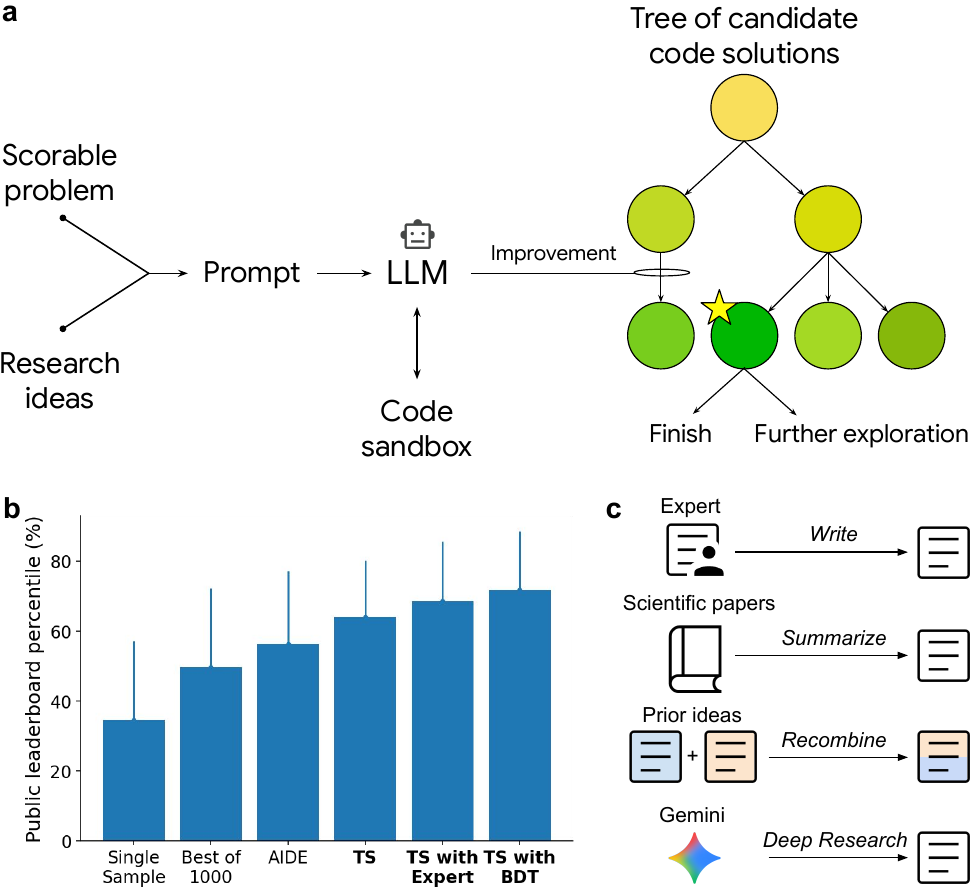}
\caption{\textbf{Schematic and performance of ERA}. \textbf{a}, Schematic of ERA algorithm. A scorable task, together with research ideas proposing methods to solve the task, are fed to an LLM, which produces code to evaluate the scorable task in a sandbox. This is then embedded within a tree search algorithm, whereby new nodes are chosen balancing exploitation and exploration, sampling from the LLM (Methods).
\textbf{b}, Performance of code generation methods on Kaggle Playground benchmark. Results report the average public leaderboard percentile performance over 16 tasks. Methods based on ERA are listed in bold. TS, ERA-based tree search. BDT, boosted decision tree. Error bars indicate standard deviation of performance between different competitions in the benchmark.
\textbf{c}, The system automates empirical software development  by iteratively optimizing predictive models and computational algorithms based on research ideas and a defined quality metric. We use LLM summaries of scientific papers or AI assisted literature as part of the prompt, and also recombine successful implementations of ideas to create more powerful methods.}
    \label{fig:schematic}
\end{figure}

Here we present Empirical Research Assistance (ERA), an AI-based system that systematically and automatically creates empirical software to solve scorable tasks. ERA is based on an LLM that rewrites software to attempt to improve its quality score. ERA creates multiple software candidate solutions, and uses Tree Search~\cite{silver2016mastering} to decide which candidates merit further exploration (\figref{fig:schematic}a).
While there are many ways of designing a code mutation system \cite{jiang2025aide,novikov2025alphaevolve,romera2024mathematical,hu2024automated}, we developed ERA by competing in Kaggle competitions (\figref{fig:schematic}b), described below. 
We augment code mutation with research ideas obtained from a range of sources including highly-cited papers, specialized textbooks, and search engine results (\figref{fig:schematic}c). In practice, these ideas can be injected either directly by the user or automatically using a search engine to access research in the literature. The LLM uses this injected guidance in writing code.

We find that ERA produces software that outperforms the state-of-the-art in scorable tasks spanning broad scientific disciplines.  This expert-level performance arises because of the ability
to exhaustively and tirelessly carry out solution searches at unprecedented scale, identifying
needle-in-the-haystack high quality solutions.

\section*{Results}\label{sec2}

\subsection*{Overview of Scorable Tasks}
We develop ERA on Kaggle playground competitions, and test it by selecting scorable tasks based on scientific or engineering problems of high relevance in diverse fields. These scorable tasks are listed below, with per-node computational costs outlined in \suptab{tab:compute_cost}.

\noindent
{\it scRNA-seq batch integration}:~\cite{xu2023inthcad} This task requires distinguishing subtle biological signals from noise in high-dimensional sparse datasets. By removing confounding factors, we can enable large-scale multi-lab transcriptomic data integration. 

\noindent
{\it CDC COVID Forecasting}:~\cite{CovidForeCastHub} This task requires predicting non-linear disease dynamics from lagged and noisy real-time data. By predicting COVID cases several weeks in advance, we can inform public health policy and resource allocations. 

\noindent
{\it Time series forecasting}:~\cite{aksu2024gift} This task requires predicting time series outcomes across a range of datasets and applications.

\noindent
{\it Geospatial segmentation}:~\cite{shao2018performance} This task requires performing dense pixel-wise multi-label semantic segmentation on complex satellite imagery. Better segmentation can lead to large improvements in environmental monitoring and disaster response.

\noindent
{\it ZAPBench}:~\cite{lueckmann2025zapbench} This task requires predicting the activity of >70,000 neurons across an entire vertebrate brain. Performing well on this benchmark may lead to a systems-level understanding of brain function and behavior.

\noindent
{\it Numerically solving difficult integrals}: This task requires solving integrals that defy standard numerical algorithms. It is useful for modeling physical and engineering systems.

\subsection*{Kaggle Playground Benchmark}\label{kaggleResults}
We designed ERA to score highly on a curated set of Kaggle competitions. Kaggle calibrates human performance 
with percentile rank on a leaderboard, and we score code by submitting directly to Kaggle.
Our benchmark consists of 16 playground competitions from the 2023 season, encompassing regression and classification tasks (\suptab{tab:kaggle_competitions}). 
Playground competitions are an ideal benchmark because they 
offer fast iteration, simplicity, and calibration against thousands of humans. Achieving a high score requires creating complex code without requiring solving a sophisticated scientific task.

Our basic strategy uses a simple prompting template (\suptab{table:playground_normal}) that concatenates the competition description with the previous trial.
\figref{fig:schematic}b evaluates the performance of ERA with the average public percentile rank across all 16 playground competitions: 
ERA substantially beats a single LLM call and best-of-$1000$ LLM calls, and also outperforms AIDE\cite{jiang2025aide}, owing to its ability to maintain a diverse tree of candidates, allowing it to backtrack when a specific line of code mutation plateaus.
During the search, the system discovers strategies leading to abrupt jumps in the score, with the 
accumulation of these jumps leading to the highest quality solutions.

Problem-specific advice added to the prompt substantially improves performance. We illustrate this with two examples. In {\sl TS with expert advice} we give ERA standard advice to win Kaggle competitions (\suptab{table:playground_expert}). In {\sl TS with Boosted Decision Tree (BDT)} we tell ERA to implement a boosted decision tree library from scratch, without using standard packages (\suptab{table:playground_bdt}). We manually verified in both cases that resulting codes followed the advice. 
 
We now evaluate ERA on six benchmarks in different scientific fields, exploring distinct ways to incorporate research ideas to improve system performance (\figref{fig:schematic}c, Methods).

\subsection*{Genomics: Batch Integration of Single-Cell RNA Sequencing Data}

Single-cell RNA sequencing (scRNA-seq) has revolutionized our ability to dissect cellular heterogeneity, discover novel cell types, infer gene regulatory networks and developmental trajectories, and improve therapeutic target prioritization~\cite{jovic2022single}, enabling hundreds of millions of cells to be individually sequenced within thousands of datasets~\cite{svensson2018exponential,regev2017human,czi2025cz}.
A major challenge required to jointly analyze many disparate scRNA-seq datasets is to computationally remove complex batch effects present across samples while preserving biological signals~\cite{stuart2019integrative}.
Nearly 300 tools exist to perform batch integration of scRNA-seq data~\cite{zappia2018exploring}, and multiple
benchmarks have been developed for assessing  metrics of batch effect removal and conservation of biological variability~\cite{tran2020benchmark,chazarra2021flexible,luecken2025}. 

\begin{figure}[htp]
   \centering
    \includegraphics[width=\textwidth]{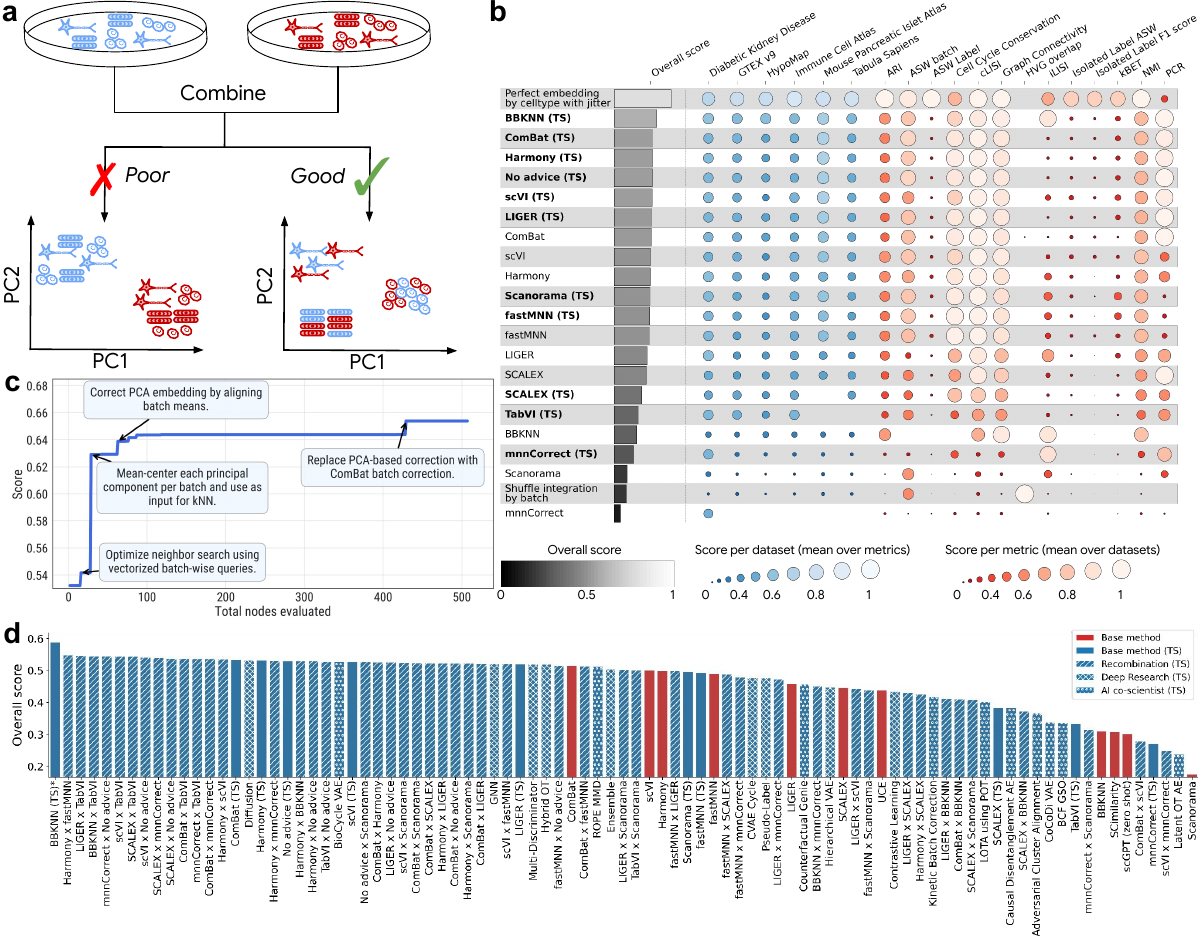}
    \caption{{\textbf{Performance of ERA on scRNA-seq batch integration.} \textbf{a}, Schematic of the batch integration task, in which disparate datasets (teal and red) are processed to remove batch effects in the data while retaining biological variability. \textbf{b}, Performance of tree search (method names bolded and suffixed by ``(TS)'') compared to the analogous published method on the OpenProblems benchmark v2.0.0~\cite{luecken2025}. ``Perfect embedding by celltype with jitter'' is a positive control method that represents the best possible performance and ``Shuffle integration by batch'' is a negative control that does not perform any batch integration. Overall score is the mean over all datasets and metrics. Each Datasets column shows the mean of all metrics computed over that dataset. Each Metrics column shows the mean of that metric computed over all datasets. Metrics were assigned a value of 0 if they could not be computed or if their performance was worse than the lowest negative control; these are displayed as empty. \textbf{c}, Performance improvements annotated with code innovation for the top-performing batch balanced $k$-nearest neighbors (BBKNN)  implementation. ComBat-based embedding generation was introduced in implementation attempt 429. \textbf{d}, Overall score for OpenProblems benchmark v2.0.0~\cite{luecken2025} non-control methods, ERA with and without recombination of ideas, Gemini Deep Research~\cite{GeminiDeepResearch}, and ERA with AI co-scientist~\cite{gottweis2025towards}. Y-axis lower bound is the overall score of the ``Shuffle integration by batch'' negative control method. Seven recombination, five base methods, and two AI co-scientist methods that do not match its performance are omitted. * indicates the method is a recombination, even if not explicitly prompted for recombination. TS, ERA-based tree search; fastMNN, batchelor fastMNN; mnnCorrect, batchelor mnnCorrect.}
    }
\label{fig:batch_integration}
\end{figure}

To assess ERA performance on 
this task,
we used the OpenProblems v2.0.0 batch integration benchmark~\cite{luecken2025}. As of July 2025, this active benchmark evaluates 15 state-of-the-art methods and eight control methods on 13 different metrics that quantify both the ability to remove batch effects in the data and retain variability attributable to true biological differences in six datasets spanning human and mouse~\cite{czi2025cz} (\figref{fig:batch_integration}a). To avoid overfitting to the benchmark, we used a separate dataset for ERA optimization (Methods,~\supfig{fig:dataflow_sc}). For each ERA run, we selected the best solution based on the performance on this training set, and report the performance on the holdout OpenProblems datasets (n=1,747,937 total cells).
We prompt the LLM with a description of the single cell batch integration problem, code for reading in the dataset, code for evaluation metrics, and optional text with a particular research idea.

First, we ran ERA without guidance, and observed
that its solution is conceptually similar to ComBat~\cite{johnson2007}, yet improved over the current OpenProblems leaderboard (\texttt{No advice (TS)} in \figref{fig:batch_integration}b).  We then evaluated whether ERA could improve upon existing algorithms. We selected nine methods from the OpenProblems benchmark, including the six highest-performing methods (Methods). For each method, we obtained the paper PDF and used Gemini 2.5 Pro to add a brief summary to the prompt (Methods).
In pairwise comparisons, ERA outperformed the corresponding published result for eight of the nine methods in overall score (\figref{fig:batch_integration}b). The top-performing method was an ERA implementation of Batch Balanced K-Nearest Neighbors (\texttt{BBKNN (TS)})~\cite{polanski2019}, yielding a 14\% overall improvement over the best published method (\texttt{ComBat}~\cite{johnson2007}) and equaled or outperformed the corresponding published \texttt{BBKNN} in every dataset and across 11/13 metrics (\figref{fig:batch_integration}b).
This performance highlights its capacity to effectively remove batch effects without compromising biological signals (\supfig{fig:bbknn_umap_immune_cell_atlas}).
We observed that ERA is also able to produce performant implementations for an algorithm without publicly-available code (TabVI~\cite{chandrashekar2025tabvi}, \extfig{fig:barplot_replicates}).
 Importantly, expert manual inspection of the code solutions proposed by ERA confirmed that nearly all implementations adhered to the requested algorithms (\extdat{table:sc_baseline_method_adherence}), with
 performance largely consistent across replicate runs of methods (\extfig{fig:barplot_replicates}). Additionally, ERA demonstrated improvements even when compared to base methods with optimized hyperparameters, indicating that its contribution extends beyond hyperparameter tuning (Methods, \supfig{fig:barplot_replicates_hparamopt}).
 \extfig{fig:single_cell_tree_breakthrough} shows the tree structure and evolution of the maximum score as a function of the number of nodes in the tree for the best performing model \texttt{BBKNN (TS)}. 

For \texttt{BBKNN (TS)}, part of the performance boost came from combining two existing methods, ComBat~\cite{johnson2007} and BBKNN, rather than simply implementing BBKNN (\figref{fig:batch_integration}c). In particular, while the original BBKNN method computes neighbors using PCA embeddings, \texttt{BBKNN (TS)} computes neighbors using ComBat-corrected PCA embeddings, removing global linear batch-associated variance. Both implementations then compute $k$-nearest neighbors across batches and construct a graph (with differences in exact implementation), thus removing local batch effects (\suptab{table:sc_bbknn_code}). Manual modification of \texttt{BBKNN (TS)} and the published BBKNN implementation confirmed that using Combat-corrected PCA embeddings is critical for improving both implementations
(\extfig{fig:bbknn_deepdive}), confirming the value in idea recombination. 

This motivated an exploration of systematic ways to generate more complex research ideas.
First, similar to how scientists often combine ideas to create a novel approach, we programmatically generated 55 ``recombinations'' of all pairs of the 11 methods described above (No advice, nine replications, and TabVI; hereafter: ``base methods'') based on summaries of the code for each method (Methods,~\suptab{table:sc_recombine_summary_prompt}). We ran ERA, prompted with each of these recombinations. For each base method and recombination group, we compared the average scores for the top nodes over the intersection of metrics that were successfully computed for all three methods. Strikingly, recombination implementations of tree search frequently outperformed their base counterparts, with 24 of the 55 recombination solutions (44\%) outperforming both of their base methods and 22 of the remaining 31 recombination solutions outperforming one of the two base methods (\extfig{fig:recomb_bar_intersection}).
Second, we used Gemini Deep Research~\cite{GeminiDeepResearch} and AI co-scientist~\cite{gottweis2025towards} to generate and implement 21 additional ideas (Methods). 
After ERA was applied to each, 6/11 base methods,
 29/55 recombination, 4/9 Deep Research, and 1/12 AI co-scientist methods (40 of 87) outperform all methods currently published on the OpenProblems leaderboard (\figref{fig:batch_integration}d).

To further understand the conceptual space explored by ERA, we obtained embeddings for each generated code using Gemini text embedding model and computed cosine similarities (\supfig{fig:sc_sim_heatmap}). Visualization of the embeddings revealed distinct clusters, generally representing deep-learning based methods and non-deep-learning methods, suggesting that ERA is able to generate a variety of solutions (\supfig{fig:sc_sim_umap}).

\subsection*{Public Health: Prediction of U.S. COVID-19 Hospitalizations}

The primary U.S. benchmark for COVID-19 forecasting is the COVID-19 Forecast Hub (\href{https://github.com/CDCgov/covid19-forecast-hub/tree/main}{CovidHub})~\cite{CovidForeCastHub}, a large, collaborative effort coordinated by the Centers for Disease Control and Prevention (CDC). CovidHub receives weekly forecasts from dozens of expert-led teams across academia, industry, and government, each using different methodologies. These weekly forecasts must cover new COVID-19 related hospitalizations across 52 U.S. states and territories for the current week and three subsequent weeks over 23 specified quantiles. Submissions are evaluated using the Weighted Interval Score (WIS), which rewards both accuracy and well-calibrated uncertainty. 

Top-performing individual models include classic autoregressive time-series approaches (e.g., UMASS-ar6\_pooled), gradient boosting machine learning models (e.g., UMASS-gbqr), and epidemiological models based on renewal equations and Bayesian estimation of the reproductive number (e.g., CEPH-Rtrend\_covid). CovidHub leverages this methodological diversity by integrating submissions into the CovidHub Ensemble, a robust aggregate forecast that has historically provided the gold standard for epidemiological prediction in the U.S.

\begin{figure}
    \centering
    \includegraphics[width=1.0\linewidth]{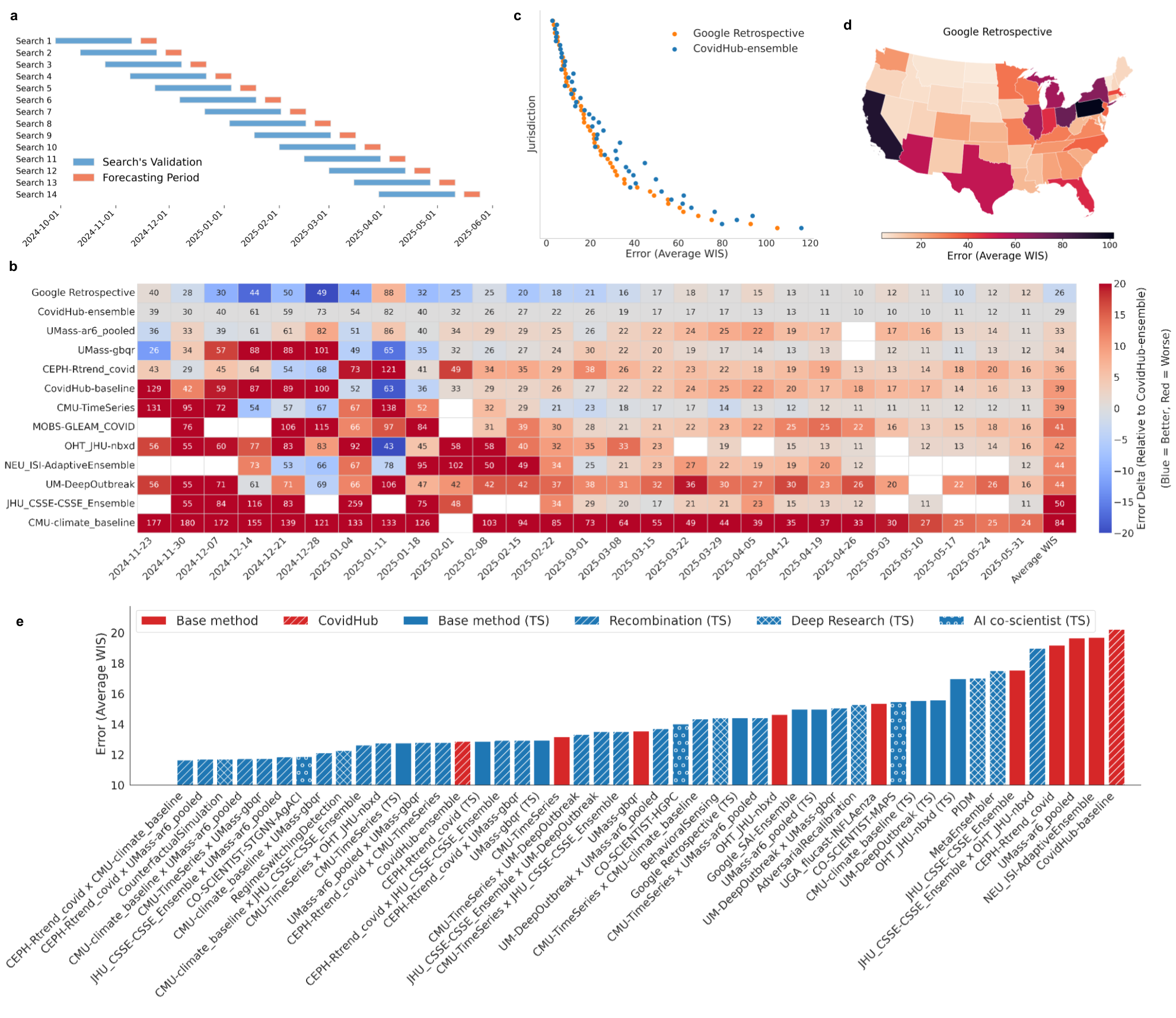}
    \caption{\textbf{Performance of ERA on COVID-19 forecasting.} 
    \textbf{a,} Rolling validation window used for the forecasting experiments. Each search's output is validated internally on a preceding block of time (blue), and the resulting model is then used to make predictions for its corresponding forecasting period (orange). Training data includes all dates on or after 2020-08-08 and prior to the validation set. 
    \textbf{b,} Time-series leaderboard showing weekly forecasting performance (Average WIS) for participating teams and our 'Google Retrospective' model, ordered by average WIS. Scores are aggregated across all 52 jurisdictions and four forecast horizons. The number within each cell is the model's absolute Average WIS for that week. The cell's background color visualizes the performance relative to the CovidHub-ensemble, with blue indicating a lower (better) WIS and red indicating a higher (worse) WIS. 
    \textbf{c,} Direct jurisdiction-level comparison of forecasting error (Average WIS) between our model and the 'CovidHub-ensemble', demonstrating our model's superior performance in a majority of locations.
    \textbf{d,} Geographic distribution of our model's forecasting error (Average WIS), aggregated over the entire 2024/25 COVID-19 season. Lower error values (lighter colors) indicate better performance.
    \textbf{e,} Comparison of aggregate forecasting performance for various modeling strategies. This includes baseline models from the CovidHub competition, our retrospective model, our replications of submitted models, novel hybrid models generated through recombination, deep research\cite{GeminiDeepResearch} and AI co-scientist\cite{gottweis2025towards}. 14 strategies (10 recombination; two Deep Research; one AI co-scientist and one replicated baseline) outperform the official CovidHub-ensemble for the 3-week (3 reference dates × 4 time horizons × 52 jurisdictions) evaluation period. Models that perform worse than CovidHub-baseline are not shown.}
    \label{fig:covid_figure}
\end{figure}

We designed a rigorous retrospective study to assess ERA performance in this competitive environment, using data available on May 1 2025. For every forecasting period, we ran ERA to optimize and select a model using data from the preceding six weeks, creating a rolling validation window throughout the 2024-2025 season (\figref{fig:covid_figure}a), with data splits elaborated in \suptab{tab:covid_splits}. The weekly performance of our resulting `Google Retrospective' model is detailed in the time-series leaderboard (\figref{fig:covid_figure}b), which visualizes our model's performance advantage relative to the CovidHub-ensemble and other top-performing teams. The temporal variation of WIS for each of the separate validation splits is shown for the best replicate (\extfig{fig:covid_retro_best_rep}) and across replicates (\supfig{fig:covid_retro_all_rep}).
A direct jurisdiction-level comparison confirms our model achieved a lower (better) WIS in a majority of states (\figref{fig:covid_figure}c,d). \supfigs{fig:supp_covid_curves} and \ref{fig:full_season_forecasts} compare prediction curves for representative locations between different models.

Overall, our model achieved the highest performance with an average WIS of 26, outperforming the official CovidHub Ensemble's average WIS of 29.
A representative tree and breakthrough plot is shown in \extfig{fig:breakthrough_tree_covid}.

Beyond this retrospective performance, we investigated ERA's ability to explore the solution space more broadly by replicating, recombining, and generating entirely new forecasting strategies (\figref{fig:covid_figure}e). First, we tested its ability to replicate existing methods from other teams using only their brief public descriptions from the CovidHub (\suptabs{table:covid-rep-methods} and \ref{table:covid-rep-prompt}). Our tree-search-based implementations (`Base Method (TS)') not only adhered to the provided instructions (\suptab{table:covid_method_adherence}) but also exceeded the performance of the original submissions in six of the eight cases tested; 
the two models that performed worse (replicating \verb|JHU_CSSE-CSSE_Ensemble| and \verb|OHT_JHU-nbxd|) did not use external data present for the original method implementations.
Next, we explored whether solutions could be improved through recombination. For this experiment, we prompted an LLM to analyze the core principles of two different parent models and then used its synthesis to instruct ERA to generate a novel hybrid strategy combining their respective strengths. Of 26 generated hybrid models (`Recombination (TS)'), 11 achieved a WIS score superior to both of their parent models (\extfig{fig:covid_recombined}). We manually verified that output code for all recombination experiments contained relevant aspects from both parent codes (\suptab{table:covid_method_adherence}). Finally, we used Gemini Deep Research\cite{GeminiDeepResearch} and AI co-scientist \cite{gottweis2025towards} to generate novel forecasting ideas which were then implemented using ERA. In total, this systematic exploration yielded 14 distinct strategies that outperformed the official CovidHub-ensemble: 10 from recombination, two from Deep Research, one from AI co-scientist, and one of our replicated baselines. Cosine similarities between embeddings for each generated code show clustering between different methods (\supfig{fig:covid_sim_heatmap}). An extended performance plot including the 9 TS models and 3 other submissions that did not outperform the baseline, alongside their corresponding validation performance, is provided in \supfigs{fig:covid_extended} and \ref{fig:covid_validation}. Examples of prediction curves are shown in \supfig{fig:prediction_covid}.

A deeper analysis of these 14 top-performing strategies reveals key patterns in how ERA achieves superior performance. The recombination models, which constitute the majority of the winners, highlight a clear pattern of synergistic hybridization. Two base models appear most frequently in these successful hybrids: the simple, climatology-based \verb|CMU-climate_baseline| and the statistical autoregressive model \verb|UMass-ar6_pooled|. This suggests ERA consistently discovers that the most effective strategies are built upon a robust foundation of historical averages and recent trends, which are then enhanced by more complex methods. Indeed, the most successful recombinations consistently fused different modeling paradigms: for instance, pairing the epidemiological \verb|CEPH-Rtrend_covid| model with the statistical \verb|UMass-ar6_pooled| model created a hybrid anchored in the theory of disease spread yet highly responsive to recent data trends, while pairing the powerful machine learning \verb|UMass-gbqr| model with the stable \verb|CMU-climate_baseline| provided a robust seasonal foundation that allowed the ML model to safely focus on learning short-term deviations.
Finally, we verified the importance of using ERA, rather than just taking the best of 1000 solution attempts, for both epidemiological modeling and scRNA-seq batch integration (\tabref{tab:era-comparison}).

\begin{table}[htbp]
\centering
\caption{\textbf{Performance comparison of ERA and Best-of-N on the scRNA-seq batch integration and epidemiological modeling tasks}. Performance is measured on the validation set for each task. Bold entries indicate better performance. BoN, best-of-N=1000.}
\label{tab:era-comparison}
\small 
\begin{tabular}{llcc}
\toprule
\textbf{Model} & \textbf{Method} & \textbf{Batch integration} & \textbf{Epidemiology} \\ 
 & & (higher better) & (lower better) \\ \midrule
\multirow{2}{*}{\textbf{Gemini 2.5 Flash}} & BoN & 0.6306 & 106.55 \\
 & ERA & \textbf{0.6552} & \textbf{93.07} \\ \midrule
\multirow{2}{*}{\textbf{Mistral Medium}} & BoN & 0.6129 & 95.73 \\
 & ERA & \textbf{0.6332} & \textbf{87.98} \\ \midrule
\multirow{2}{*}{\textbf{Claude Sonnet 4.6}} & BoN & 0.6502 & 85.03 \\
 & ERA & \textbf{0.6575} & \textbf{84.56} \\ \midrule
\multirow{2}{*}{\textbf{GPT-5}} & BoN & \textbf{0.6740} & 78.04 \\
 & ERA & 0.6671 & \textbf{74.55} \\ \midrule
\multirow{2}{*}{\textbf{Gemini 3.1 Pro}} & BoN & 0.6461 & 92.39 \\
 & ERA & \textbf{0.6641} & \textbf{72.70} \\ \bottomrule
\end{tabular}
\end{table}

\subsection*{Time Series Forecasting: GIFT-Eval}
We next evaluated ERA using General Time Series Forecasting Model Evaluation (GIFT-Eval)~\cite{aksu2024gift}, a standard benchmark for time series forecasting. GIFT-Eval includes 28 datasets across seven diverse domains, with data frequencies ranging from seconds to years. It receives $\sim$4 new submissions per month that span diverse methodologies including black-box deep learning methods and foundation models. Submissions are scored on official train/validation/test splits
using a normalized 
Mean Absolute Scaled Error (MASE) metric, 
calculated relative to a seasonal naive baseline. 

We applied ERA in two phases. We began with a \texttt{per-dataset solution} in which ERA discovers an independent solution for each dataset. 
The second \texttt{unified solution} created a single general-purpose forecasting model 
using only basic libraries by hill-climbing against the average score for the entire GIFT-Eval.

\paragraph*{Per-dataset solution}
Here we allowed ERA to use a full suite of Python libraries, including \texttt{scikit-learn}, \texttt{statsmodels}, and \texttt{xgboost}.
ERA outperformed the entire May 18, 2025 leaderboard, which included foundation models, deep learning models, and standard time series methods (\suptab{tab:leaderboard_snapshot_mase_full}).
The discovered solutions showed strong convergence towards \texttt{gradient boosting} and \texttt{ensemble/decomposition} models (\supfig{fig:gift-category}). 

\paragraph*{Unified solution}
We wondered whether  the code mutation system could create a unified, general-purpose forecasting library from scratch, by hill climbing with a {\sl single} code on the average MASE on the entire GIFT-Eval dataset. 
To manage the benchmark's diversity, we allowed the library to have an adaptive configuration system, whereby it could generate up to 8 preset hyperparameter configurations to adapt to the diversity of datasets, with a validation step selecting the best performing configuration for each dataset.
As the search progressed, 
 date and trend-related features often led to performance breakthroughs leading to a model that sequentially forecasts and subtracts individual time series components, including a base level, trend, seasonality, datetime-based features, and a final residual correction (\extfig{fig:tree_breakthrough_gift}).
 The configurations (\suptab{tab:unified_config_list}) include date-specific features, including one that featurizes holidays in a specific set of countries ([`US', `DE', `CN', `GB', `CA', `AU']) . 
The resulting \texttt{unified solution} was highly competitive on the May 2025 leaderboard (\suptab{tab:leaderboard_snapshot_mase_full}).

 \subsection*{Other Problems: Geospatial Analysis, Neuroscience and Numerical Analysis}
 We also evaluated ERA on problems from three additional distinct domains: geospatial semantic segmentation, vertebrate neural activity forecasting, and numerical analysis (Supplementary Notes). ERA achieved expert-level performance in each case (\supfigs{fig:dlrsd_example}--\ref{fig:integrals-scores}, \suptab{tab:dlrsd_performance_summary}).

\section*{Discussion}
Our work introduces ERA, an AI-based system that drives a  Tree Search (TS) with a Large Language Model (LLM) to systematically create and improve software for scientific tasks. By defining the problem of creating scientific software as a search for a program whose output maximizes a quality score, we convert software creation into a ``scorable task'', producing empirical software. ERA is novel in its LLM-driven rewriting approach, which allows for the flexible integration of domain knowledge and external research ideas. The ability of frontier LLMs to closely follow instructions enables efficient exploration of research ideas.
ERA builds upon ideas from several distinct but related areas of research: Genetic Programming, Generative Programming, the application of LLMs to code, Automated Machine Learning (AutoML), Combining LLMs and Search, and agents for scientific discovery.

{\it Genetic Programming} — Genetic Programming (GP) provides a foundation to our work. In GP, a population of programs is iteratively improved using evolutionary principles like selection, crossover, and mutation. The fitness of each program is determined by a ``fitness function,'' which is directly analogous to our ``quality score''~\cite{koza1994genetic}. While GP has been successful, it traditionally relies on random mutations and structured recombination of code fragments (e.g., swapping sub-trees in an abstract syntax tree). Prior to the widespread adoption of large language models, GP systems successfully incorporated human expertise by restricting the search space using formal grammars or leveraging structural code metrics to bias the evolutionary search~\cite{schweim2021using}. A key difference in ERA is the use of an LLM to perform intelligent, semantic-aware "mutations" by rewriting the code, which can produce more complex and meaningful variations than the random changes typical in GP.

{\it Generative Programming} — ERA can be viewed as a modern AI-driven realization of traditional generative programming, where, a developer creates a program generator (using techniques like templates, domain-specific languages~\cite{mernik2005and}, or metaprogramming) that produces tailored source code for a family of related problems~\cite{czarnecki2002generative}. In contrast, we use an LLM guided by a tree search as the  generative engine. This approach offers greater flexibility, allowing ERA to synthesize novel programs by exploring a vast solution space and integrating diverse domain knowledge in ways not easily achievable with more template-based methods.

{\it LLMs for Code Generation} — The advent of large language models pre-trained on vast code corpora has revolutionized code generation. Systems like AlphaCode~\cite{li2022competition} and OpenAI Codex~\cite{chen2021evaluating} have demonstrated the ability to generate correct and complex code from natural language descriptions. These systems are typically used for ``one-shot'' generation from a prompt. Our approach differs by using the LLM in an iterative refinement loop. Instead of generating code from scratch, our LLM rewrites existing software candidates, guided by a search algorithm (TS) that uses the quality score as a signal.

{\it AutoML} — AutoML systems aim to automate the process of building machine learning pipelines by searching for optimal model architectures and hyperparameters. The goal is to maximize a performance metric (e.g., accuracy, F1-score) on a validation dataset~\cite{hutter2019automated}, which fits our definition of a scorable task. While AutoML focuses specifically on finding the best model within a fixed set of ML frameworks, ERA is more general. It can rewrite any software, including pre-processing steps, complex simulations, mathematical heuristics, and other areas that fall outside the typical scope of AutoML.

{\it Combining LLMs and Search} — The most closely related work involves combining LLMs with search algorithms to overcome the limitations of one-shot generation.  A rapidly growing body of literature formulates automated algorithm design as an Evolutionary Algorithm (EA)\cite{wu2024evolutionary, ma2025toward}. FunSearch uses an LLM to search for new mathematical discoveries by pairing a LLM suggesting code improvements with an automated evaluator~\cite{romera2024mathematical}. Very recently, agentic frameworks like AlphaEvolve~\cite{novikov2025alphaevolve} have expanded these evolutionary coding loops towards needle-in-the-haystack discovery problems, similar in spirit to the tree search algorithms explored here, albeit without idea exploration required for scientific problem solving.    

{\it Agents for science problems} — This sub-field has seen remarkable performance from highly specialized systems,  with agents focusing on problems ranging from data science~\cite{ifargan2024data} to computational biology~\cite{xiao2024cellagent}. Instead of specializing in one domain, ERA demonstrates 
a general capability for empirical software optimization, achieving expert-level
performance on public leaderboards and in academic literature across multiple fields.

To our knowledge, this is the first work demonstrating a system that beats human performance on a wide range of relevant and well-studied scientific problems. Across different fields, by prompting with the method description from cutting edge papers, ERA not only produces code that follows the methods of the paper but often has superior results.  

We would like to emphasize the critical distinction between optimizing empirical predictive models and performing genuine scientific discovery, the latter of which requires reasoning about underlying theories, causal mechanisms, and mathematical frameworks. While the core problems evaluated in this paper primarily emphasize advanced empirical software engineering to allow for rigorous, automated scoring, our underlying system goes well beyond this, towards scientific discovery (Supplementary Notes).

 The ability of LLM-based systems to autonomously produce expert-level empirical software carries broader safety risks, particularly when applied to domains that could directly impact human well-being. By automating complex engineering workflows, our system significantly lowers the technical expertise required to execute sophisticated computational tasks. While this democratization accelerates beneficial scientific discovery, it concurrently lowers the barrier to entry for deploying advanced models in sensitive or potentially dangerous domains. This phenomenon represents a broader, systemic risk associated with the combination of large-scale inference-time compute and exponentially increasing foundation model quality.

To summarize, ERA combines a code mutation system based on Tree Search~\cite{silver2016mastering} with the ability to integrate complex research ideas. Such research ideas could come from the published literature, from research agents (e.g.~\cite{gottweis2025towards,GeminiDeepResearch}) or from combining previous ideas and solutions that the LLM has found itself. Because ERA creates code that can follow a specific idea, it can search over externally-supplied research ideas. 
ERA created 40 methods that beat the best known method for scRNA-seq batch integration and 14 methods that outperformed the CDC ensemble for epidemiological prediction. Additionally, ERA achieved expert-level performance on geospatial reasoning, neural activity prediction and algorithms for computational mathematics and a novel rule-based strategy for time series prediction.

Trial and error is essential to scientific progress, both for humans and for the automated approaches we outline here. 
ERA generates expert-level solutions extraordinarily quickly, reducing exploration of a set of ideas from weeks or months, to hours or days. Accelerating research in this way
has profound consequences for scientific advancement. Based on this work, we believe that progress
in scientific fields where solutions can be scored by machines is on the precipice of a significant
acceleration.


\newpage
\section*{Methods}

\subsection*{Code mutation system}

We prompt an LLM providing a description, the evaluation metric and the relevant data. The LLM produces Python code, which is then executed and scored on a sandbox. 
Searching over strategies dramatically increases performance:
The system uses the score together with output logs and other information to hill climb towards a better score. We used a  tree search (TS) strategy with an upper confidence bound (UCB) inspired by AlphaZero \cite{silver2017mastering}. A critical difference from AlphaZero is that our problems don't allow exhaustive enumeration of all possible children of a node, so every node is a candidate for expansion. We therefore modify the UCB algorithm to count visits and compute mean values using the tree. However, when sampling a node to expand, we sample directly from the whole set instead of recursing from the root like AlphaZero. This makes our method closer to Flat UCB \cite{coquelin2007bandit} than any MCTS variant like AlphaZero.

We also note that the algorithm differs from traditional TS, in that the scoring of the nodes do not involve random rollouts (e.g. of a game) to estimate the value of a node. Yet there is still randomness for scoring each node,  caused by the sampling of the LLM itself, which produces a distribution of different codes (scores) for each fixed prompt.  

We use a PUCT tree search algorithm to explore the space of code~\cite{silver2016mastering}. The PUCT (Predictor + Upper Confidence bound applied to Trees) algorithm is described in Algorithm~\ref{alg:puct}. For tree $T$, and executed candidate $u$, we define the flat prior $P_T(u) = \frac{1}{|T|}$. To make it easier to tune  the exploration constant $c_{puct}$ across tasks, we convert task-specific scores $\text{TaskScore}(u)$ to rank scores $\text{RankScore}_T(u)$ in the PUCT formula. We define $\text{RankScore}_T(u) = \frac{\text{Rank}_T(u) - 1}{|T| - 1}$, when $|T| > 1$, and 1 otherwise, where $\text{Rank}_T(u)$ gives ascending-order ranks to the candidates.

To select the next node for expansion, the algorithm balances exploitation of high-scoring solutions with exploration of the search space by computing a PUCT-inspired (Polynomial Upper Confidence Trees) acquisition score for each node $i$:
\begin{equation}
    \text{PUCT}_i = r_i + c_{\text{puct}} \cdot E(i)
\end{equation}
where $c_{\text{puct}}$ is the exploration constant and $E(i)$ is an exploration term based on the node's visit count relative to the total number of visits across the entire tree. We tuned $c_{\text{puct}}$ on the Kaggle benchmark to maximize performance, finding $c_{\text{puct}}=1$ works well. 
The node with the globally maximum $\text{PUCT}_i$ score is selected. The language model then generates a single novel child solution conditioned on this parent's code and score. The new code is executed, assigned an empirical score, and appended to the search tree, followed by a backpropagation of the visit count to its ancestors. This global, flat-tree structure allows the system to seamlessly backtrack and branch from \textit{any} historical node if the current optimization path plateaus.

We emphasize that in our algorithm mutations occur at the {\sl code level} -- ideas are part of prompts that produce code, that is then scored and iterated upon. With increasing nodes in the tree, the
score saturates after 300-1000 nodes (See Breakthrough plots in Supplementary Figures). We search over ideas by combining different tree searches with different prompts together. 

Finally, the implementation we outline here was chosen for optimal performance on the Kaggle Benchmark. Many alternatives were explored, including different agentic configurations and the simple implementation outlined here had the highest overall performance across the benchmark. All experiments in this paper were carried out with Gemini 2.5 Flash, with the improvement with Gemini 2.5 Pro modest.

\begin{algorithm}[h!]
\caption{UCB tree search (PUCT)}\label{alg:puct}
\begin{algorithmic}[1]
\Require GenerateAndExecute(), TaskScore() to define rank scores $\text{RankScore}_T(u)$, exploration constant $c_{puct}$, and a root node $r$.
\State $T \gets \{ r \}$ \Comment{Initialize the tree with a root node.}
\State $V(r) \gets 1$
\ForAll{iterations}
   \State $N_{total} \gets \sum_{u \in T} V(u)$ \Comment{Get total visits across all nodes}
   \State Select $u^* \gets \text{argmax}_{u \in T} \left( \text{RankScore}_T(u) + c_{puct} P_T(u) \frac{\sqrt{N_{total}}}{1+V(u)} \right)$ \Comment{Select node with highest PUCT score}
   \State $u_c \gets \text{GenerateAndExecute}(u^*)$ \Comment{Expand the selected node and Execute}
   \State $T \gets T \cup \{u_c\}$
   \State $V(u_c) \gets 1$
   \ForAll{ancestors $u_a$ of $u_c$ (excluding $u_c$)} \Comment{Backpropagate results}
        \State $V(u_a) \gets V(u_a) + 1$
   \EndFor
\EndFor
\State \Return $\text{argmax}_{u \in T} \text{TaskScore}(u)$ \Comment{Best solution found}
\end{algorithmic}
\end{algorithm}

It is useful to explicitly contrast this algorithm with standard heuristic search algorithms. Our approach does not rely on beam search (which aggressively prunes unselected candidates at each depth layer) nor a $(1+\lambda)$ evolution strategy (which maintains only the most recent generation and discards historical states).  

Similarly, our system differs from  emergent efforts to incorporate LLMs into GPs~\cite{wu2024evolutionary, ma2025toward,hemberg2024evolving,meyerson2024language,grishina2025fully}.

More recently, agentic AutoML frameworks encompass automated feature engineering, meta-learning, dynamic pipeline construction, and multi-objective optimization \cite{he2021automl}, including agentic systems \cite{fang2024mlzero,li2024autokaggle}, which have shown more impressive performance\cite{grosnit2024agentk}. 

\subsection*{Adding research ideas to the code mutation system}

When an expert solves difficult scientific problems, they often search for prior work for ideas. Prior work could be sourced from highly cited papers, specialized textbooks, or search engines. The search for prior work can also be powered by LLMs \cite{gottweis2025towards,GeminiDeepResearch,PerplexityDeepResearch,du2025deepresearch,coelho2025deepresearchgym,xu2025comprehensive,baek2024researchagent}. 

We emulate the expert behavior by injecting instructions for carrying out research ideas into the prompt of our code mutation system (\figref{fig:schematic}). 
We applied the research instruction injection for scRNA-seq batch integration, COVID prediction, segmenting remote sensing images, and whole-brain neural activity prediction. While the most successful outcomes used top methods from the literature, we also used two LLM driven search strategies:
Deep Research from Gemini 2.5 Flash~\cite{GeminiDeepResearch} and AI co-scientist~\cite{gottweis2025towards}. 

For running these searches, we provided the tools with background information from the main problem description, and instructed the models to create distinct ideas (\suptab{table:sc_deep_research_prompt}). After manually filtering proposals and removing one proposed scRNA-seq batch integration method, we prompted Gemini to format the ideas into a structure consistent with our baseline method descriptions (\suptab{table:sc_deep_research_format}). Finally, we ran ERA on these ideas to create empirical codes that could be scored.

\subsection*{ERA versus best-of-N}
We use $N=128$ ERA search nodes and compare the performance of Gemini 2.5 Flash, Mistral Medium, Claude Sonnet 4.6, GPT-5, and Gemini 3.1 Pro on the scRNA-seq batch integration task and an epidemiological flu forecasting task. This explores a range of models that were contemporaneous with the core model used in this paper (Gemini 2.5 Flash) while also using 
Gemini 3.1 Pro as a state-of-the-art model at the time of paper publication. The scores use the same scoring rules as in all other experiments: for the epidemiological forecasting task it is the Weighted Interval Score evaluated over a rolling evaluation window, while for scRNA-seq batch integration it is the average of metrics measuring preservation of biological information while eliminating batch effects, both measured on the validation set. Note that significant improvements in batch integration happen with much smaller change in validation scores than in epidemiological forecasting.   We present here the best performance over 3 different experiments (with both Best-of-N and ERA).  Typical variation in performance is of order $0.01$ for batch integration and $O(1)$ for epidemiological forecasting, with some model dependence. For example, GPT-5 has much more experiment-to-experiment variability than other models. 

We note that ERA explores the solution space more efficiently than Best-of-N, and outperforms Best-of-N for all models and problems, with the exception of GPT-5's performance on batch integration. On the batch integration problem, GPT-5's one shot performance is sufficiently good that this distinction isn't important.  Indeed we expect that as frontier models improve, tasks that were once difficult will saturate. Tree search is useful for pushing model performance on difficult tasks.

\subsection*{Recombination experiments}

For both scRNA-seq batch integration problem and COVID-19 forecasting, we combined ideas from methods already generated using tree search. For the scRNA-seq batch integration problem, we used the first versions of our 11 baseline methods. For the COVID-19 prediction problem, we used the eight replications of models submitted to CovidHub. We first took the top-performing node from each tree search run seeded with one of these methods, based on its score on the validation set (for COVID-19 prediction, this included six weeks of reference dates from 2025-02-22 to 2025-03-29). Then, for every pair of these methods, we prompted Gemini 2.5 Flash to compare the two methods and explain the core technical similarities and differences between the two parent models using a consistent prompt (\suptab{table:sc_recombine_summary_prompt}). The explanatory response was then added to the the prompt, along with a statement instructing tree search to recombine the ideas by combining the best parts of both approaches (\suptab{table:sc_recombine_ts_prompt}). Subsequently, we ran ERA to generate new hybrid strategies. This process yielded 55 recombined methods for the scRNA-seq batch integration problem, and 28 for the COVID-19 prediction problem (evaluated on the three-week holdout set 2025-04-05 to 2025-04-19, see \figref{fig:covid_figure}e).

Our method for recombination differs from previous efforts where the LLM acts as a mutation and crossover operator\cite{meyerson2024language,lehman2023evolution, grishina2025fully, hemberg2024evolving,vanstein2024llamea,liu2024evolution,ye2024reevo}.  Our approach to recombining expert ideas is conceptually related to evolutionary methods such as RHEA, with the comparative advantage that our tree search recombines expert ideas directly in conceptual space via LLM prompts rather than solely in distilled model space\cite{meyerson2024unlocking}.

\subsection*{Gemini embeddings} For each tree search implementation, we input the code snippets to the Gemini text embedding model~\cite{lee2025gemini}, and the resulting 3,072-dimensional output vectors served as the semantic representations of their respective implementations.

\subsection*{scRNA-seq batch integration}

For all scRNA-seq experiments, we ran tree search with 500 nodes. Each experiment took roughly seven hours to execute on our infrastructure.

\paragraph{Dataset}
We sourced a dataset from CZ CELLxGENE Discover~\cite{czi2025cz} to use for hill climbing with tree search. To identify datasets distinct from the six OpenProblems.bio test datasets but that have similar characteristics, we filtered to datasets that contain only healthy human cells, with primary cell count $\geq$ 2,000, at least 10 unique cell types, at least seven unique donor ids (i.e. number of batches), and contain at least two unique assays that are also present in the OpenProblems.bio datasets. This filtering process identified 22 candidate datasets. After manually investigating the candidate datasets, we selected the dataset \texttt{364bd0c7-f7fd-48ed-99c1-ae26872b1042} version \texttt{ffdaa1f0-b1d1-4135-8774-9fed7bf039ba}~\cite{xu2023inthcad}.

Within the selected dataset, we applied quality control metrics and data processing steps identical to the processing performed on the OpenProblems.bio datasets~\cite{oprepo,tbirepo}, yielding a processed dataset with normalized expression values, highly variable genes, principal components, and $k$-nearest neighbors all computed. For computational efficiency, we randomly selected two disjoint subsets of $N=20,000$ cells each, attempting to match \texttt{(batch, cell type)} distributions of the entire processed dataset. The ``train'' dataset was used for model training and selection of the highest-performing node in a single tree search. The ``validation'' dataset was used to select the best tree search for methods in which we ran multiple replicates of the same algorithm (\supfig{fig:dataflow_sc}).

\paragraph{Evaluating scRNA-seq Batch Integration on the OpenProblems.bio Benchmark}
We downloaded the OpenProblems \texttt{v2.0.0} input and solution data from \url{s3://openproblems-data/resources/task_batch_integration/datasets/cellxgene_census/} and raw performance metrics from \url{s3://openproblems-data/resources/task_batch_integration/results/run_2025-01-23_18-03-16/score_uns.yaml}. We computed control-scaled metric results identically to the published OpenProblems results. Briefly, for each \texttt{(dataset, metric)}, lower and upper bounds on raw scores are defined as the minimum and maximum values achieved by the seven ``control'' methods. Raw values were linearly scaled between those extrema and clamped to be in $[0, 1]$. Overall score was computed as the arithmetic mean over all 78 measurements (13 metrics computed for each of 6 datasets) with NaN values replaced by 0 (i.e., failure to compute a metric causes it to be considered the worst possible score).

\paragraph{Replication of Existing Methods for Batch Integration}

The OpenProblems.bio benchmark profiles the performance of several state-of-the-art existing methods. As of July 11, 2025 there were 19 different methods. Three methods have implementations in both R and Python: \texttt{LIGER} and \texttt{pyliger}, \texttt{Harmony} and \texttt{Harmonypy}, and \texttt{batchelor mnnCorrect} and \texttt{mnnpy}. After grouping reimplementations of the same method, there are 16 separate research ideas. From this list, we excluded all six foundation model methods (\texttt{UCE}, \texttt{SCimilarity}, \texttt{scGPT (zero shot)}, \texttt{scGPT (fine-tuned)}, \texttt{Geneformer}, and \texttt{scPRINT}) because they perform very poorly on the benchmark and use a much larger training set. For example, only a single foundation model (\texttt{UCE}) performs better than the negative control of ``No integration'' which simply performs PCA on the dataset. We further excluded \texttt{scANVI}, which is a modification of \texttt{scVI} that is trained using cell type information. Since cell type information is used to define the metrics, this represents data leakage and consequently we consider \texttt{scANVI} a control method. This resulted in nine existing different research methods to optimize with tree search.

For each of the nine existing methods, we obtained the manuscript PDF corresponding to the method. To obtain a short method description from the manuscript, we used Gemini 2.5 Pro Thinking to summarize the paper (prompt in \suptab{table:sc_baseline_prompt}, example output in \suptab{table:sc_baseline_prompt_output_bbknn}). For \texttt{batchelor fastMNN}, which is a faster implementation of \texttt{batchelor mnnCorrect}, there is no separate publication and thus we provided the paper PDF of \texttt{batchelor mnnCorrect} as well as the docstring corresponding to \texttt{batchelor fastMNN} from \url{https://rdrr.io/github/LTLA/batchelor/man/fastMNN.html} (Details section) with a slightly adjusted prompt. Finally, the method summary is added to the tree search prompt, and is used to come up with better code solutions given the method summary.

For each of the nine methods, we ran three replicates of tree search. For \figref{fig:batch_integration}, we selected the replicate that had the best performance based on the validation set score. We show the performance of all replicates in \extfig{fig:barplot_replicates}. Code for the best performing method is in \suptab{table:sc_bbknn_code}.

\paragraph{Hyperparameters}To determine optimal hyperparameters for each base method, we employed Optuna, an automated hyperparameter optimization framework~\cite{optuna_2019}. Search spaces were defined across integer, float, and categorical parameter types by experts. The optimization process ran for a total of five times the number of parameters. In each trial, a model was trained using a sampled parameter set and evaluated based on a performance metric that Optuna's Tree-structured Parzen Estimator (TPE) sampler aimed to maximize. All hyperparameter optimization was conducted solely on the training dataset. The best identified hyperparameter set was then utilized to train the final base methods and evaluate them on the held-out OpenProblems dataset.

\subsection*{COVID-19 prediction}

\paragraph{Dataset} Our primary data source was historical confirmed COVID-19 hospital admissions, which corresponds to the target variable specified by CovidHub. These data are published weekly by the CDC within the National Healthcare Safety Network (NHSN) Hospital Respiratory Data (HRD) dataset~\cite{CDC_HRD_Data_2024_data}. Preprocessing was kept minimal--missing values in the dataset were replaced by zeros to enable tree search to find executable code with the criterion score (WIS).  The only additional data source used to augment the target for our model was static jurisdiction-specific population values from the CovidHub GitHub Repository~\cite{CovidForeCastHub}.
For comparing model performance in~\figref{fig:covid_figure}c, we use all of the models submitted to Forecast Hub which make predictions at a state by state level and  have forecasts for at least 75 percent of the season and time horizons.
We ran tree search with 2000 nodes for each reported run. We note that  we use  available as of 2025-05-01 for the entire retrospective season, thus ignoring potential differences in data available at the date of forecast.

\paragraph{Replication of existing COVID-19 prediction models}
We selected eight models for replication from those that had submitted to CovidHub based on the following inclusion criteria: (1) The method must be reproducible solely using historical COVID-19 hospitalization data, without reliance on external predictor variables, (2) The model submission must include predictions across all specified time horizons, and (3) Model submissions must be available for over three months (12 weeks) to enable meaningful comparison. Three models were excluded for failing these criteria: two were ensembles of external forecasts, and one relied entirely on additional data. An additional five models were excluded because they did not provide predictions for all forecast horizons. These five models originated from the same forecasting team. As all our analysis involves aggregating model performance across horizons, we have excluded these five models from all comparisons. Overall this gave a selection of eight models for replication.

To instruct the search algorithm, we provided the method descriptions from the original authors' official submission metadata. For example, the metadata for the \verb|UMASS-arc6-pooled| model states: ``\textit{AR(6) model after fourth root data transform. AR coefficients are shared across all locations. A separate variance parameter is estimated for each location.}'' We integrated these concise descriptions directly into the tree search prompt as part of the model directions, transforming them into instructions by prepending `Use a/an' (see Methods, \suptab{table:covid-rep-prompt}).

\subsection*{GIFT-Eval benchmark}

We applied our tree search methodology to the General Time Series Forecasting Model Evaluation (GIFT-Eval) benchmark \cite{aksu2024gift}.  The search begins from a root node defined by an initial code template and proceeds via hill climbing, where new candidate solutions are generated and evaluated against the GIFT-Eval validation folds. At the end of a tree search, we evaluated the solution on the held-out test set using MASE point forecast as the scoring metric. 
Our results are based on a 5/18/2025 snapshot of the dataset, official leaderboard and scoring, all of which have been updated since. See \suptab{tab:leaderboard_snapshot_mase_full} for a complete snapshot of the leaderboard.

We remark that our snapshotting to a fixed date ensures stability. The GIFT-Eval protocols underwent significant structural changes after our experiments concluded, including major scoring/dataset corrections on July 24, 2025, and the introduction of an "Agentic" category on August 5, 2025. Later updates on August 25 redefined "Zero-shot" to account for widespread test data leakage in foundation models using the large pre-train dataset (Note that our method does not use the large pre-training data).
To avoid unsound comparisons against baselines developed under these revised protocols, we deliberately chose not to submit to the live public leaderboard, restricting our comparison to the May 18th snapshot to ensure a valid, stable baseline. We also note that Gemini models driving our Tree Search have evolved significantly since our experiments.

We adhered to the benchmark's framework, utilizing the official dataset source from \href{https://huggingface.co/datasets/Salesforce/GiftEval}{Hugging Face}, its pre-defined training, validation, and test splits, as well as the scoring and evaluation code commonly used in the existing submission \href{https://github.com/SalesforceAIResearch/gift-eval/tree/main/notebooks}{notebooks}.

{\sl Per-dataset Solution}
We conducted separate tree searches for 92 of the 97 GIFT-Eval datasets, excluding the five largest  due to computational constraints; for these, the naive baseline score was used in order to produce the aggregated leaderboard score. For each dataset, we used a search of 300 nodes, with the system permitted to use a broad suite of machine learning libraries, including \texttt{scikit-learn}, \texttt{XGBoost}, and \texttt{statsmodels}.  \supfig{fig:gift-category} shows an analysis of the types of models used across the 92 different solutions.

{\sl Unified Solution}
Here, we created a single, unified forecasting library that could generalize across all 97 datasets. We used a tree search of over 1,000 nodes, guided by the geometric mean of the normalized MASE scores across all datasets, providing a single objective function to optimize. To force the model to reason from first principles, its access was restricted to basic libraries (\texttt{numpy}, \texttt{pandas}, and \texttt{holidays}).

The resulting solution consists of two components: a single forecasting library and a list of eight preset configurations. For each dataset, the best-performing configuration is identified on the validation set. This selected configuration is then used with the unified library to produce the final forecast on the test set, allowing the model to adapt its strategy without seeing test data.

The final solution was developed iteratively. An initial search yielded a base model with a MASE of 0.82. A key breakthrough occurred in a subsequent run when the search space was expanded to ten configurations and the system was advised to use the \texttt{holidays} library, which improved the MASE to 0.77 (\extfig{fig:tree_breakthrough_gift}). A final 500 node refinement run pruned the configurations to an optimized set of eight, achieving the final MASE of 0.734. 

The final solution  sequentially models and removes fundamental components of the series, with the final forecast being the sum of the individual component forecasts. This approach allows the model to be highly configurable while systematically accounting for different sources of variation in the data. This process is outlined with the following steps:

\begin{enumerate}
    \item \textbf{Preprocessing:} The input series first undergoes basic cleaning, including median imputation for any missing values. An optional log-transform (\texttt{log1p}) can be applied to stabilize variance in series with exponential growth patterns.
    \item \textbf{Base/Level Component:} A base level is established using simple but robust methods like a seasonal naive forecast or a rolling median of recent data points. This component captures the basic magnitude of the series.
    \item \textbf{Trend Component:} The residuals from the base component are then modeled to capture linear or polynomial trends. This step includes a \texttt{damping\_factor} to prevent unrealistic long-term extrapolation by gradually flattening the trend.
    \item \textbf{Seasonality Component:} The residuals from the trend component are analyzed to model cyclical patterns (e.g., weekly, yearly). The model identifies the cycle length and forecasts seasonality by averaging values at the same point in the cycle (e.g., the average value for all Mondays).
    \item \textbf{Datetime and Holiday Features:} To capture special events and non-seasonal cycles, features are extracted from the timestamp (e.g., \texttt{dayofweek}, \texttt{is\_holiday\_flag}). The model calculates the median effect of each feature category from the remaining residuals and adds it to the forecast.
    \item \textbf{Residual Correction:} As a final step, a correction is made by modeling the median of the most recent unexplained errors. This autoregressive-like step helps correct for short-term biases in the model. A \texttt{decay\_factor} fades its impact over the forecast horizon.
\end{enumerate}

To apply the unified solution to a new dataset, one would first split the historical data into training and validation sets. Using the library's adaptive configuration system, one can then find a suitable forecasting strategy by evaluating the eight preset configurations on the validation data to select the best-performing one. This provides a strong, data-driven starting point that can be used directly. For more specialized applications, one can also create a custom configuration, allowing for manual refinement of the model's components and making the library both powerful out-of-the-box and flexible enough for expert tuning.

\section*{Code Availability} 
A reference implementation of ERA is available at \href{https://github.com/google-research/era}{https://github.com/google-research/era}. The best candidate solutions generated for each of the six scientific problems in this paper are publicly available at \href{https://google-research.github.io/era}{https://google-research.github.io/era}, along with a user interface enabling examination of the full tree search data for a representative run for each of the six scientific problems. The interface allows inspecting the solution progression and breakthrough plot as the tree search proceeds and highlights code diffs.

\section*{Author Contributions Statement}
Code Mutation System (E.A., A.B., G.C., M.C., H.C., P.N., D.S., J.T., S.V., M.P., J.K., P.R., J.W., L. W., S.M. and M.P.B.)
Single Cell RNA-seq Batch Integration (A.Be., C.Y.M., C.H., Y.Z., M.P.B.)
COVID Forecasting (Z.S., S.M., M.P., M.C., M.P.B.)
Geospatial Analysis (R.J., Je.C., Q.Z., M.P.B.)
ZAPBench (B.P.W., J-M.L, Q.Z.)
GIFT-Eval (J.G., M.P.B.)
Integrals (A.K., R.K., M.P.B.)
User Interfaces (E.A., G.C., P.N., A.K., M.K., M.P.B., J-M.L., D.L., J.K., C.C., S.E.)
Graphical Design (G.J.)
Program Management (M.A., E.B.)
Leadership (K.C., J.M., Y.M., J.C.P., L.D., S.M., M.P.B.)

\section*{Acknowledgements}
We are grateful to our colleagues in Google Research and Google DeepMind for the incredible environment within which to do this work. We would like to specifically thank Niv Efron, Viren Jain, Anupam Pathak and Jamie Smith for many incisive discussions, Pablo Ruggia and Petkov Yotov for their help with LLM inference and scaling, and Nicholas Reich for comments on the manuscript.

\section*{Competing Interest Declaration}
The authors affiliated with Google are employees of Google Inc. and  hold Alphabet stock.

\clearpage
\newpage

\renewcommand{\thetable}{\arabic{table}}
\renewcommand{\thefigure}{\arabic{figure}}
\renewcommand{\figurename}{Extended Data Fig.} 
\renewcommand{\tablename}{Extended Data Table}
\setcounter{table}{0}
\setcounter{figure}{0}
\newpage

\begin{appendices}

\clearpage

\section*{Extended Data Figure Legends}

\begin{figure}[ht]
  \centering
  \includegraphics[width=1.0\textwidth]{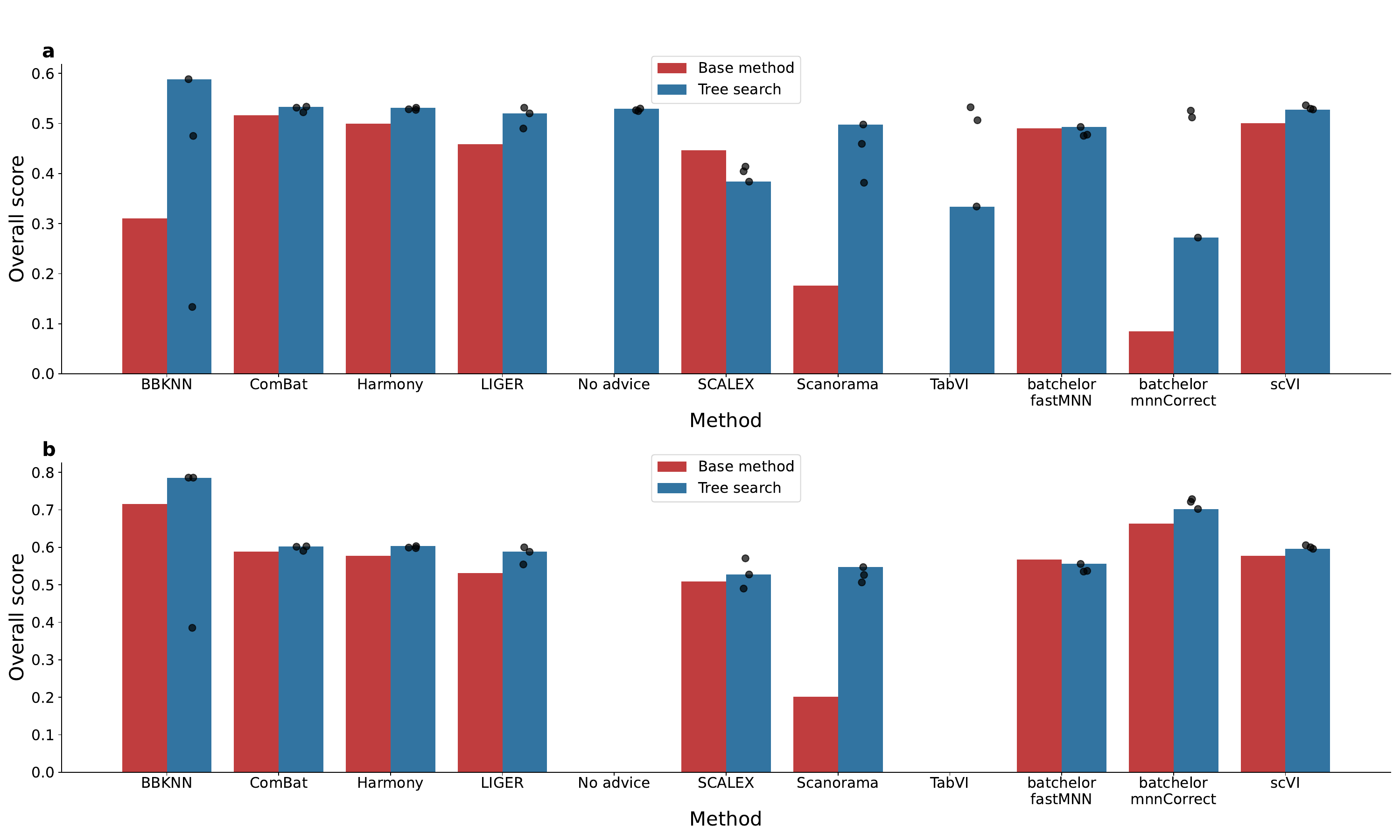}
  \caption{\textbf{Relative performance of base methods and ERA replicates.} \textbf{a,} Overall scores on the holdout OpenProblems datasets for all replicates of methods evaluated in~\figref{fig:batch_integration}. For tree search implementations, three replicates of the full process were performed. Dots indicate the overall score of the replicate on the holdout OpenProblems datasets. The bar shows the performance of the replicate with highest performance in the validation dataset (identical values to those shown in~\figref{fig:batch_integration}). The lowest performing tree search replicates for BBKNN, Scanorama, and TabVI only successfully computed 30, 57, and 45 of the 78 metrics, respectively. We note that failures due to out of memory or compute time issues were not explicitly selected against in our algorithm since all optimization was performed on datasets of only 20k cells.
  \textbf{b,} Average scores for each method when restricting to only \texttt{(method, dataset, metric)} combinations that have non-NaN values for the base method and all three tree search replicates. \texttt{No advice} and \texttt{TabVI} are absent since they have no base method comparator.}
\label{fig:barplot_replicates}
\end{figure}

\begin{figure}[h!]
    \centering 
    \begin{subfigure}[b]{0.9\textwidth}
        \centering
        \includegraphics[width=\linewidth]{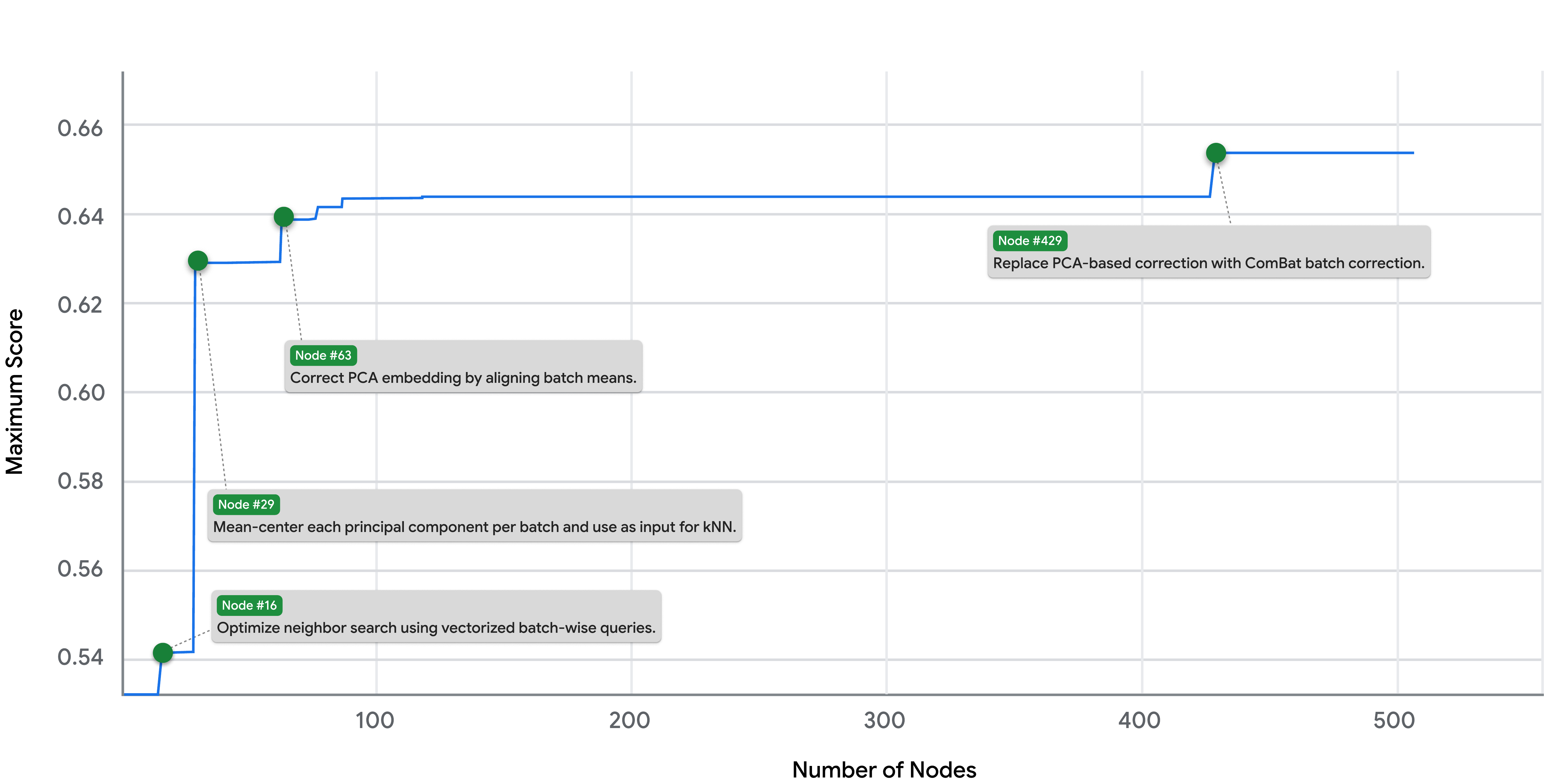}
    \end{subfigure}
    \hfill 
    \begin{subfigure}[b]{0.9\textwidth}
        \centering

        \includegraphics[width=\linewidth]{Figures_FINAL/Brenner_EDfig2b.pdf}
    \end{subfigure}
    
    \caption{\textbf{Breakthrough plot and solution tree for the scRNA-seq batch integration task.} {\sl Top Figure} Breakthrough plot for the \texttt{BBKNN (TS)} tree search, showing the evolution of the maximum score as a function of the number of nodes. The green dots label places where the score abruptly increases due to an improvement in the code, and the label describes the change in the code that resulted in the score increase. {\sl Bottom Figure} Structure of the tree for this same search. The color range consists of orange (lower scores) to green (higher scores) with the highest score denoted by a diamond node.}
    \label{fig:single_cell_tree_breakthrough}
\end{figure}

\begin{figure}[ht]
  \centering
  \includegraphics[width=1.0\textwidth]{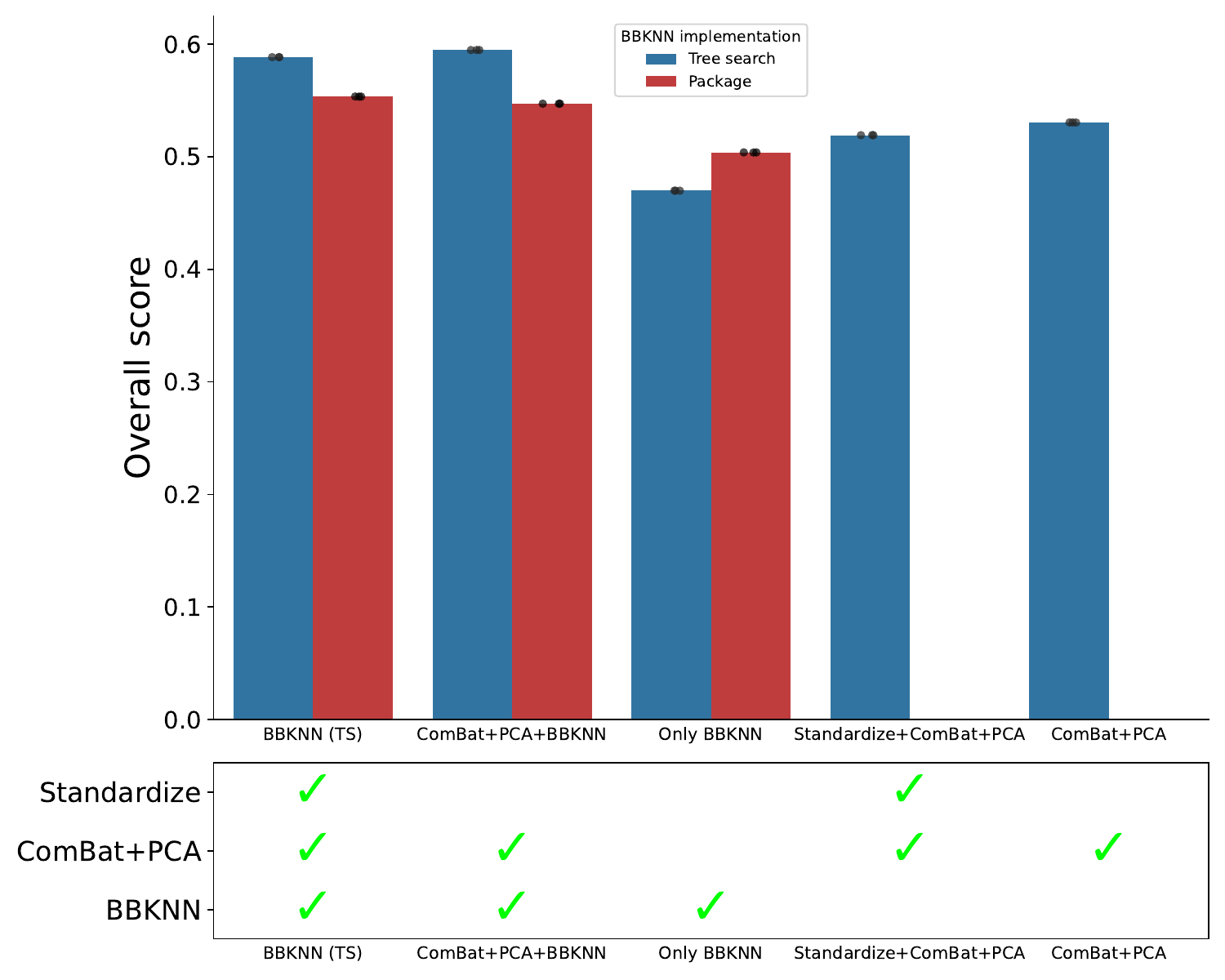}
  \caption{\textbf{Ablation analysis of the top-performing \texttt{BBKNN (TS)} method.} The \texttt{BBKNN (TS)} method performed standard linear expression scaling to $10^4$ total counts followed by \texttt{log1p} transformation. It then applied three additional transforms: ``Standardize'' called \texttt{sc.pp.scale} to further scale the data to mean 0 and unit variance, ``ComBat+PCA'' called \texttt{sc.pp.combat} followed by \texttt{sc.tl.pca} to generate the expression embedding, and ``BBKNN'' applied an implementation of batch-balanced $k$-nearest neighbors writted by ERA. Bars here show the overall performance in the OpenProblems datasets for ablations that include one or more of these components. For each ablation that includes the ``BBKNN'' component, comparison of the written BBKNN implementation (``Tree search'') and the \texttt{bbknn} package implementation (``Package'') is shown. Black dots show individual performance of three replicates of each method.}
  \label{fig:bbknn_deepdive}
\end{figure}

\begin{figure}[ht]
  \centering
  \includegraphics[height=0.75\textheight]{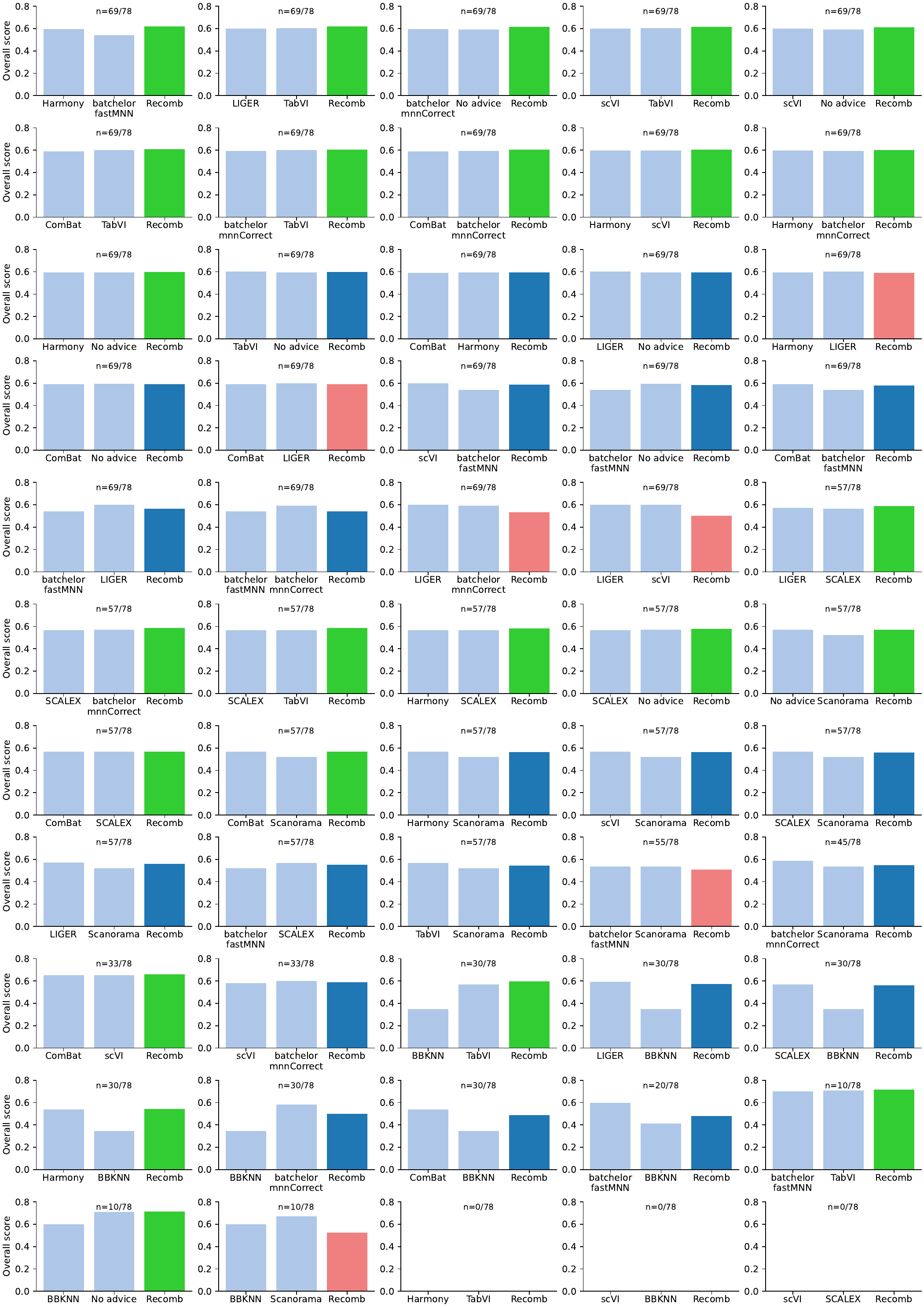}
  \caption{\textbf{Comparison of tree search performance on base methods and their ``recombination'' over an intersection of successfully calculated metrics.} We ran ``recombination'' experiments by seeding tree search with the top variants from two base method runs (see Methods). We compare the performance of two base methods and ``recombination'' on the OpenProblems test dataset for all 55 pairwise combinations of the 11 base methods. Since sometimes methods may fail getting a score for certain evaluation metrics due to errors like out of memory, we compare the performance on a subset of metrics that were successfully computed for all three methods. ``n=X/78'' on each subplot shows the number of successfully computed metrics, X, that we averaged over. For each subplot, we show the base methods on the left in light blue, and the recombination method on the right (labeled as ``Recomb''), where a green bar means the recombination method outperforms both of its base methods, dark blue means the recombination method outperforms one of the base methods, and red means the recombination method does not outperform either of the base methods.}
  \label{fig:recomb_bar_intersection}
\end{figure}

\begin{figure}[ht]
  \centering
  \includegraphics[width=1.0\textwidth]{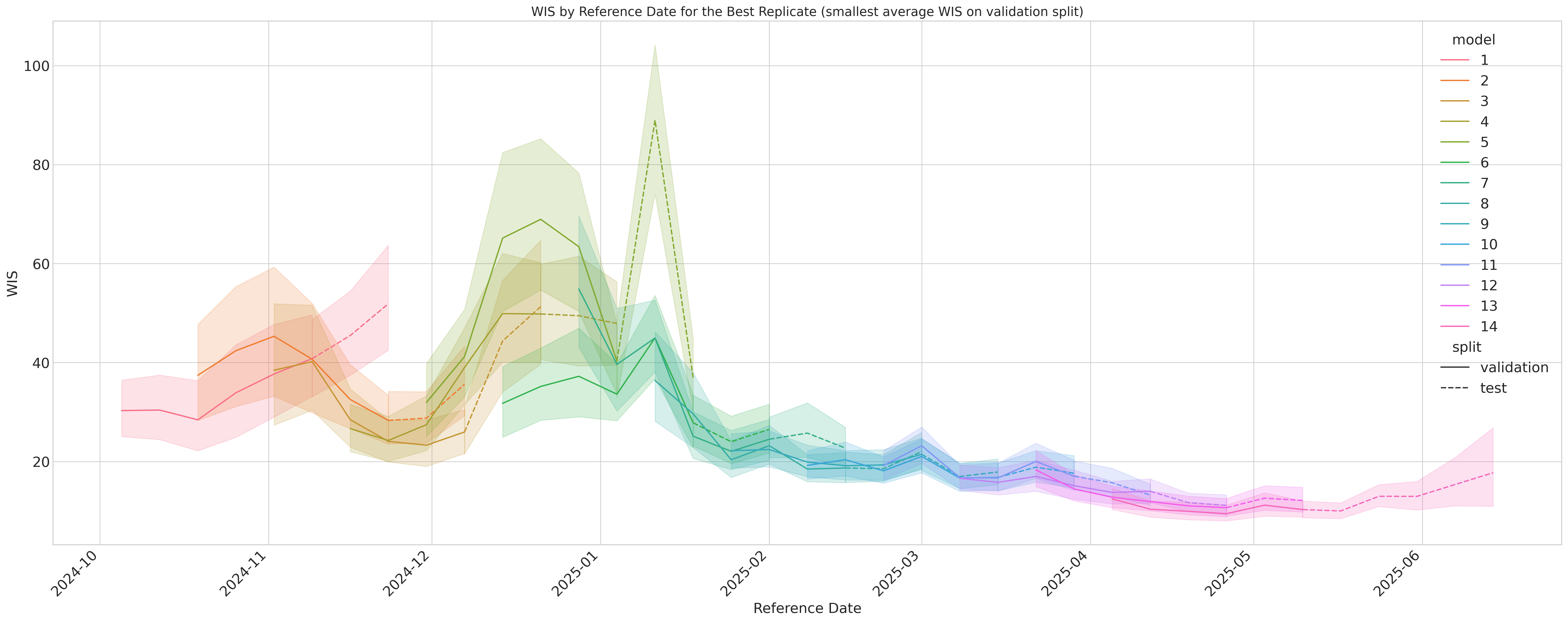}
  \caption{\textbf{Performance of the best retrospective COVID-19 hospitalization forecast replicates.} This figure presents WIS by reference date for the single best-performing replicate of each validation window in our retrospective COVID-19 forecasting study. The best models are selected based on their performance on the validation dates. The plot shows how finding optimum models on a handful of validation dates (6 weeks) generalizes to the next two weeks of unseen reference dates.}
  \label{fig:covid_retro_best_rep}
\end{figure}

\begin{figure}[h!]
    \centering 
    
    \begin{subfigure}[b]{0.9\textwidth}
        \centering
        \includegraphics[width=\linewidth]{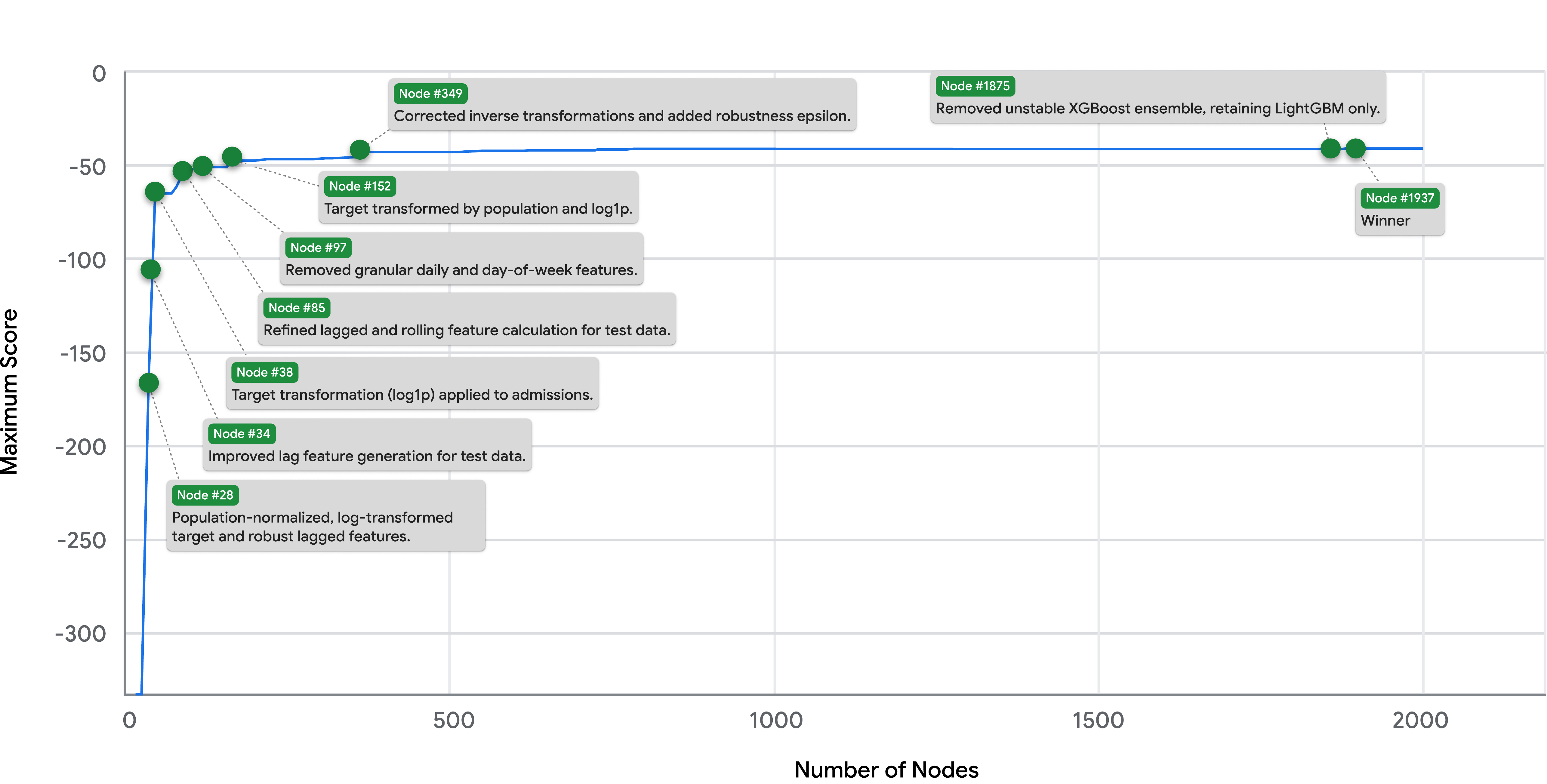}
    \end{subfigure}
    \hfill 
    \begin{subfigure}[b]{0.9\textwidth}
        \centering

        \includegraphics[width=0.6\linewidth]{Figures_FINAL/Brenner_EDfig6b.pdf}
    \end{subfigure}
    
    \caption{\textbf{Breakthrough plot and solution tree for the COVID-19 prediction task.} {\sl Top Figure} Breakthrough plot for the retrospective COVID-19 prediction, showing the evolution of the maximum score as a function of the number of nodes. The green dots label places where the score abruptly increases due to an improvement in the code, and the label describes the change in the code that resulted in the score increase. {\sl Bottom Figure} Structure of the tree for this same search. The color range consists of orange (lower scores) to green (higher scores) with the highest score denoted by a diamond node.}
    \label{fig:breakthrough_tree_covid}
\end{figure}

\begin{figure}[ht]
  \centering
  \includegraphics[width=1.0\textwidth]{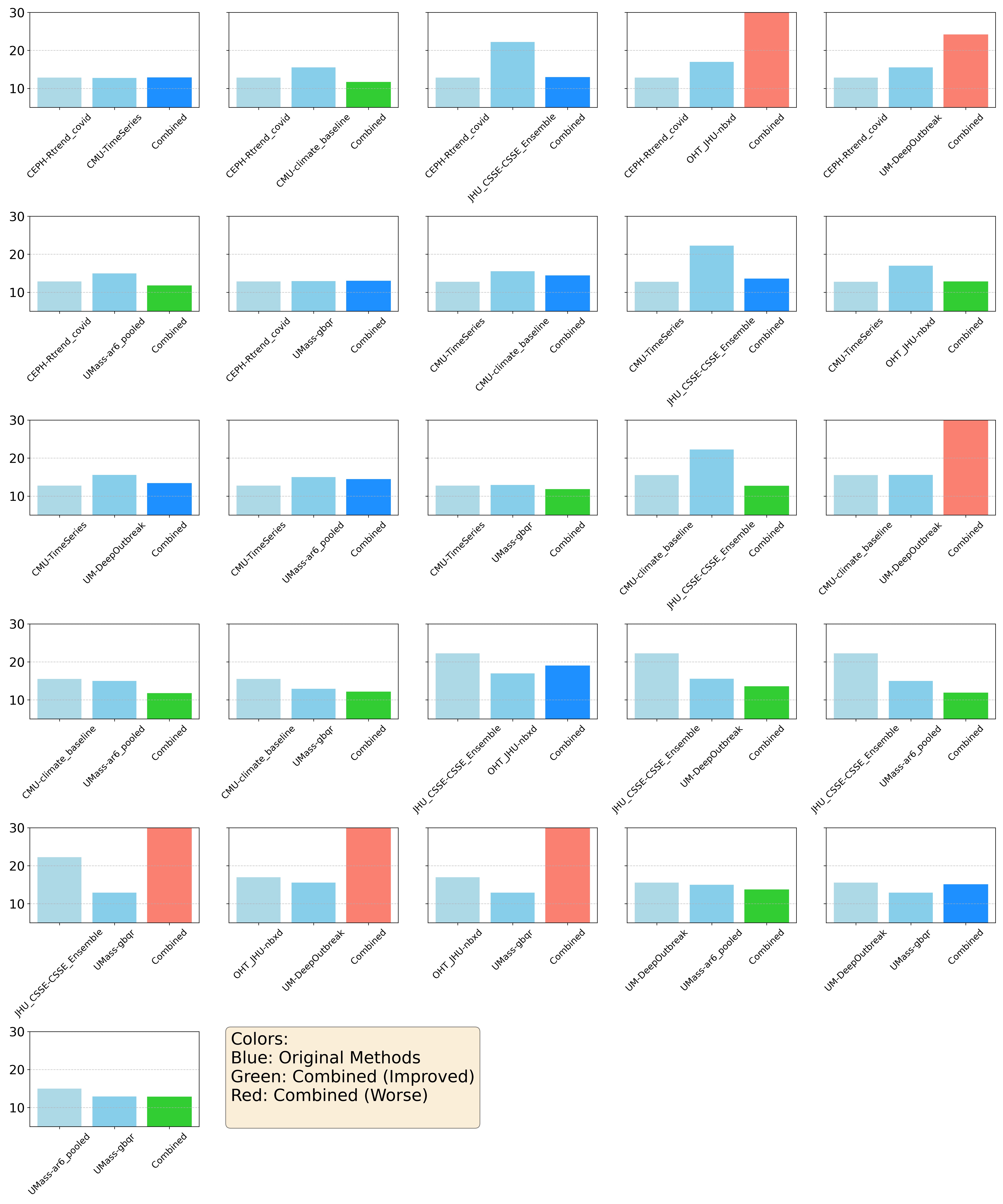}
  \caption{\textbf{Performance of recombination experiments for COVID-19 forecasting.} This series of bar plots illustrates the average WIS achieved by various hybrid models (right bar, labeled "Recomb") compared to their constituent baseline models (left bars, typically light blue) from the CovidHub competition. Each subplot represents a recombination experiment, demonstrating the success of our system in synthesizing novel forecasting strategies. Green bars indicate that the recombination outperformed both parent models, dark blue indicates it outperformed one, and red indicates it outperformed neither. These results emphasize the search system's ability to combine the strengths of existing methodologies to achieve superior predictive performance.}
  \label{fig:covid_recombined}
\end{figure}

\begin{figure}[h!]
    \centering 
    
    \begin{subfigure}[b]{0.9\textwidth}
        \centering
        \includegraphics[width=\linewidth]{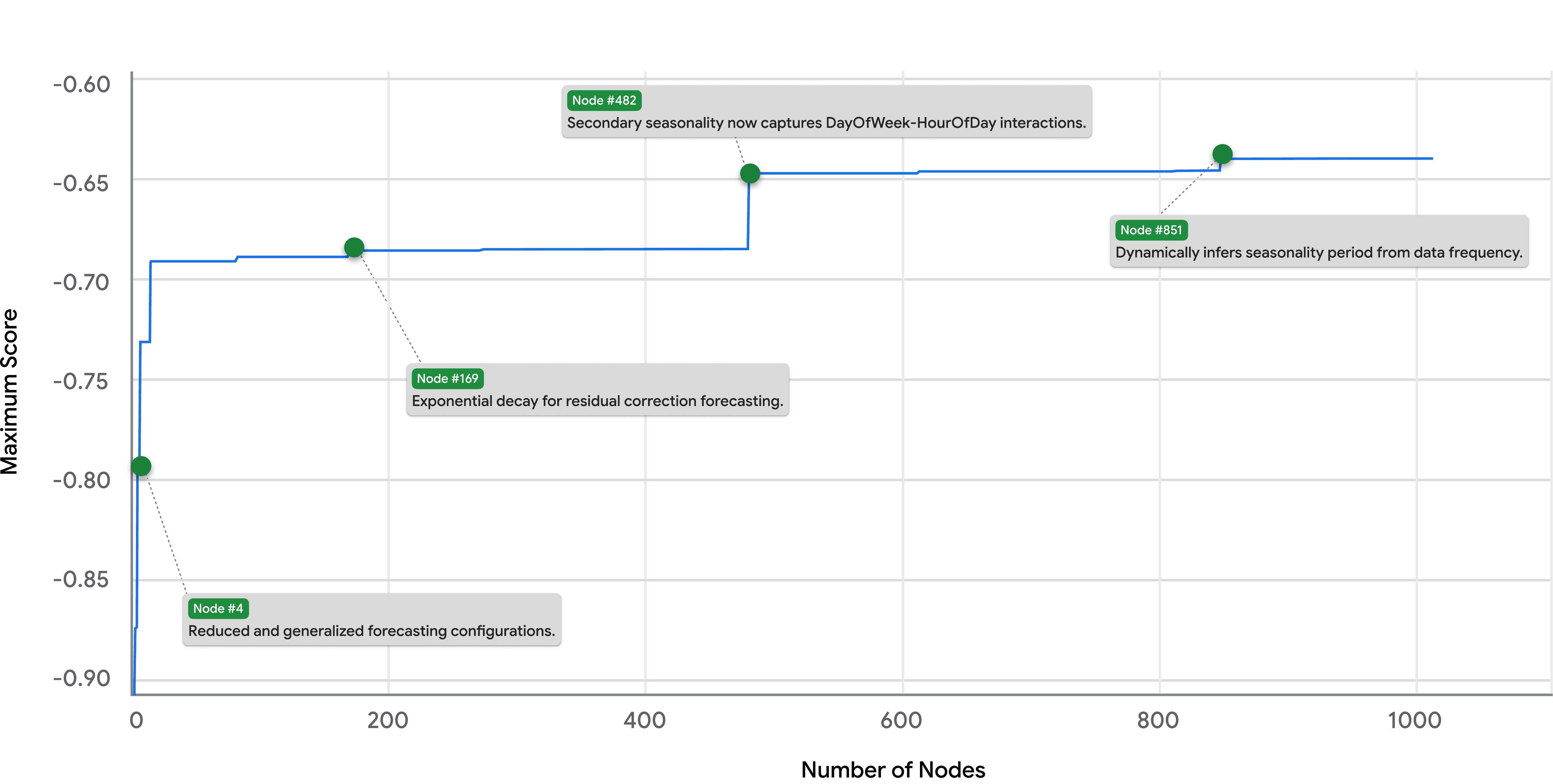}
    \end{subfigure}
    \hfill 
    \begin{subfigure}[b]{0.9\textwidth}
        \centering

        \includegraphics[width=\linewidth]{Figures_FINAL/Brenner_EDfig8b.pdf}
    \end{subfigure}
    
    \caption{\textbf{Breakthrough plot and solution tree for the time series prediction task.} {\sl Top Figure} Breakthrough plot for the GIFT-Eval tree search, showing the evolution of the maximum score as a function of the number of nodes. The green dots label places where the score abruptly increases due to an improvement in the code, and the label describes the change in the code that resulted in the score increase. {\sl Bottom Figure} Structure of the tree for this same search. The color range consists of orange (lower scores) to green (higher scores) with the highest score denoted by a diamond node.}
    \label{fig:tree_breakthrough_gift}
\end{figure}

\clearpage
\newpage
\subsection*{Extended Data Tables}

\begin{table}[h!tbp]
\renewcommand{\small}{\fontsize{9}{11}\selectfont}
\caption{\textbf{Expert manual inspection of adherence of \ourmethod implementation to scRNA-seq batch integration method.}}
\label{table:sc_baseline_method_adherence}
\centering
\fontsize{9}{11}\selectfont
\begin{tabular}{lclp{8cm}}
\toprule[1.5pt]
\textbf{Method} & \textbf{Replicate} & \textbf{Judgment} & \textbf{Notes} \\
\midrule
{\small batchelor fastMNN} & {\small 0} & {\small Follow} & {\small } \\
{\small batchelor fastMNN} & {\small 1} & {\small Follow} & {\small } \\
{\small batchelor fastMNN} & {\small 2} & {\small Follow} & {\small } \\
{\small batchelor mnnCorrect} & {\small 0} & {\small Follow} & {\small } \\
{\small batchelor mnnCorrect} & {\small 1} & {\small Follow} & {\small } \\
{\small batchelor mnnCorrect} & {\small 2} & {\small Follow} & {\small } \\
{\small BBKNN} & {\small 0} & {\small Follow} & {\small Adds distances between batches, performs spectral clustering on the graph. Does not compute connectivities.} \\
{\small BBKNN} & {\small 1} & {\small Follow + Innovative} & {\small Standardize + ComBat + PCA for embedding. BBKNN implemented on that embedding.} \\
{\small BBKNN} & {\small 2} & {\small Follow} & {\small Corrects the data, computes neighbors, final embedding is UMAP supposedly based on neighbors.} \\
{\small ComBat} & {\small 0} & {\small Follow} & {\small } \\
{\small ComBat} & {\small 1} & {\small Follow} & {\small } \\
{\small ComBat} & {\small 2} & {\small Follow} & {\small } \\
{\small Harmony} & {\small 0} & {\small Follow} & {\small Entropy-based diversity penalty.} \\
{\small Harmony} & {\small 1} & {\small Follow} & {\small Linear diversity penalty.} \\
{\small Harmony} & {\small 2} & {\small Follow} & {\small Linear diversity penalty.} \\
{\small LIGER} & {\small 0} & {\small Follow} & {\small Uses \texttt{sklearn.NMF} with multiplicative update solver.} \\
{\small LIGER} & {\small 1} & {\small Follow} & {\small Writes NMF function from scratch. Builds single global KNN graph rather than by batch.} \\
{\small LIGER} & {\small 2} & {\small Does not follow} & {\small Uses ComBat + SVD.} \\
{\small No advice} & {\small 0} & {\small Follow} & {\small Uses batch-specific mean+std for all genes to rescale. Then PCA.} \\
{\small No advice} & {\small 1} & {\small Follow} & {\small ComBat + SVD} \\
{\small No advice} & {\small 2} & {\small Follow} & {\small ComBat + PCA} \\
{\small SCALEX} & {\small 0} & {\small Follow} & {\small Adds log\_var clipping and weight normalization.} \\
{\small SCALEX} & {\small 1} & {\small Follow} & {\small Learns batch embedding. Learns gamma and beta conditioned on batch index. Batch index not supplied to first layer of decoder.} \\
{\small SCALEX} & {\small 2} & {\small Follow} & {\small Uses min\_delta for robust early stopping. batch\_index not supplied to the first layer of the decoder.} \\
{\small Scanorama} & {\small 0} & {\small Follow} & {\small } \\
{\small Scanorama} & {\small 1} & {\small Does not follow} & {\small Implements mnnpy via \texttt{sc.external.pp.mnn\_correct}.} \\
{\small Scanorama} & {\small 2} & {\small Follow} & {\small } \\
{\small scVI} & {\small 0} & {\small Follow} & {\small Applies log1p scaling with ZINB loss. Fits global dispersion theta rather than batch-specific.} \\
{\small scVI} & {\small 1} & {\small Follow} & {\small Applies optional log1p scaling with ZINB loss. Fits global dispersion theta rather than batch-specific.} \\
{\small scVI} & {\small 2} & {\small Follow} & {\small Expression frequency exponentiated rather than softmaxed. Applies log1p scaling with ZINB loss. Fits global dispersion theta rather than batch-specific.} \\
{\small TabVI} & {\small 0} & {\small Follow} & {\small } \\
{\small TabVI} & {\small 1} & {\small Follow} & {\small } \\
{\small TabVI} & {\small 2} & {\small Follow} & {\small } \\
\bottomrule[1.5pt]
\end{tabular}
\vspace{-0.2cm}
\end{table}



\clearpage
\newpage

\section{Supplementary Notes}
\subsection{Geospatial Analysis: Segmentation of Remote Sensing Images}

We ran ERA on a problem in geospatial analysis: 
semantic segmentation of high-resolution remote sensing images. Semantic segmentation is a computer vision task that involves assigning a specific class label to every single pixel in an image. It is essential for diverse applications, ranging from  monitoring land use, assessing the environmental impacts of human activity and managing natural disasters. The primary difficulty is  significant visual heterogeneity. Aerial and satellite images of the same location can differ dramatically due to variations in time of day, season, and weather conditions, while even objects
 within a single class (e.g. buildings) exhibit substantial diversity in size, shape, height, function and lighting conditions. 

A recent paper \cite{shao2018performance} introduces the ``dense labeling remote sensing dataset'' (DLRSD)  for 
advanced remote sensing tasks, including multi-label classification, image retrieval, and pixel-based applications like semantic segmentation. This dataset is a densely labeled version of the UC Merced Land Use Dataset \cite{yang2010bag}, a widely-used benchmark for image-level land use classification, whereby individual pixels of each image are labeled with 17 class labels. 

We prompted \ourmethod to train a model to classify pixels into the land cover classes and provided a pre-specified, reproducible 80/20 train/test split of imagery in the DLRSD dataset. 
For each experiment, we validated model performance on the held out test set of 420 randomly selected images using a standard ``mean intersection over union'' (mIoU) metric. 

The three top performing solutions generated by tree search significantly outperformed reported results in recent academic papers on the DLRSD benchmark, achieving mIoU greater than 0.80 (\suptab{tab:dlrsd_performance_summary}, \supfig{fig:dlrsd_example}). All three solutions build upon existing models, libraries and strategies. Solutions 1 and 3 leverage standard UNet++ and U-Net models but paired with powerful encoders (efficientnet-b7 and se-resnext101-32x4d)  pre-trained on ImageNet \cite{russakovsky2015imagenet}. Solution 2 uses SegFormer, a state of the art Transformer-based architecture.  
Key differentiators among the models included their data augmentation and prediction strategies. The U-Net++ and U-Net models leveraged extensive augmentation from the Albumentations library, whereas the Segformer model used a more basic set of transforms. 
All three solutions employ extensive Test-Time Augmentation (TTA) \cite{Krizhevsky} by predicting masks for multiple augmented versions of a single test image (e.g., horizontal flips, vertical flips, rotations) which are then reverse-transformed and averaged to produce a final, more robust mask which smooths out prediction errors and boosts performance.  A representative tree and breakthrough plot for Solution 3 is shown in \supfig{fig:tree_breakthrough_geospatial}.

\subsection{Neuroscience: Whole-Brain Neural Activity Prediction}

We consider
the Zebrafish Activity Prediction Benchmark (ZAPBench), a recent dataset designed to test predictions of cellular-resolution neural activity in an entire vertebrate brain  \citep{lueckmann2025zapbench}.  
The benchmark uses a novel dataset capturing brain activity of a larval zebrafish over a two-hour session using light-sheet fluorescent microscopy, resulting in 3D brain volumes recorded over time. Throughout the recording, the animal was exposed to distinct visual stimulus conditions designed to elicit a range of different behaviors. The raw volumetric video data was extensively processed to align, motion-stabilize, and segment into activity traces,  resulting in a final data matrix of activity traces for 71,721 neurons across 7,879 time steps.

Several state-of-the-art forecasting methods were evaluated on the benchmark \citep{lueckmann2025zapbench}, including time-series forecasting methods that operate on the extracted activity traces per neuron, as well as a volumetric video prediction model (a Unet variant) that directly processes the 3D brain volumes over time \citep{immer2025forecasting}. The video-based approach  exploits spatial information that is lost when converting the data to time series, but is computationally expensive. Among the different methods evaluated on the benchmark, the video-based Unet model achieved the best overall performance, especially in the setting where only a short window of past context is available.

We prompted \ourmethod to solve the multivariate time-series forecasting problem, predicting the output activity of all neurons for up to 32 time steps ahead in the time-series domain, given their past 4 time steps of activity as context, using the 
dataset splits provided by ZAPBench~\citep{lueckmann2025zapbench} which split each stimulus condition into $70\%$ for training, $10\%$ for validation, and $20\%$ for testing  per stimulus condition. We used the validation set for model selection, including hyperparameter tuning and early stopping, and to obtain a score to guide the tree search. We score solutions using mean absolute error (MAE) averaged across the prediction horizon, and compare solutions found by tree search against the methods included in ZAPBench: These include a linear model \citep{zeng2023transformers}, TiDE \citep{das2024longterm}, TSMixer \citep{chen2023tsmixer}, Time-Mix (a variant of TSMixer where feature mixing is ablated), and a custom Unet architecture \citep{immer2025forecasting}.

Our initial experiment using tree search led to a best-performing model that uses a rich feature set from the input window, combining temporal convolutions, a learned ``global brain state'', and neuron-specific embeddings. The model then processes these features through a series of weight-shared residual blocks and a final dense layer to generate the multi-step prediction in one shot. \supfig{fig:zapbench} shows the result of this model, compared to other baselines. In that figure, the mean baseline predicts the average over the context window, while the stimulus baseline predicts the average for each stimulus phase. Remarkably, the model produced by tree search outperformed all other baselines, including the best-performing video model, except for 1-step-ahead predictions. A representative example of the breakthrough plot and tree is shown in \supfig{fig:tree_breakthrough_zapbench}.

We then developed a separate model tuned specifically for 1-step-ahead predictions with another tree search. The resulting solution is conceptually similar to the first in that both architectures generate a learned global context vector to inform their per-feature predictions. However, this model computes its global context using a dynamic attention mechanism for weighted aggregation and modulates feature representations through a FiLM-like layer \citep{perez2018film} for interactive conditioning. This model achieved leading performance on 1-step-ahead predictions (\supfig{fig:zapbench}).

Both of these solutions are orders of magnitude faster to train than the best-performing video model--less than two hours on a single T4 GPU, as compared to 36 hours on 16 A100 GPUs for the Unet model. In addition, our solutions effectively use cross-neuron information to generate predictions, a major challenge highlighted in previous work~\citep{lueckmann2025zapbench}.

A key future direction is the development of models that incorporate biophysical information and are more interpretable. The forthcoming synaptic-level structural reconstruction of the larval zebrafish brain used for ZAPBench provides a unique opportunity to develop such models by integrating anatomical wiring diagrams. As an initial exploratory step, we prompted \ourmethod to use Jaxley~\citep{deistler2024differentiable}, a JAX-based library for differentiable simulation of biophysically detailed neuron models, for the tree search.
The resulting best-performing solution simulates each neuron independently using single-compartment Hodgkin-Huxley models\cite{hodgkin1952quantitative}. Crucially, it dynamically modulates each neuron's biophysical parameters based on its recent activity history. To account for inter-neuronal interactions without the computational cost of direct synaptic simulation, the model then processes the outputs of these independent simulations through a latent autoencoder. This learns a system-wide corrective signal, effectively modeling a \textit{functional} connectome--a reasonable hybrid approach in the absence of the structural connectome. While this model did not outperform the top-performing video model, it was competitive with time-series baselines (\supfig{fig:zapbench}).

\subsection{Numerical Analysis: Library for Numerical Evaluation of Difficult Integrals}

We applied ERA to the problem of numerical evaluation of difficult integrals using Gaussian quadratures. 
The gold standard, QUADPACK~\cite{piessens}, was developed in the 1980s and underlies the popular Python function \texttt{scipy.integrate.quad()}. 
However, this function can fail 
in multiple ways, among them: the underlying algorithm can fail to converge; the algorithm samples its integrand, and the sampling may miss important features; the algorithm loses precision when the problem exhibits precise cancellations.

While standard techniques exist to address these problems, we asked whether ERA could build a general-purpose method superior to \texttt{quad()}, by hill-climbing on a benchmark set of integrals for which the standard algorithm fails but the analytic solution is known. Our dataset consists of 38 oscillatory integrals with infinite upper limits and without other pathologies, for which \texttt{quad()} returned an incorrect answer (\supfig{fig:integrals-datasets}).

To build this list of integrals, we started with a long list of integrals in LaTeX form from Gradshteyn and Ryzhik~\cite{gandr}. 
We converted both the question and solution into a python expression using SymPy~\cite{meurer2017sympy}.
Most expressions included free parameters, often with value constraints. 
To enable numeric evaluations, we generated random values for all parameters consistent with the constraints.

Once an integral and its answer were in the form of SymPy expression objects, we evaluated answers numerically by substituting our chosen parameter values using \texttt{sympy.Expr.subs()} and evaluating via \texttt{sympy.evalf()}.
We build integrand functions suitable for \texttt{scipy.integrate.quad()} via \texttt{sympy.lambdify} for efficient evaluation.
We compared each numerical answer to the number returned by \texttt{scipy.integrate.quad()} and discarded cases where the numbers agreed within the latter's error estimate. 
We also discarded cases where that error estimate was greater than 2\% of the numbers' magnitude.

All conversions from LaTeX to SymPy and all constrained parameter generations were performed by Gemini using specialized prompts. The resulting SymPy expressions and parameter values were examined manually for correctness. These manual steps were the limiting factor on the scale of our dataset.

We randomly split these into training and evaluation datasets (n=19 integrals each). We initialized ERA with a simple invocation of \texttt{quad()} and prompted the system to improve it on a tree search over 1000 nodes. We scored solutions with the logarithm of the absolute fractional error (discrepancy between the generated solution's number and the answer's number), where the logarithm prevented the search from over-weighting outliers.

\begin{equation}
    \text{score} = - \log \left( 1 + \left| \frac{\text{response} - \text{answer}}{\text{answer}} \right| \right)
\end{equation}

The resulting breakthrough plot and tree structure for the search are shown in \supfig{fig:tree_breakthrough_integrals}.

The best solution builds on \texttt{quad()} by partitioning the infinite domain into a sequence of contiguous, finite subintervals whose lengths may increase geometrically to cover the domain's tail more efficiently.
The definite integral is thus transformed into an infinite series, where each term is the numerical integral of the integrand over one of these finite segments, calculated using \texttt{quad()}.
For integrals that converge slowly, such as those with oscillatory integrands, direct summation of this series is impractical.
The algorithm therefore applies Euler's transformation, a powerful series acceleration technique, to this sequence of segment integrals. 
By repeatedly averaging adjacent terms, the transformation extrapolates the limit of the slowly converging series from a finite number of its initial terms, providing an accurate estimate of the integral's true value. The evolved code correctly evaluated 17 out of 19 held-out integrals to within a fractional error of less than 3 percent (\supfig{fig:integrals-scores}).

The evolved code always applies \texttt{quad()} first. It only falls back to its more specialized methods if \texttt{quad()} returns a large error estimate, returns \texttt{NaN} or \texttt{Inf}, or raises an exception. This means the evolved code is as accurate as \texttt{quad()} in less pathological cases and so could reasonably be used as a drop-in replacement.

\subsection{Towards Genuine Discovery}
Several recent deployments  of our core tree-search algorithm  demonstrate the potential for genuine discovery:
\begin{itemize}
    \item \textbf{Mathematical Discovery} Our system was recently used to derive the exact analytical power spectrum of gravitational radiation emitted by cosmic strings \cite{brenner2026solving}. Operating without empirical data or predictive modeling, the system's search navigated the space of symbolic mathematics to discover six novel analytical derivations to an integral that had not been previously solved, yielding a closed-form exact solution.
    \item \textbf{Neural Mechanisms:} In computational neuroscience, the system was tasked with discovering transition models of neural activity in an \textit{in silico} larval zebrafish \cite{lueckmann2026discovering}. While unconstrained search yielded predictive models relying on statistical shortcuts, providing the system with structural priors (wiring diagrams) allowed the algorithm to autonomously discover the true, underlying mechanisms governing the neural circuit, successfully recovering causal effective connectivity.
\end{itemize}

 Within predictive modeling tasks, the scoring schemes are by construction standardized and straightforward to implement. Yet, when applying this automated framework to novel, open-ended scientific domains, inventing an appropriate scoring metric requires significant domain expertise and scientific creativity.
We illustrate how this can work for mathematical discovery \cite{brenner2026solving}, with the objective of discovering a closed-form analytical equation for the radiation spectra of cosmic strings. Our scoring metric measured how closely a proposed analytical solution matches a high-precision numerical solution:  the system evaluated the AI-proposed analytical formulas at random physical parameter values and computed the residual difference against the numerical integration. Minimizing this difference lets the  search discover six different  analytical solutions and also a novel asymptotic expansion.
In general, the design of metrics for complex scientific problems is an iterative endeavor. The scientific process  implicitly includes creating scoring functions to try to reach a hypothesized goal, iterating when the goal is not reached.  Within ERA,
this requires that the user iterates the scoring function.

\subsection{Computational Cost and Resource Utilization}
Since search-based LLM optimization relies on iterative generation and empirical evaluation, it is inherently more resource-intensive than single-shot or zero-shot prompting. Here we detail representative computational costs required per search node for representative search runs across our evaluated benchmarks in \suptab{tab:compute_cost}.


\clearpage

\renewcommand{\thetable}{S\arabic{table}}
\renewcommand{\thefigure}{S\arabic{figure}}
\renewcommand{\figurename}{Supplementary Fig.} 
\renewcommand{\tablename}{Supplementary Table}
\setcounter{table}{0}
\setcounter{figure}{0}

\newpage
\section*{Supplementary Figures}
\begin{figure}[h!t]
  \centering
  \includegraphics[width=0.84\textwidth]{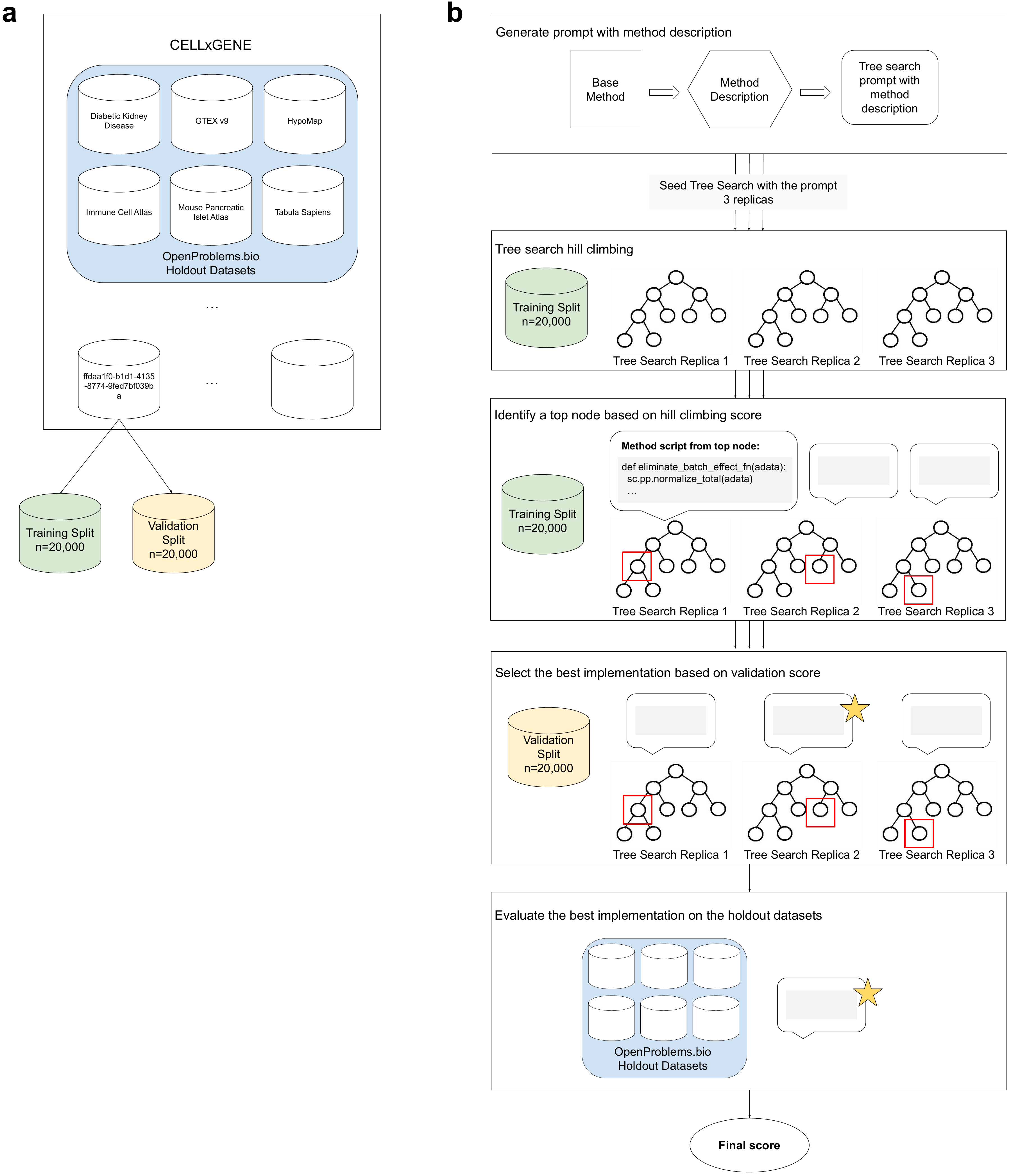}
  \caption{\textbf{Experimental design for single-cell batch integration.} \textbf{a,} We sourced our tree search development dataset from CELLxGENE. After filtering and manually selecting the dataset \texttt{364bd0c7-f7fd-48ed-99c1-ae26872b1042} version \texttt{ffdaa1f0-b1d1-4135-8774-9fed7bf039ba} (see Methods), which has a similar profile to the six datasets used in the OpenProblems.bio Batch Integration benchmark (distinct datasets also in CELLxGENE), we sampled 20,000 cells for the training split and 20,000 for the validation split. \textbf{b,} For each of the 11 base methods, we generated a detailed method description and inserted it into a prompt to initialize the tree search. We ran three independent tree search replicas per method, using the training split for hill climbing. From each tree, we selected the top-performing node based on its training score. We then evaluated each top node's script on the validation split and selected the best one based on validation performance. The best implementation per method was finally evaluated on the OpenProblems.bio holdout datasets, and the corresponding scores are reported as final results.}
  \phantomsection
  \addcontentsline{toc}{subsection}{\protect\numberline{Fig. S1:}Experimental design for single-cell batch integration.}
  \label{fig:dataflow_sc}
\end{figure}

\begin{figure}[ht]
  \centering
  \includegraphics[width=1.0\textwidth]{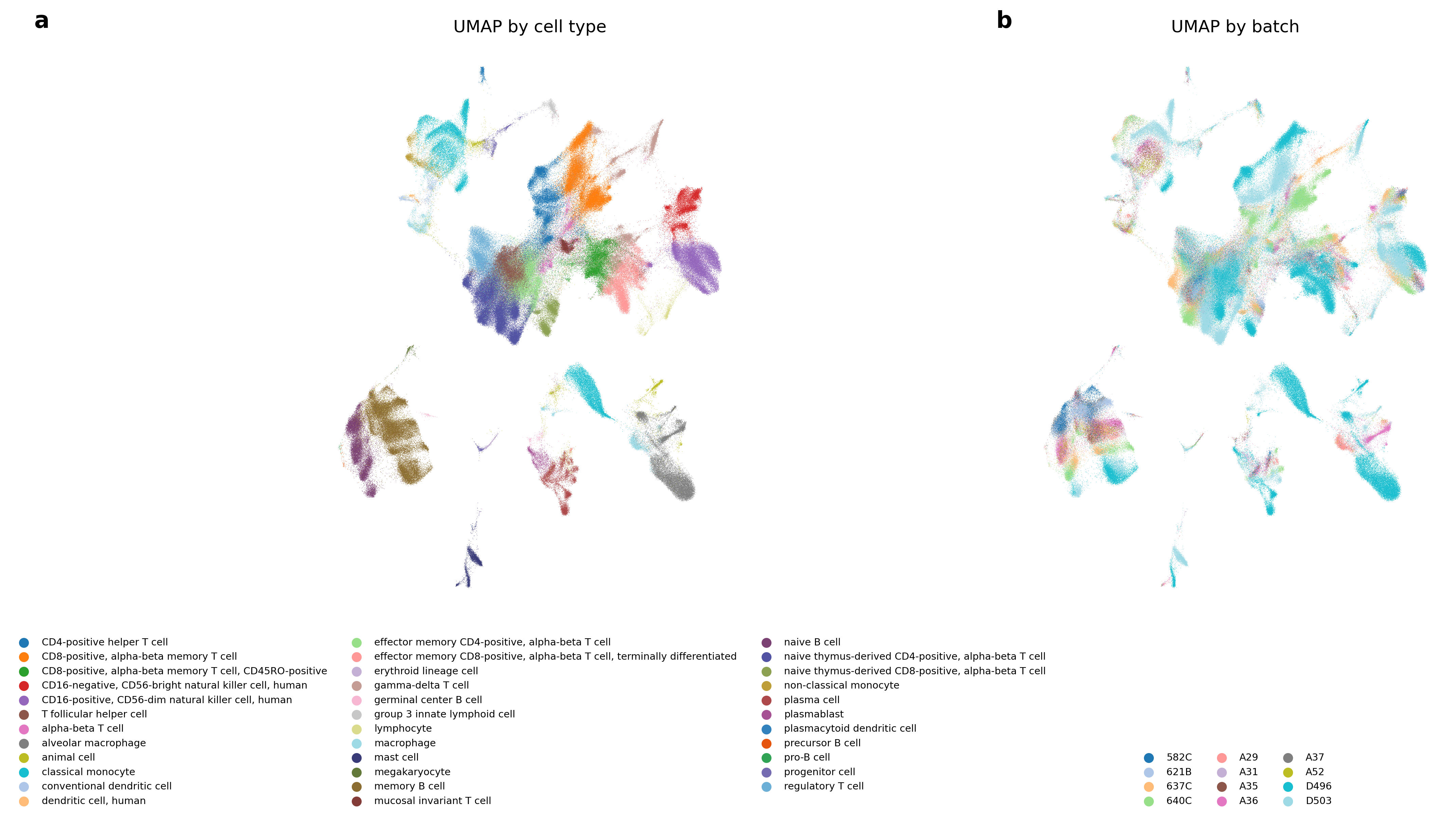}
  \caption{\textbf{Uniform Manifold Approximation and Projection~\cite{mcinnes2018} of \texttt{BBKNN (TS)} on the Immune Cell Atlas dataset.} \textbf{a,} The UMAP projection colored by cell type shows cell-type-specific clusters. \textbf{b,} The UMAP projection colored by data batch shows good batch mixing across the dataset.}
  \phantomsection
  \addcontentsline{toc}{subsection}{\protect\numberline{Fig. S2:}Uniform Manifold Approximation and Projection of \texttt{BBKNN (TS)} on the Immune Cell Atlas dataset.}
  \label{fig:bbknn_umap_immune_cell_atlas}
\end{figure}

\newpage

\begin{figure}[ht]
  \centering
  \includegraphics[width=1.0\textwidth]{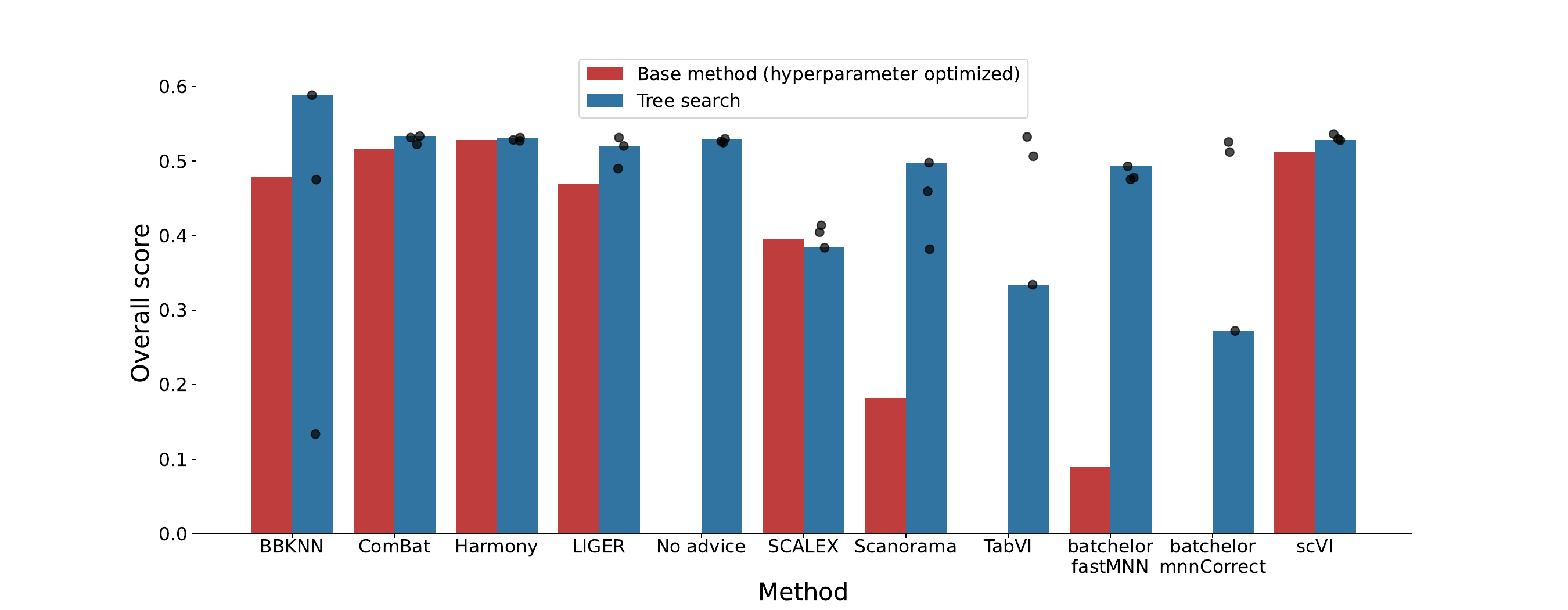}
  \caption{\textbf{Relative performance of base methods with optimized hyperparameters and ERA replicates.} Overall scores on the holdout OpenProblems datasets for all replicates of methods evaluated in Fig.\ 2. Hyperparameters for the base methods were optimized using the training dataset. For tree search implementations, three replicates of the full process were performed. Dots indicate the overall score of the replicate on the holdout OpenProblems datasets. The bar shows the performance of the replicate with highest performance in the validation dataset (identical values to those shown in Fig.\ 2). The \texttt{No advice} and \texttt{TabVI} methods have no base method code available. The \texttt{batchelor mnnCorrect} hyperparameter-optimized base method code failed to compute embeddings on every OpenProblems dataset owing to out-of-memory errors.}
  \phantomsection
  \addcontentsline{toc}{subsection}{\protect\numberline{Fig. S3:}Relative performance of base methods with optimized hyperparameters and ERA replicates.}
\label{fig:barplot_replicates_hparamopt}
\end{figure}

\newpage
\begin{figure}[ht]
  \centering
  \includegraphics[height=0.75\textheight]{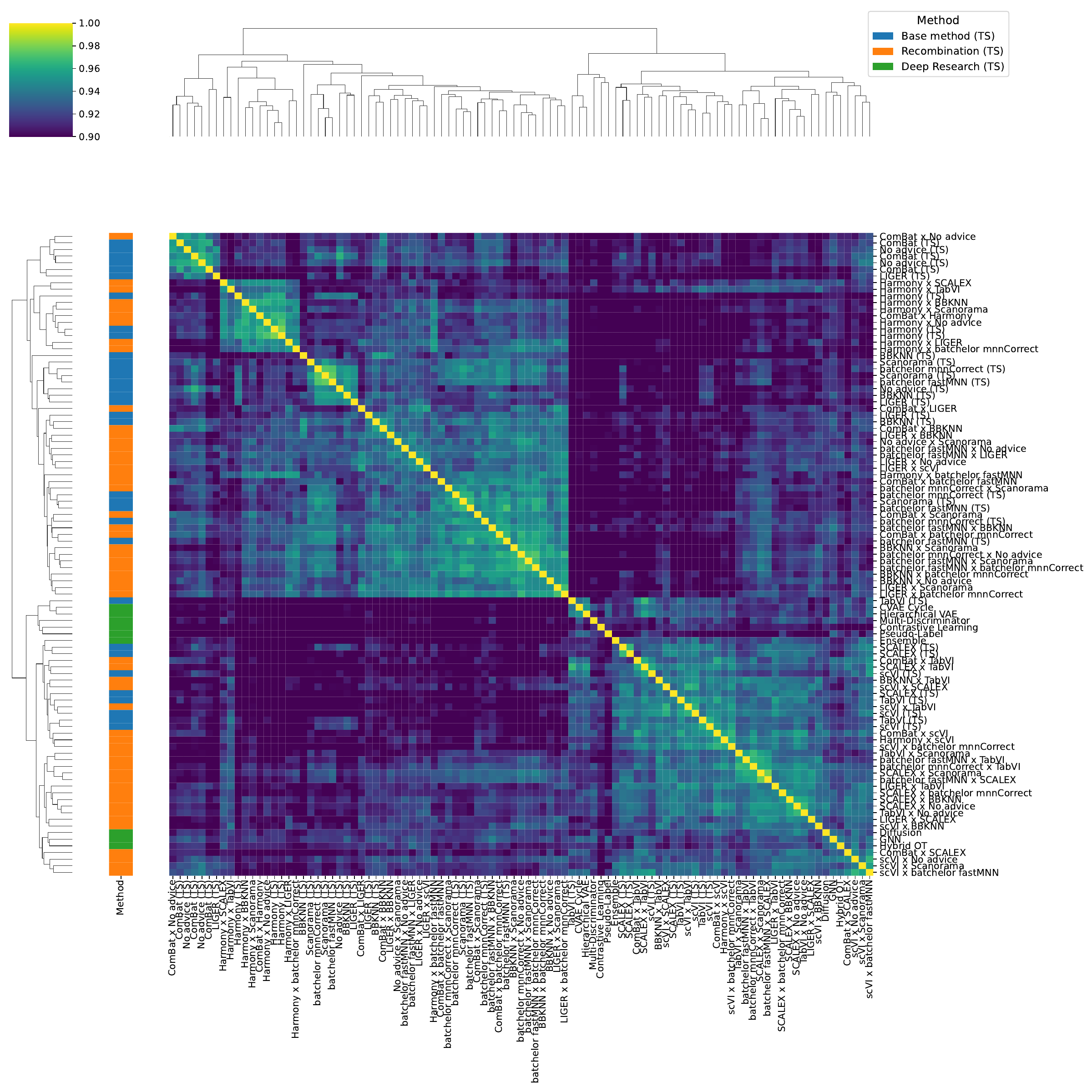}
  \caption{\textbf{Heatmap of text embedding cosine similarities among ERA-generated methods.} The similarity matrix was hierarchically clustered along rows and columns and reordered to group similar methods together. Three distinct color bars denote major method categories. The pairwise cosine similarities between tree search-generated solutions were greater than 0.85. For context, the lower bound of cosine similarity, established by averaging the similarities between GIFT-Eval’s methods (a completely different benchmark) and batch integration methods, was 0.74.}
  \phantomsection
  \addcontentsline{toc}{subsection}{\protect\numberline{Fig. S4:}Heatmap of text embedding cosine similarities among ERA-generated methods.}
  \label{fig:sc_sim_heatmap}
\end{figure}

\begin{figure}[ht]
  \centering
  \includegraphics[height=0.5\textheight]{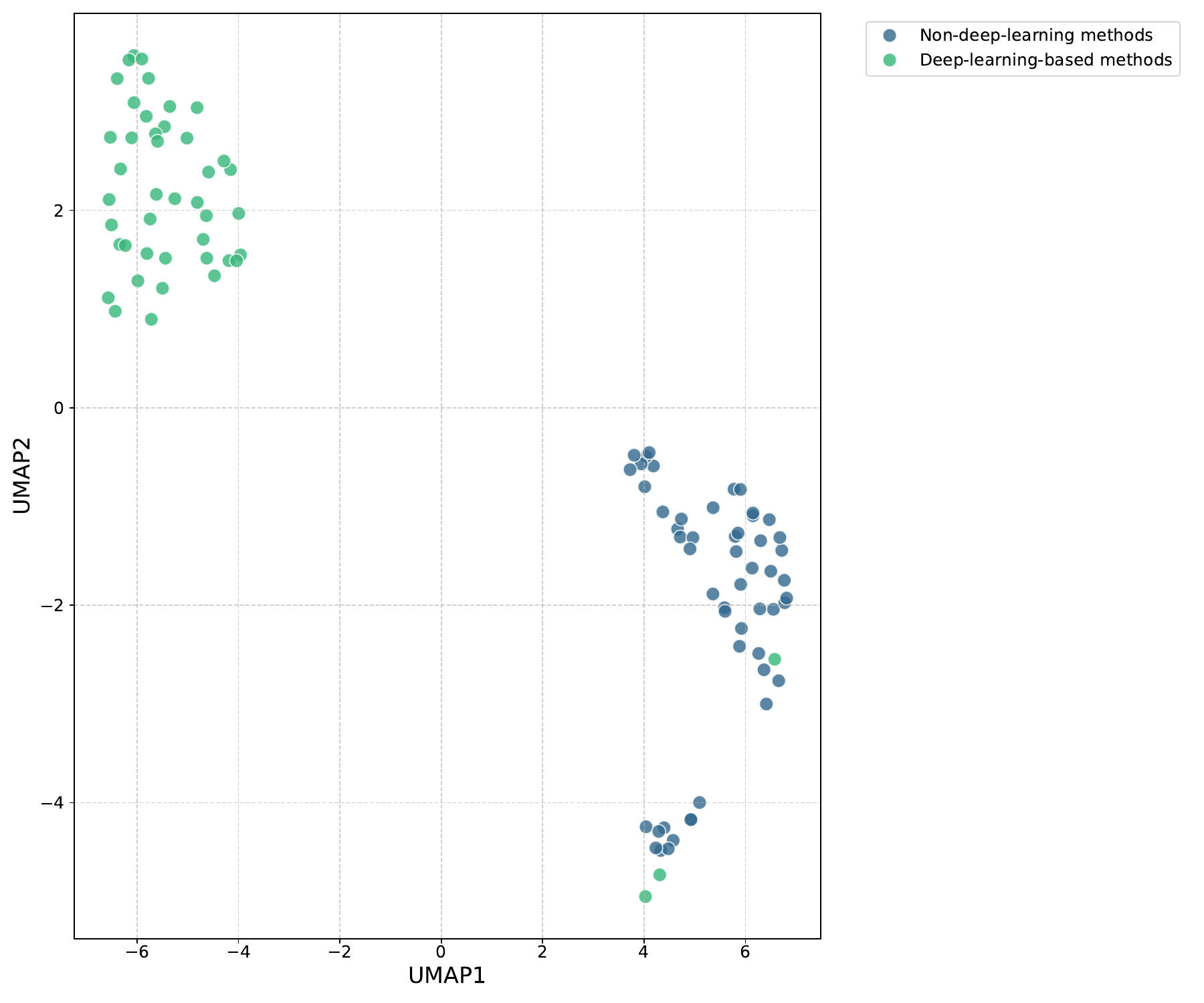}
  \caption{\textbf{UMAP of text embeddings representing ERA-generated methods for single-cell batch integration.} The UMAP shows two major clusters: non-deep-learning methods (blue) and deep-learning-based methods (green). We confirmed this clustering by using Gemini to classify the code associated with each method.}
  \phantomsection
  \addcontentsline{toc}{subsection}{\protect\numberline{Fig. S5:}UMAP of text embeddings representing ERA-generated methods for single-cell batch integration.}
  \label{fig:sc_sim_umap}
\end{figure}
\clearpage
\newpage

\begin{figure}[ht]
  \centering
  \includegraphics[width=1.0\textwidth]{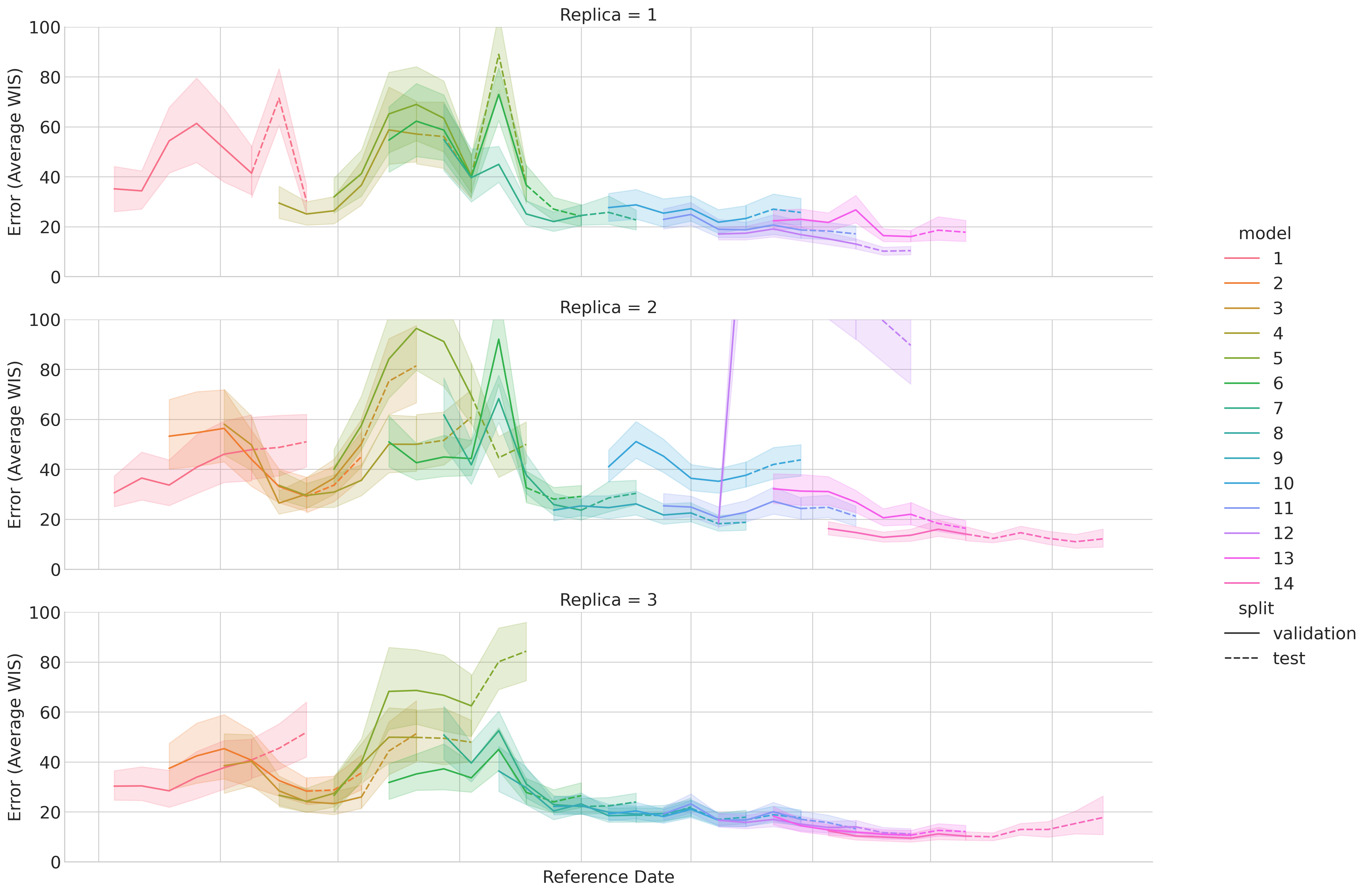}
  \caption{\textbf{Performance of retrospective COVID-19 hospitalization forecasts across all replicates.} Each panel displays the average WIS by reference date for individual replicates of our proposed models, for all rolling validation dates. Lower WIS values indicate superior forecasting accuracy and calibration. The consistent trends across replicates demonstrate the robustness and reproducibility of tree search's ability to generate high-performing probabilistic forecasts.}
  \phantomsection
  \addcontentsline{toc}{subsection}{\protect\numberline{Fig. S6:}Performance of retrospective COVID-19 hospitalization forecasts across all replicates.}
  \label{fig:covid_retro_all_rep}
\end{figure}
\clearpage
\newpage

\begin{figure}[htbp]
    \centering
    \begin{subfigure}{\textwidth}
        \centering
        \includegraphics[height=0.17\textheight, keepaspectratio]{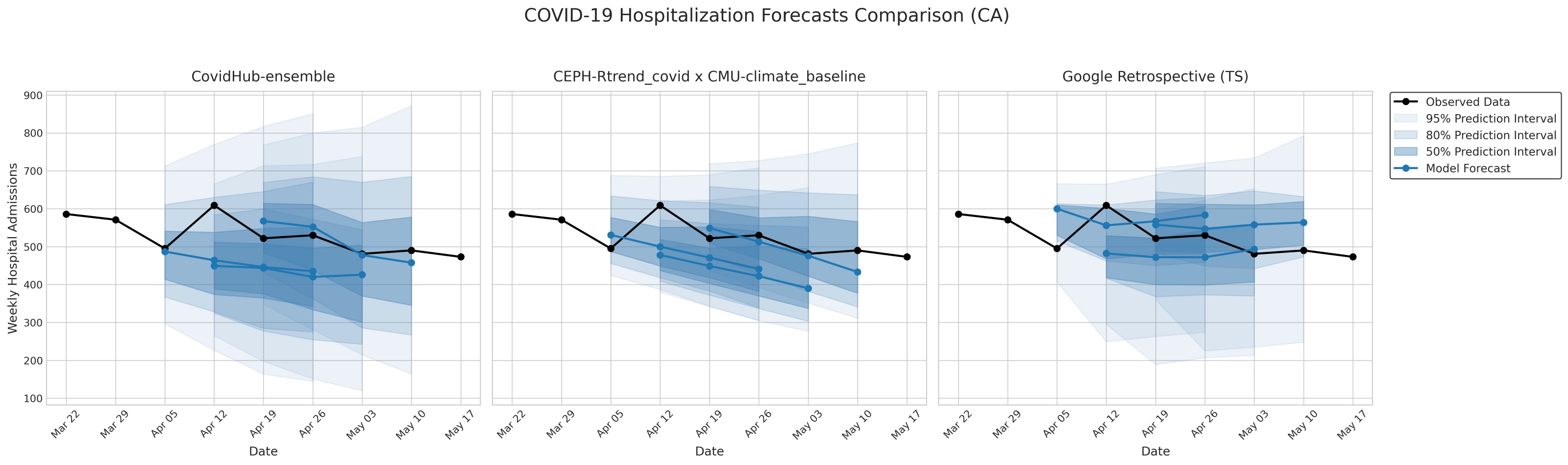}
        \vspace{2pt}
    \end{subfigure}
    
    \begin{subfigure}{\textwidth}
        \centering
        \includegraphics[height=0.17\textheight, keepaspectratio]{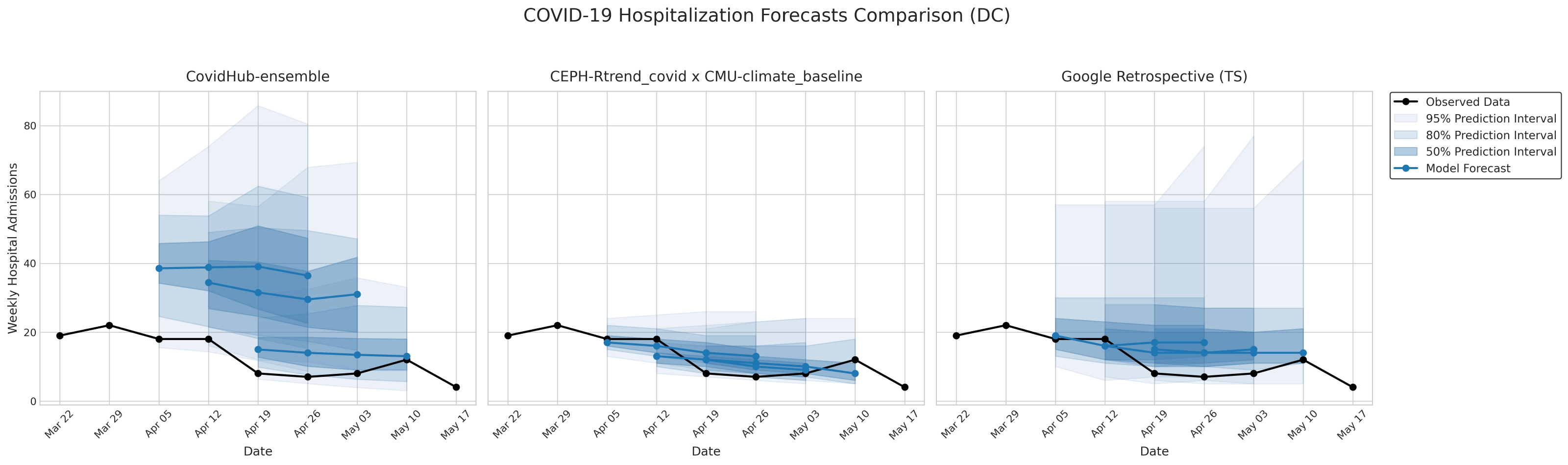}
        \vspace{2pt}
    \end{subfigure}
    
    \begin{subfigure}{\textwidth}
        \centering
        \includegraphics[height=0.17\textheight, keepaspectratio]{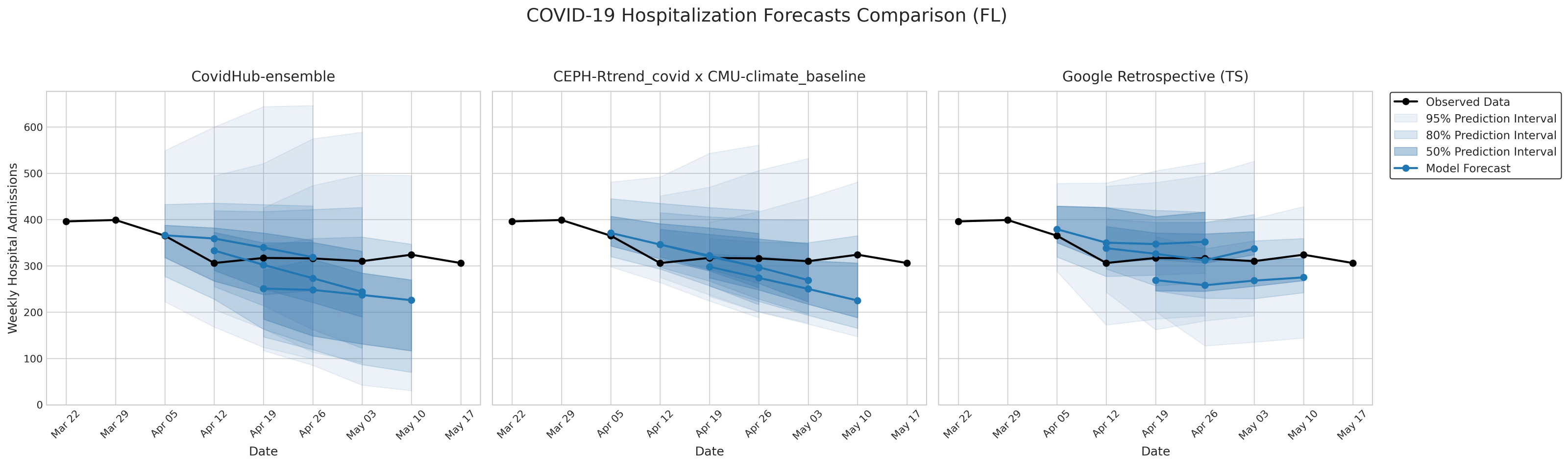}
        \vspace{2pt}
    \end{subfigure}
    
    \begin{subfigure}{\textwidth}
        \centering
        \includegraphics[height=0.17\textheight, keepaspectratio]{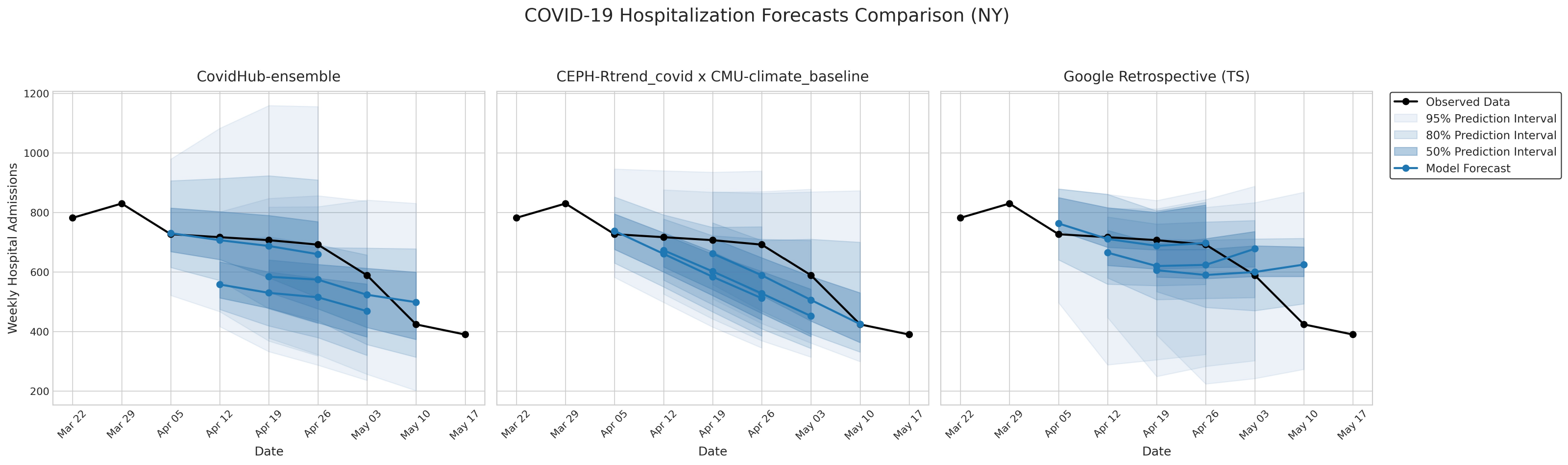}
        \vspace{2pt}
    \end{subfigure}
    
    \begin{subfigure}{\textwidth}
        \centering
        \includegraphics[height=0.17\textheight, keepaspectratio]{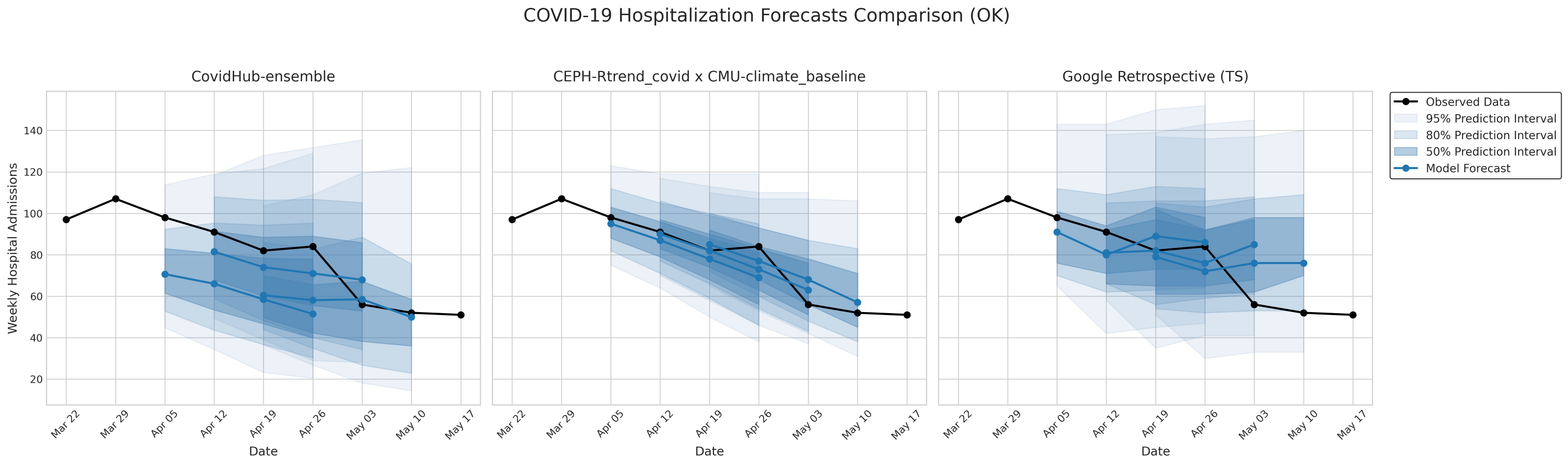}
    \end{subfigure}

    \caption{\textbf{Visual validation of COVID-19 hospitalization forecasts.} Comparison of ground truth (Observed Data) against the CovidHub-ensemble, the lowest WIS recombination solution (CEPH-Rtrend\_covid $\times$ CMU-climate\_baseline), and the Google Retrospective (TS) across five representative jurisdictions (CA, DC, FL, NY, OK). The ERA-developed models demonstrate a visible qualitative improvement in tracking non-linear dynamics and providing better-calibrated prediction intervals. This comparison confirms that the numerical improvements in WIS are driven by superior predictive accuracy across diverse temporal regimes, rather than statistical artifacts or a bias toward low-constant values.}
    \phantomsection
    \addcontentsline{toc}{subsection}{\protect\numberline{Fig. S7:}Visual validation of COVID-19 hospitalization forecasts.}
    \label{fig:supp_covid_curves}
\end{figure}

\newpage
\begin{figure}[p] 
    \centering
    \caption{\textbf{Full-season COVID-19 forecasts across all validation splits.} Comparison between the CovidHub-ensemble (left) and the Google Retrospective (TS) model (right). The overlapping colored ribbons represent successive 6-week validation splits throughout the 2024-2025 season.}
    \phantomsection
    \addcontentsline{toc}{subsection}{\protect\numberline{Fig. S8:}Full-season COVID-19 forecasts across all validation splits.}
    \label{fig:full_season_forecasts}
    \vspace{10pt}

    \begin{subfigure}{\textwidth}
        \centering
        \includegraphics[height=0.21\textheight, keepaspectratio]{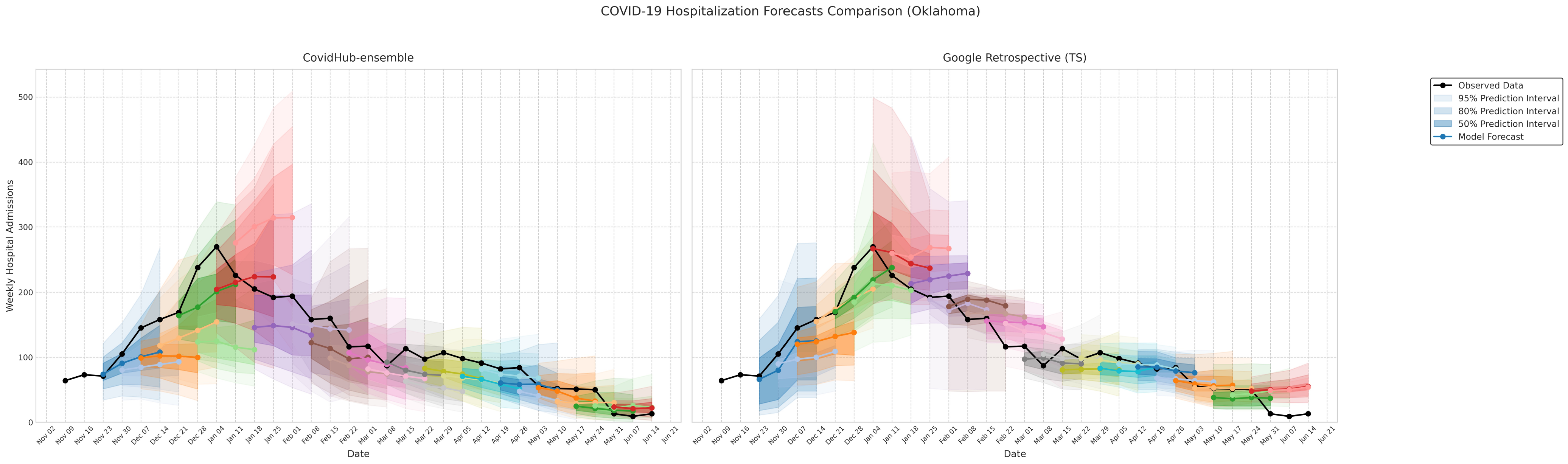}
        \subcaption{Oklahoma}
    \end{subfigure}
    \vfill

    \begin{subfigure}{\textwidth}
        \centering
        \includegraphics[height=0.21\textheight, keepaspectratio]{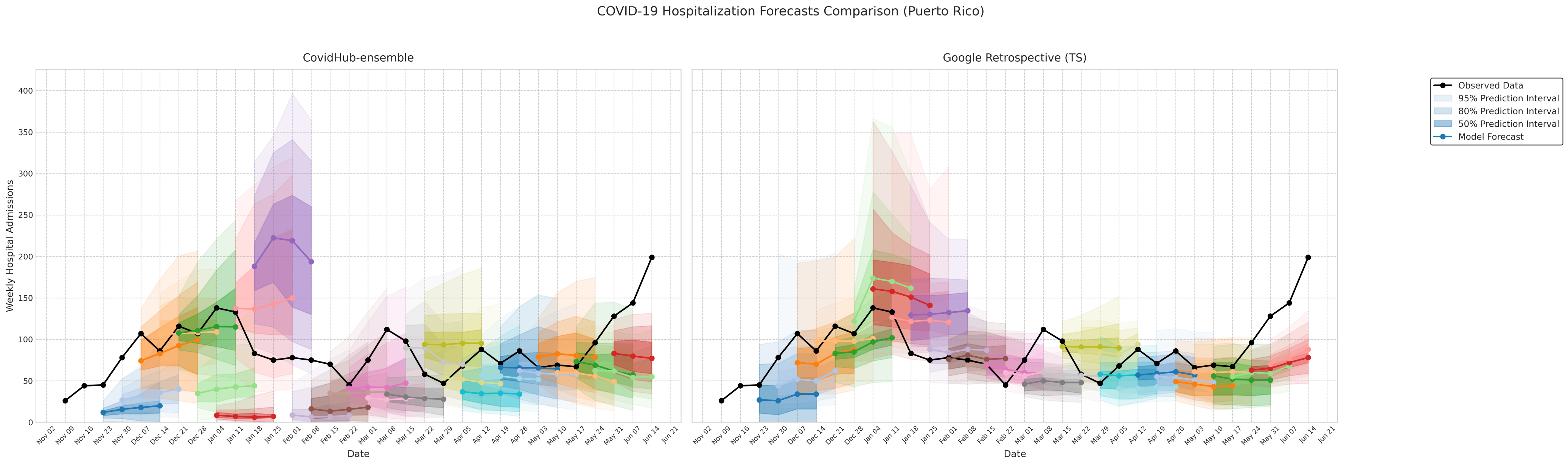}
        \subcaption{Puerto Rico}
    \end{subfigure}
    \vfill

    \begin{subfigure}{\textwidth}
        \centering
        \includegraphics[height=0.21\textheight, keepaspectratio]{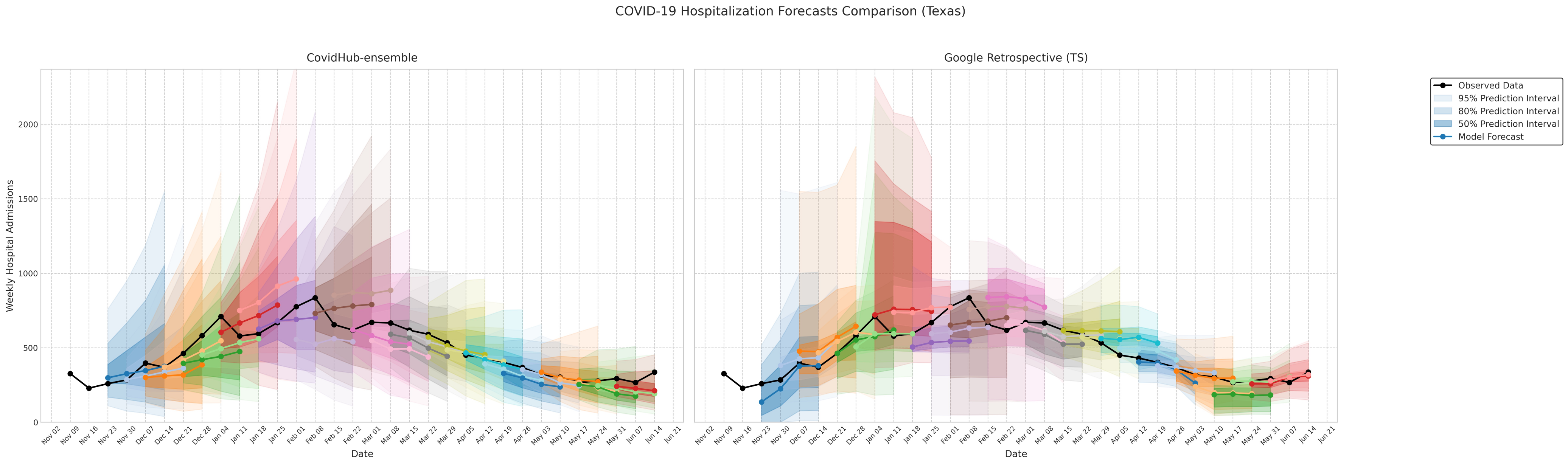}
        \subcaption{Texas}
    \end{subfigure}
    \vfill

    \begin{subfigure}{\textwidth}
        \centering
        \includegraphics[height=0.21\textheight, keepaspectratio]{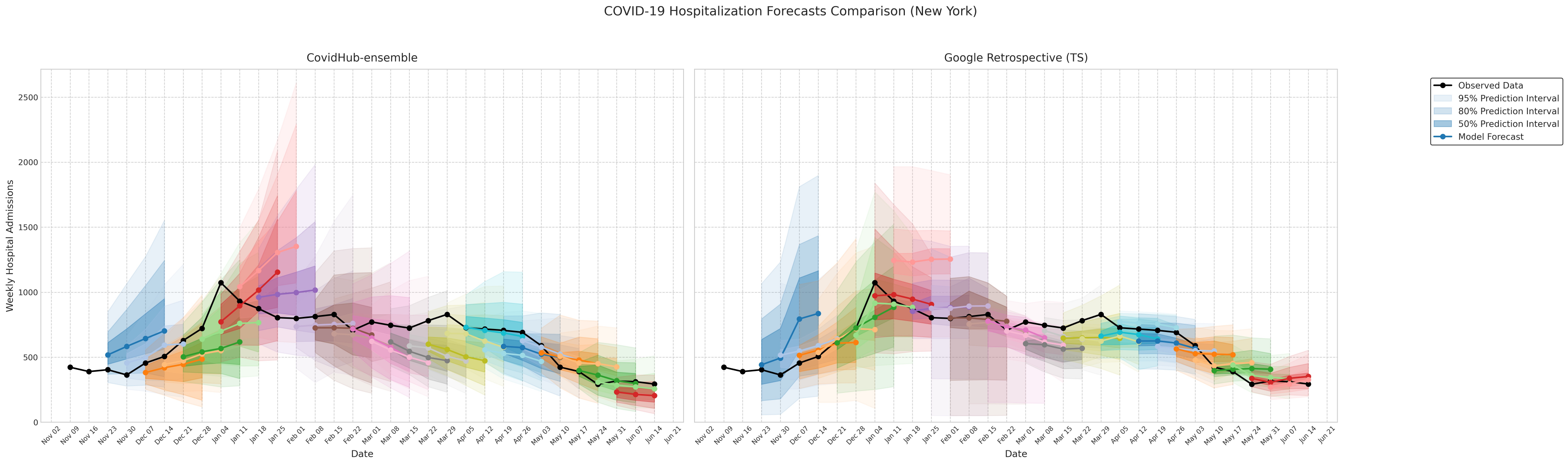}
        \subcaption{New York}
    \end{subfigure}
\end{figure}
\clearpage
\newpage

\begin{figure}[ht]
  \centering
  \includegraphics[height=0.75\textheight]{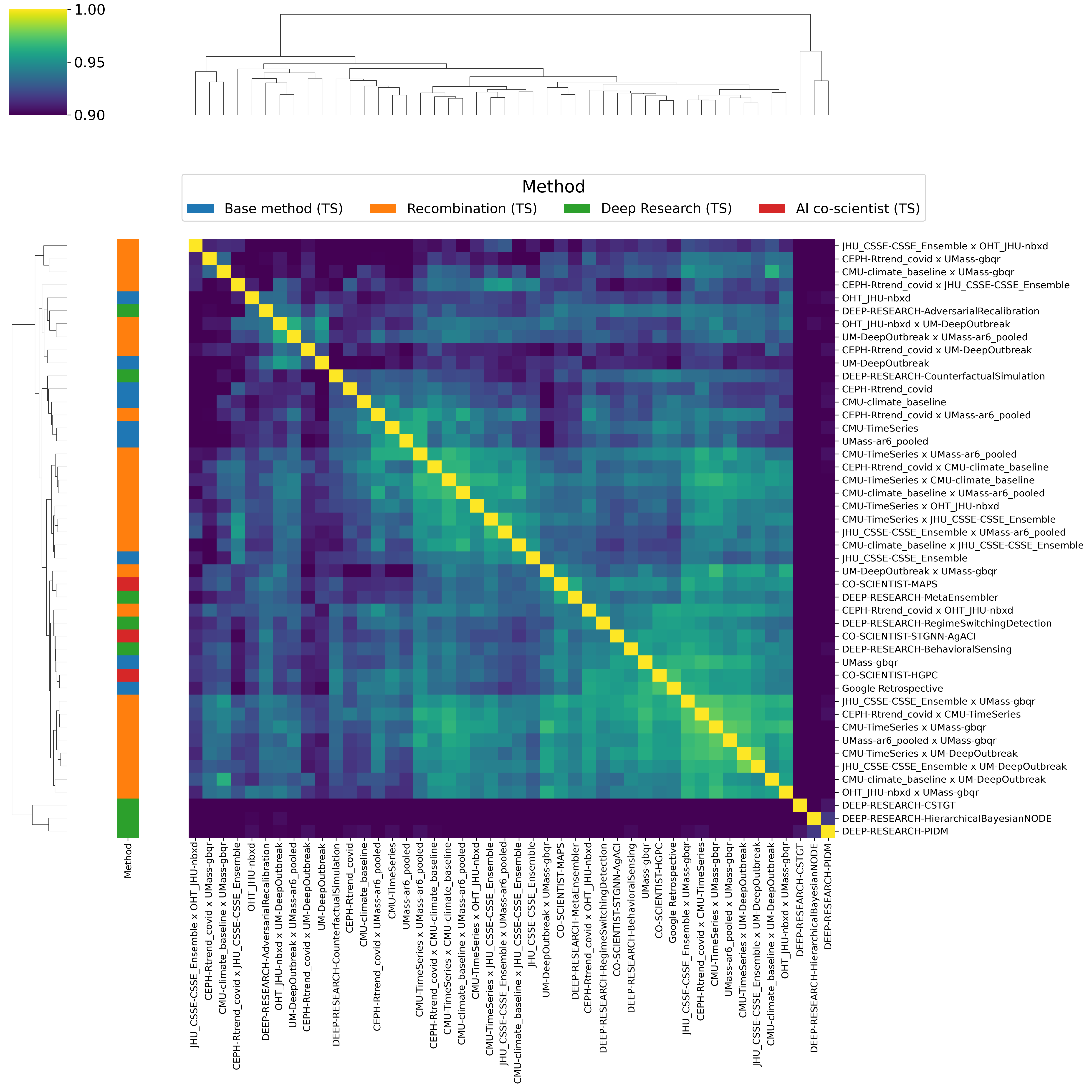}
  \caption{\textbf{Heatmap of conceptual similarities among COVID-19 forecasting generated codes for methods.} This figure displays the pairwise cosine similarities between text embeddings of all forecasting models generated by tree search for the COVID-19 prediction task. Text embeddings were produced using a Gemini model~\cite{lee2025gemini}. The similarity matrix was then hierarchically clustered and reordered to group conceptually related strategies. The color-coded sidebar categorizes each method by its origin illustrating the composition of the emergent conceptual clusters. The No Advice methods are from the Google Retrospective study.}
  \phantomsection
  \addcontentsline{toc}{subsection}{\protect\numberline{Fig. S9:}Heatmap of conceptual similarities among COVID-19 forecasting generated codes for methods.}
  \label{fig:covid_sim_heatmap}
\end{figure}

\begin{figure}[ht]
  \centering
  \includegraphics[width=1.0\textwidth]{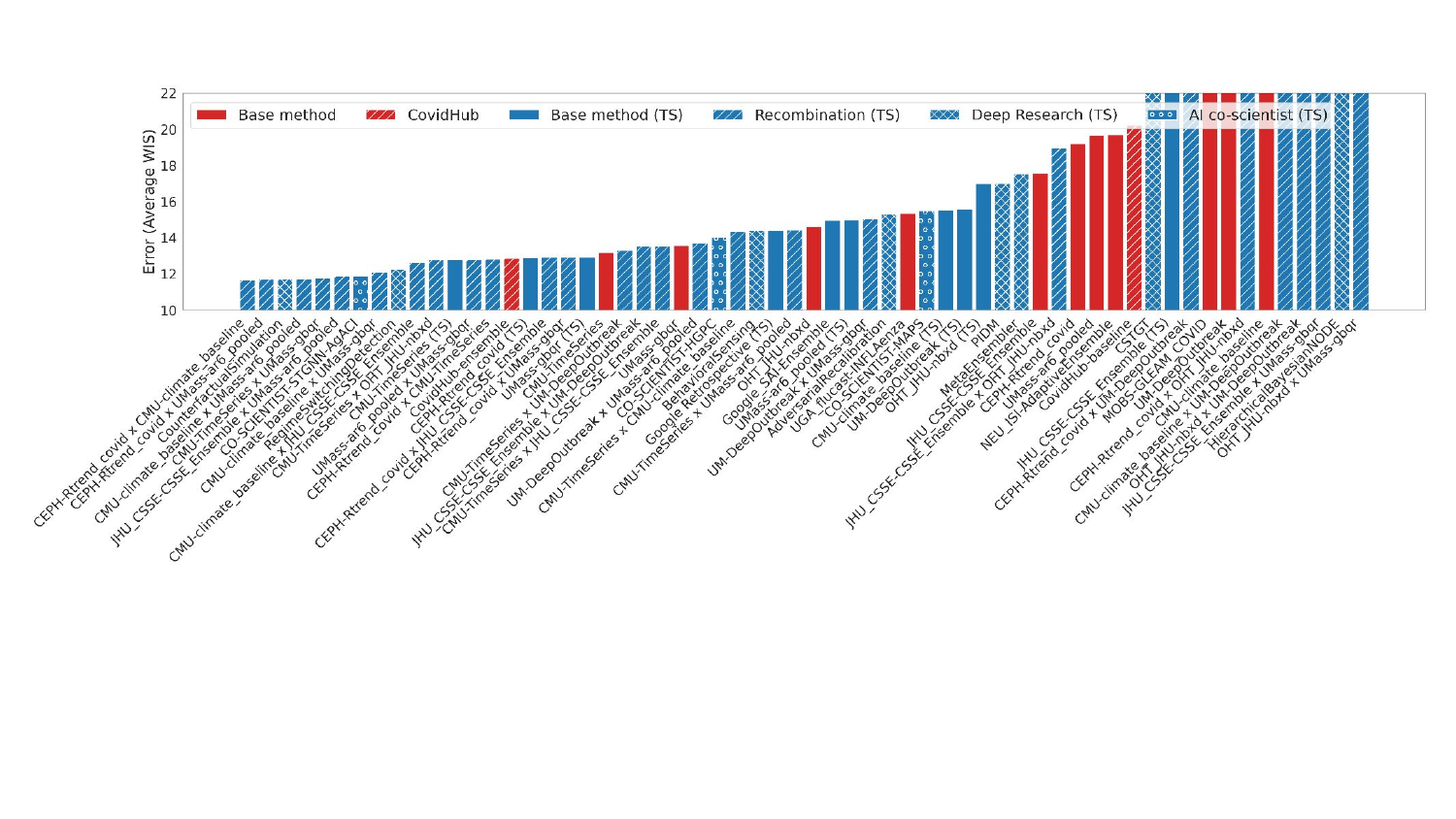}
  \caption{\textbf{Performance of recombination experiments for COVID-19 forecasting.} Average WIS across all evaluated models on the common held-out test set. This extended plot highlights the full distribution of search outcomes, including the 9 Tree Search generated models and 3 external models that performed worse than the CovidHub-baseline (indicated by red bars). Models are color-coded by strategy: base methods, CovidHub submissions, and Tree Search optimized variants (Base method, Recombination, Deep Research, and AI co-scientist).}
  \phantomsection
  \addcontentsline{toc}{subsection}{\protect\numberline{Fig. S10:}Performance of recombination experiments for COVID-19 forecasting.}
  \label{fig:covid_extended}
\end{figure}
\clearpage
\newpage

\begin{figure}[ht]
  \centering
  \includegraphics[width=1.0\textwidth]{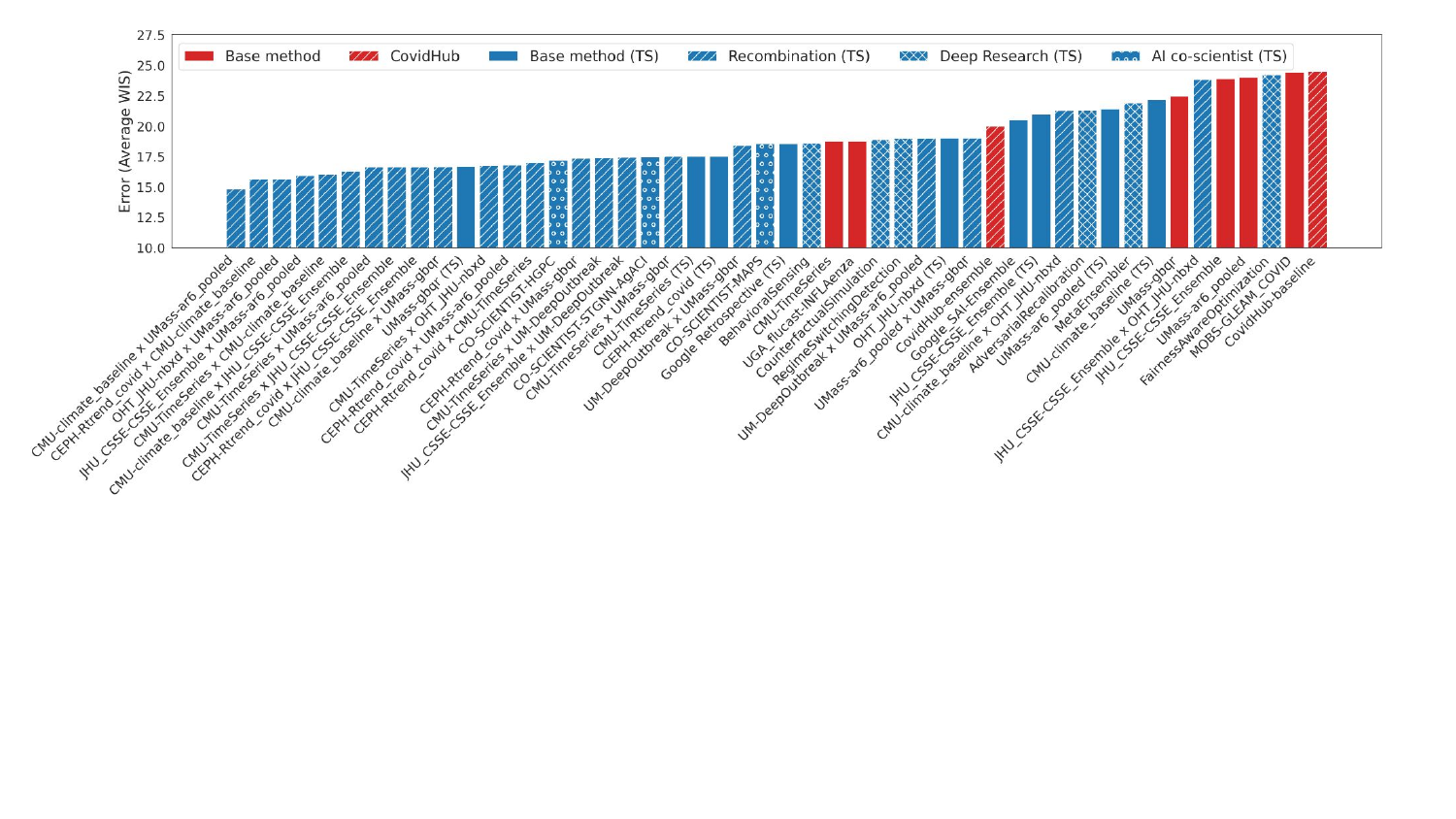}
  \caption{\textbf{Validation performance for COVID-19 forecasting.} Average WIS for models evaluated on the six-week rolling validation window covering the reference dates 2025-02-22 to 2025-03-29. Performance on the validation dates correlates with the held-out test set results; for instance, the best-performing test model (CEPH-Rtrend\_covid x CMU-climate\_baseline) was identified as the second-best performer during this validation phase.}
  \phantomsection
  \addcontentsline{toc}{subsection}{\protect\numberline{Fig. S11:}Validation performance for COVID-19 forecasting.}
  \label{fig:covid_validation}
\end{figure}
\clearpage
\newpage
\begin{figure}[ht]
  \centering
  \includegraphics[trim={0cm 0cm 13cm 0cm}, clip, width=1.0\textwidth]{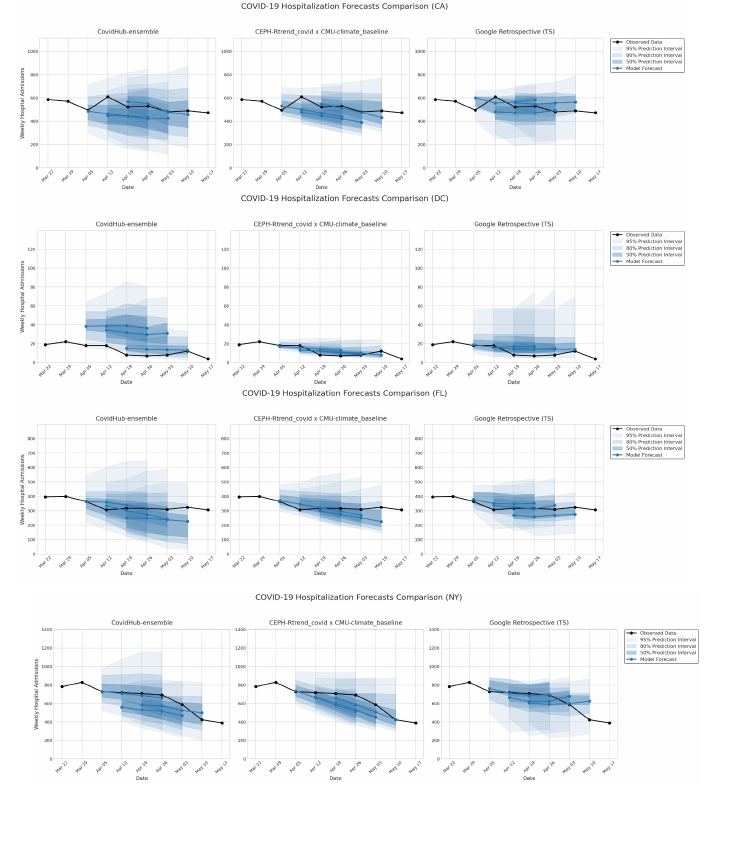}
  \caption{\textbf{Examples of COVID-19 forecasts.} Observed ground truth data (black line) plotted against predictive models across four major jurisdictions (CA, DC, FL, and NY) for the reference dates 2025-04-05, 2025-04-12, and 2025-04-19. These dates are used as the held out test set in the analysis in the paper. Columns represent the CovidHub-ensemble (left), the best-performing tree search (TS) discovered recombination model (CEPH-Rtrend\_covid x CMU-climate\_baseline, middle), and the Google Retrospective TS model (right). The average WIS for this period are 11.63 for the top recombination model, 12.85 for the CovidHub-ensemble, and 14.39 for the Google Retrospective model. The TS methods generate significantly narrower confidence bands—indicating higher predictive precision—while consistently containing the ground truth data.}
  \phantomsection
  \addcontentsline{toc}{subsection}{\protect\numberline{Fig. S12:}Examples of COVID-19 forecasts.}
  \label{fig:prediction_covid}
\end{figure}

\clearpage
\newpage
\begin{figure}[ht]
    \includegraphics[width=0.75\linewidth]{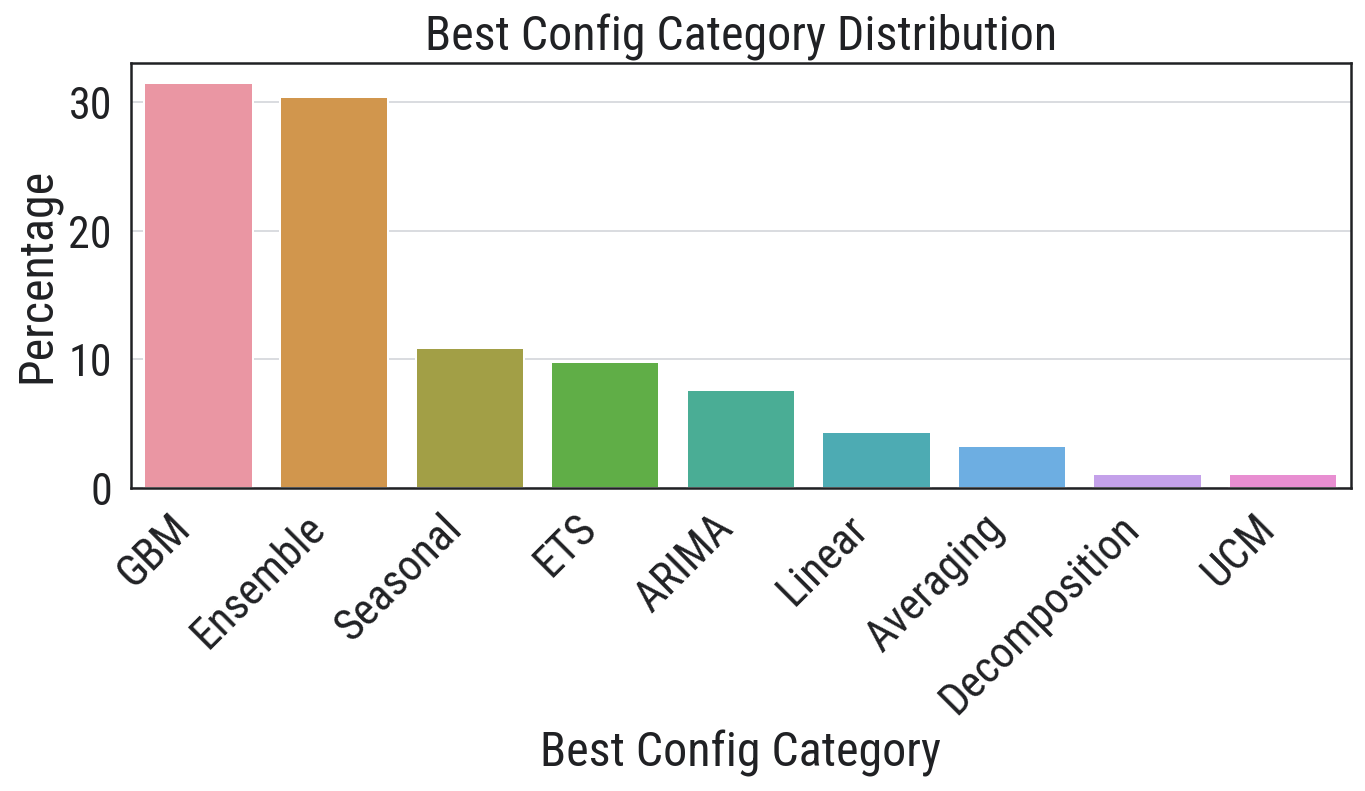}
  \caption{\textbf{Categories of solutions on the GIFT-Eval benchmark on the \texttt{per-dataset solution (v1)}.} We prompted an LLM (Gemini 2.5 Pro) to categorize the code from each of the solutions into a class of methods. The figure shows the percentage of the best codes for each of the 92 competitions in the specified categories:  Gradient Boosted Method (GBM); Ensemble; Seasonal; Error, Trend and Seasonality (ETS); Arima \cite{ho1998use}; Linear; Averaging; Decomposition and Unobserved components model (UCM).}
  \phantomsection
  \addcontentsline{toc}{subsection}{\protect\numberline{Fig. S13:}Categories of solutions on the GIFT-Eval benchmark on the \texttt{per-dataset solution (v1)}.}
  \label{fig:gift-category}
\end{figure}

\clearpage

\begin{figure}
    \centering
    \includegraphics[width=0.8\linewidth]{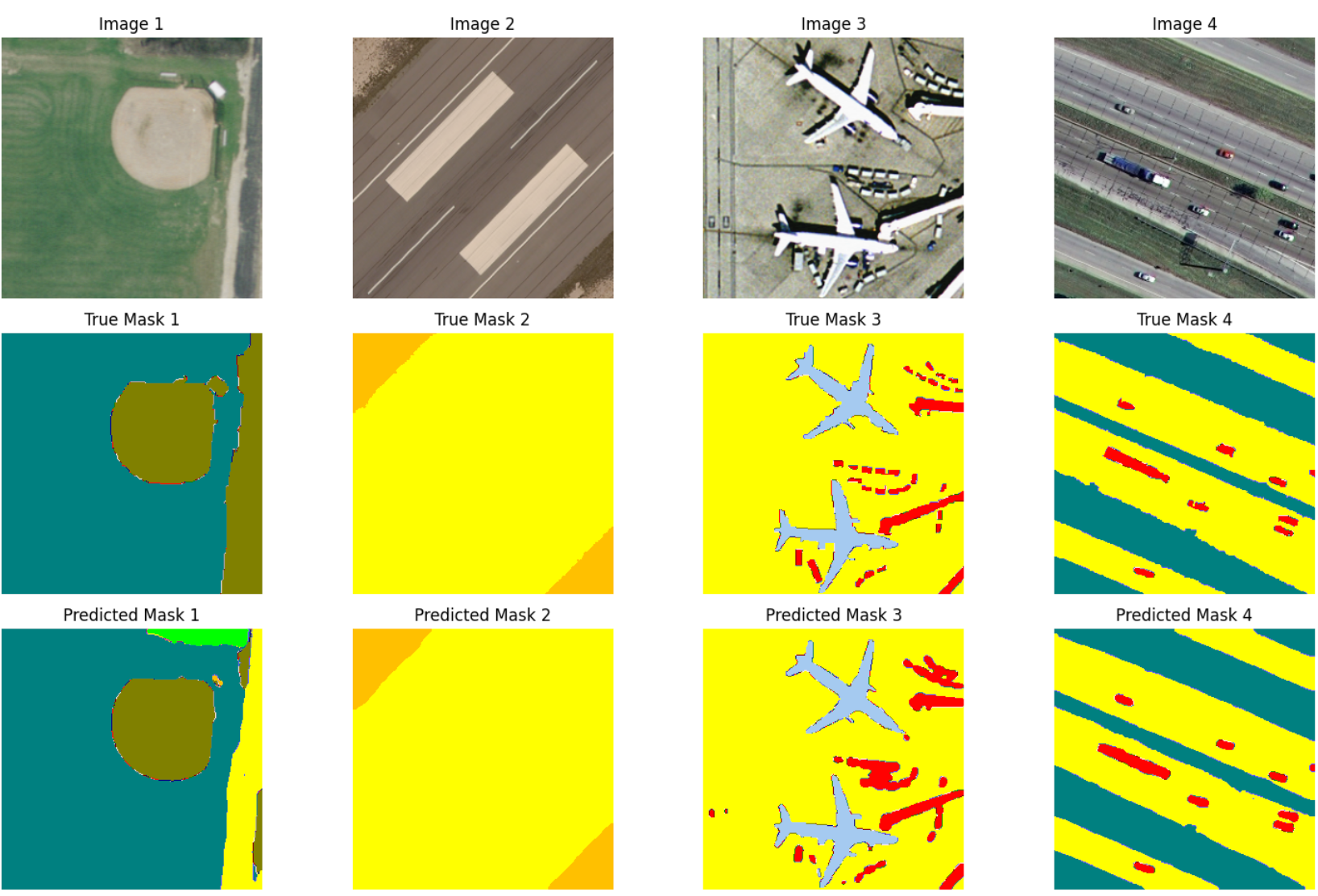}
    \caption{\textbf{Example output segmenting DLRSD image pixels from \ourmethod Solution 1 (U-Net++).}}
    \phantomsection
    \addcontentsline{toc}{subsection}{\protect\numberline{Fig. S14:}Example output segmenting DLRSD image pixels from \ourmethod Solution 1 (U-Net++).}
    \label{fig:dlrsd_example}
\end{figure}

\begin{figure}[h!]
    \centering 
    
    \begin{subfigure}[b]{0.9\textwidth}
        \centering
        \includegraphics[width=\linewidth]{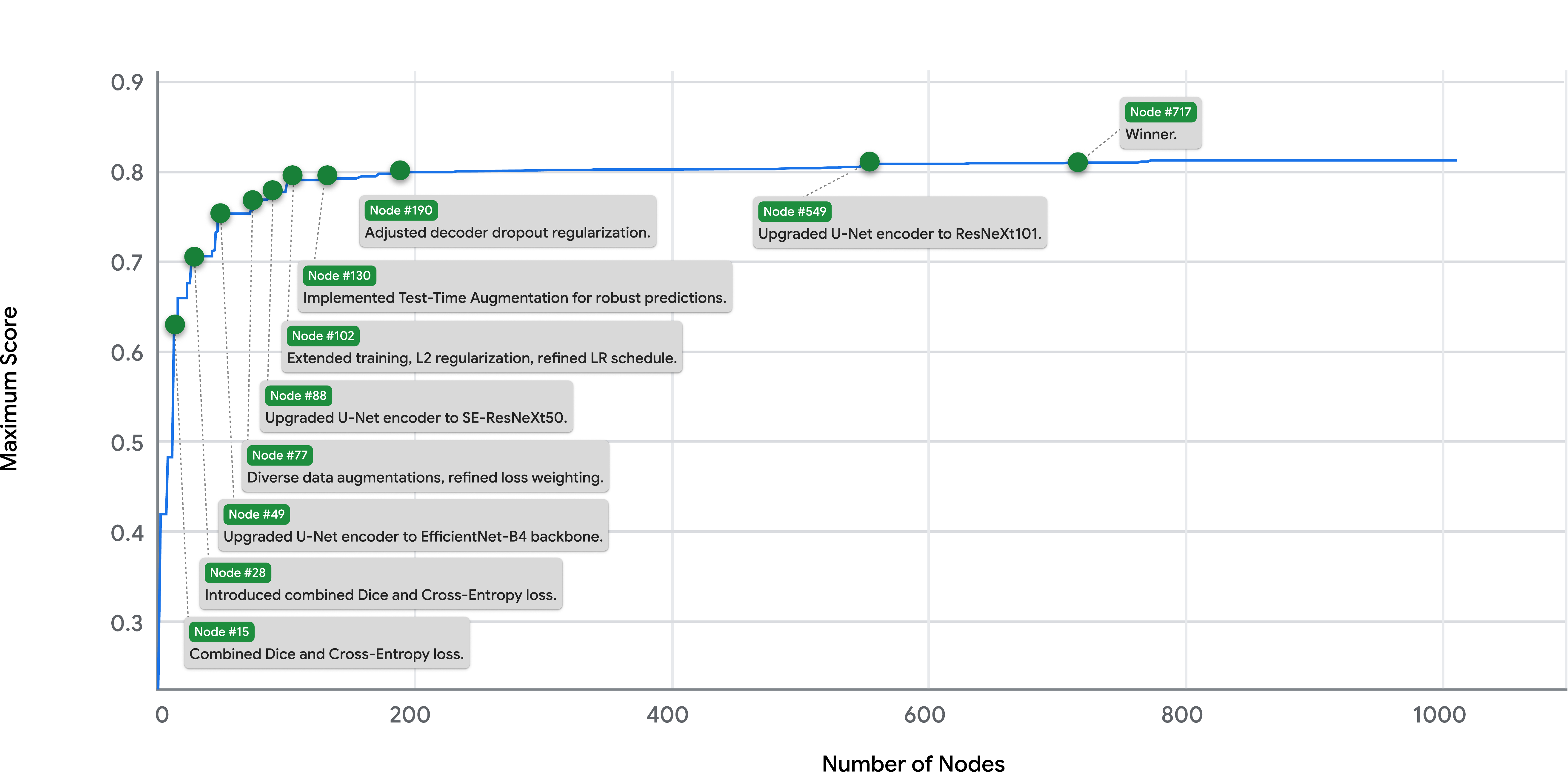}
    \end{subfigure}
    \hfill 
    \begin{subfigure}[b]{0.9\textwidth}
        \centering

        \includegraphics[width=0.8\linewidth]{Figures/tree_visualization_pretty_geospatial.pdf}
    \end{subfigure}
    
    \caption{\textbf{Breakthrough plot and solution tree for the geospatial segmentation task.} {\sl Top Figure} Breakthrough plot for the U-Net Geospatial DLRSD solution (solution 3), showing the evolution of the maximum score as a function of the number of nodes. The green dots label places where the score abruptly increases due to an improvement in the code, and the label describes the change in the code that resulted in the score increase. {\sl Bottom Figure} Structure of the tree for this same search. The color range consists of orange (lower scores) to green (higher scores) with the highest score denoted by a diamond node.}
    \phantomsection
    \addcontentsline{toc}{subsection}{\protect\numberline{Fig. S15:}Breakthrough plot for the geospatial segmentation task.}
    \label{fig:tree_breakthrough_geospatial}
\end{figure}
\newpage

\begin{figure}[bh]
    \centering
\includegraphics[width=1.0\linewidth]{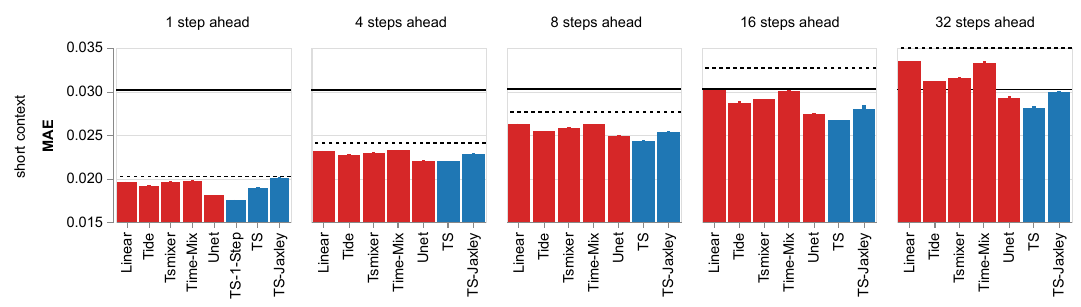}
    \caption{\textbf{Comparison of solutions to time-series and video forecasting methods across conditions on ZAPBench.} Solutions are evaluated using average mean absolute error (MAE) across conditions (lower is better).
    For \ourmethod, we report the performance of three different solutions (blue), and compare them against baselines (red). Alongside our best general solution (TS), we include results from two specialized runs: a tree search that was optimized for 1-step ahead forecasting as well as a solution prompted to use Jaxley, a differentiable biophysical neuron simulator.
    The dotted and solid lines represent the mean and stimulus baselines, respectively. To account for variability due to random number generator seeding, each method was run three times. We report the mean, with error bars indicating 95\% confidence intervals.}
    \phantomsection
    \addcontentsline{toc}{subsection}{\protect\numberline{Fig. S16:}Comparison of solutions to time-series and video forecasting methods across conditions on ZAPBench.}
    \label{fig:zapbench}
\end{figure}

\begin{figure}[h!]
    \centering 
    
    \begin{subfigure}[b]{0.9\textwidth}
        \centering
        \includegraphics[width=\linewidth]{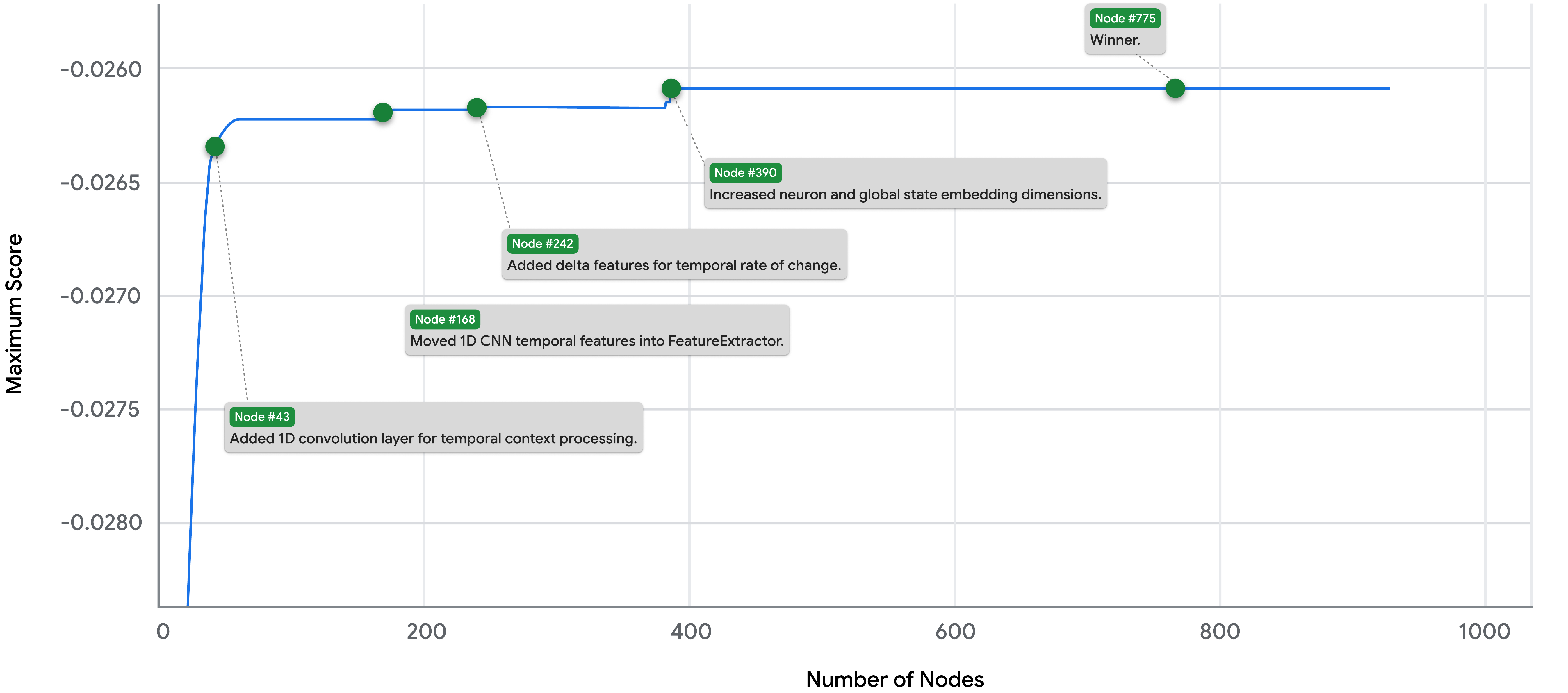}
    \end{subfigure}
    \hfill 
    \begin{subfigure}[b]{0.9\textwidth}
 
        \includegraphics[
    width=0.8\linewidth
]{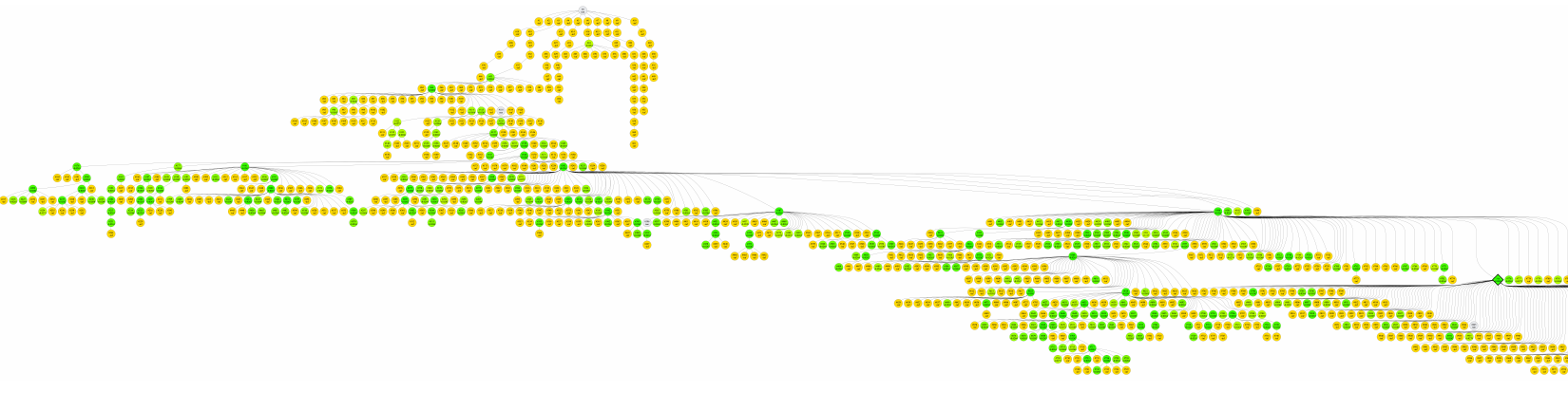}
    \end{subfigure}
 
    \caption{\textbf{Breakthrough plot and solution tree for the ZAPBench task.} {\sl Top Figure} Breakthrough plot for the ZAPBench task, showing the evolution of the maximum score as a function of the number of nodes. The green dots label places where the score abruptly increases due to an improvement in the code, and the label describes the change in the code that resulted in the score increase. {\sl Bottom Figure} Structure of the tree for this same search. The color range consists of orange (lower scores) to green (higher scores) with the highest score denoted by a diamond node.}
    \phantomsection
    \addcontentsline{toc}{subsection}{\protect\numberline{Fig. S17:}Breakthrough plot and solution tree for the ZAPBench task.}
    \label{fig:tree_breakthrough_zapbench}
\end{figure}

\clearpage
\newpage

\begin{figure}
    \begin{center}
        \begin{tabular}{ll|ll}
            train set & & test set & \\
            \hline
445.001 & $ \int\limits_{0}^{\infty} \sin{\left(x^{2} \right)}\, dx $ &
446.021 & $ \int\limits_{0}^{\infty} \left(\sin^{4}{\left(a x^{2} \right)} - \sin^{4}{\left(b x^{2} \right)}\right)\, dx $ \\
445.017 & $ \int\limits_{0}^{\infty} \sin{\left(a x^{2} \right)} \cos{\left(2 b x \right)}\, dx $ &
446.045 & $ \int\limits_{0}^{\infty} x \cos{\left(a x^{2} \right)} \cos{\left(2 b x \right)}\, dx $ \\
447.012 & $ \int\limits_{0}^{\infty} \sin{\left(a x^{2} + \frac{b^{2}}{a} \right)} \cos{\left(2 b x \right)}\, dx $ &
449.013 & $ \int\limits_{0}^{\infty} x^{\mu - 1} \sin{\left(a x \right)} \cos{\left(b x \right)}\, dx $ \\
458.031 & $ \int\limits_{0}^{\infty} \left(\frac{\gamma + x}{\beta^{2} + \left(\gamma + x\right)^{2}} - \frac{\gamma - x}{\beta^{2} + \left(\gamma - x\right)^{2}}\right) \sin{\left(a x \right)}\, dx $ &
465.002 & $ \int\limits_{0}^{\infty} \frac{\left(3 - 4 \sin^{2}{\left(a x \right)}\right) \sin^{2}{\left(a x \right)}}{x}\, dx $ \\
462.034 & $ \int\limits_{0}^{\infty} \frac{x \sin{\left(a x \right)} \cos{\left(b x \right)}}{c^{2} + x^{2}}\, dx $ &
465.013 & $ \int\limits_{0}^{\infty} \frac{\sin^{2 m + 1}{\left(x \right)} \sin{\left(x \left(6 m + 3\right) \right)}}{a^{2} + x^{2}}\, dx $ \\
477.049 & $ \int\limits_{0}^{\infty} \frac{x \sin{\left(a x \right)} + \cos{\left(a x \right)}}{x^{2} + 1}\, dx $ &
467.025 & $ \int\limits_{0}^{\infty} \frac{\sin{\left(x \right)} \cos{\left(x \right)}}{x \sqrt{\sin^{2}{\left(x \right)} + 1}}\, dx $ \\
478.036 & $ \int\limits_{0}^{\infty} \frac{\left(\cos{\left(a \right)} - \cos{\left(a n x \right)}\right) \sin{\left(m x \right)}}{x}\, dx $ &
478.031 & $ \int\limits_{0}^{\infty} \sin{\left(a x^{p} \right)}\, dx $ \\
487.011 & $ \int\limits_{0}^{\infty} \frac{1}{x} \frac{\sin{\left(x \right)}}{\left(a^{2} \cos^{2}{\left(x \right)} + b^{2} \sin^{2}{\left(x \right)}\right)^{2}}\, dx $ &
478.050 & $ \int\limits_{u}^{\infty} \frac{\cos{\left(a x \right)}}{\sqrt{- u + x}}\, dx $ \\
487.026 & $ \int\limits_{0}^{\infty} \frac{1}{x} \frac{\sin{\left(x \right)} \cos^{2}{\left(x \right)}}{\left(a^{2} \cos^{2}{\left(x \right)} + b^{2} \sin^{2}{\left(x \right)}\right)^{2}}\, dx $ &
484.059 & $ \int\limits_{0}^{\infty} \left(\sin{\left(a - x^{2} \right)} + \cos{\left(a - x^{2} \right)}\right)\, dx $ \\
488.014 & $ \int\limits_{0}^{\infty} \frac{1}{x} \frac{\sin^{3}{\left(x \right)} \cos{\left(x \right)}}{\left(a^{2} \cos^{2}{\left(2 x \right)} + b^{2} \sin^{2}{\left(2 x \right)}\right)^{4}}\, dx $ &
487.068 & $ \int\limits_{0}^{\infty} \frac{\cos{\left(x \right)} \cos{\left(a \cos{\left(x \right)} \right)} \cos{\left(2 n x \right)} \sinh{\left(a \sin{\left(x \right)} \right)}}{x}\, dx $ \\
491.004 & $ \int\limits_{0}^{\infty} \frac{\cos^{2 m}{\left(x \right)}}{a^{2} + x^{2}}\, dx $ &
494.006 & $ \int\limits_{0}^{\infty} x \sin{\left(2 b x \right)} \cos{\left(a x^{2} \right)}\, dx $ \\
491.006 & $ \int\limits_{0}^{\infty} \frac{\cos^{2 m + 1}{\left(x \right)}}{a^{2} + x^{2}}\, dx $ &
496.037 & $ \int\limits_{0}^{\infty} \frac{\sin^{3}{\left(x \right)}}{\left(a^{2} \cos^{2}{\left(x \right)} + b^{2} \sin^{2}{\left(x \right)}\right)^{3}} \frac{1}{x} \, dx $ \\
491.014 & $ \int\limits_{0}^{\infty} \frac{x \sin{\left(2 a x \right)} \cos^{2}{\left(b x \right)}}{\beta^{2} + x^{2}}\, dx $ &
504.025 & $ \int\limits_{0}^{\infty} \frac{\sin{\left(a x^{p} \right)}}{x}\, dx $ \\
493.056 & $ \int\limits_{0}^{\infty} \frac{\sin{\left(2 a x \right)} \cos^{2}{\left(b x \right)}}{x}\, dx $ &
504.061 & $ \int\limits_{0}^{\infty} \frac{\sin^3{\left(x \right)} \cos{\left(x \right)}}{x \sqrt{\sin^{2}{\left(2 x \right)} + 1}}\, dx $ \\
495.029 & $ \int\limits_{0}^{\infty} \frac{\sin^{3}{\left(a x \right)} \sin^{2}{\left(b x \right)}}{x}\, dx $ &
505.006 & $ \int\limits_{0}^{\infty} \frac{\sqrt{- b + \sqrt{b^{2} + x^{2}}} \sin{\left(a x \right)}}{\sqrt{b^{2} + x^{2}}}\, dx $ \\
504.057 & $ \int\limits_{0}^{\infty} \frac{\sin^3{\left(x \right)} \cos{\left(x \right)}}{x \sqrt{\cos^{2}{\left(2 x \right)} + 1}}\, dx $ &
505.008 & $ \int\limits_{0}^{\infty} \frac{\sin{\left(x \right)}}{x \left(a^{2} \sin^{2}{\left(x \right)} + b^{2} \cos^{2}{\left(x \right)}\right)}\, dx $ \\
512.029 & $ \int\limits_{0}^{\infty} \frac{\cos{\left(b x \right)} \cos{\left(p \sqrt{a^{2} + x^{2}} \right)}}{c^{2} + x^{2}}\, dx $ &
505.023 & $ \int\limits_{0}^{\infty} \frac{\left(\cos{\left(a \right)} - \cos{\left(a n x \right)}\right) \sin{\left(m x \right)}}{x}\, dx $ \\
512.037 & $ \int\limits_{0}^{\infty} \frac{\cos{\left(b x \right)} \cos{\left(p \sqrt{a^{2} + x^{2}} \right)}}{a^{2} + x^{2}}\, dx $ &
513.033 & $ \int\limits_{0}^{\infty} \frac{\sin^{3}{\left(a x \right)} \cos{\left(3 b x \right)}}{x^{2}}\, dx $ \\
550.003 & $ \int\limits_{0}^{\infty} \frac{\sin{\left(a x \right)} \coth{\left(\frac{\pi x}{2} \right)}}{x^{2} + 1}\, dx $ &
551.027 & $ \int\limits_{0}^{\infty} \frac{\sin^{3}{\left(a^{2} x^{2} \right)}}{x^{2}}\, dx $
    \end{tabular}
    \end{center}
    \caption{\textbf{The dataset of 38 definite integrals with oscillatory integrands
    on semi-infinite domains~\cite{gandr}, none of which were solved correctly by \texttt{scipy.integrate.quad()}}. Parameters
    like $a$, $b$, $c$ were chosen randomly between 0 and 5 with exponents constrained to be integers.}
    \phantomsection
    \addcontentsline{toc}{subsection}{\protect\numberline{Fig. S18:}The dataset of 38 definite integrals with oscillatory integrands on semi-infinite domains.}
    \label{fig:integrals-datasets}
\end{figure}

\begin{figure}[h!]
    \centering 
    
    \begin{subfigure}[b]{0.9\textwidth}
        \centering
        \includegraphics[width=\linewidth]{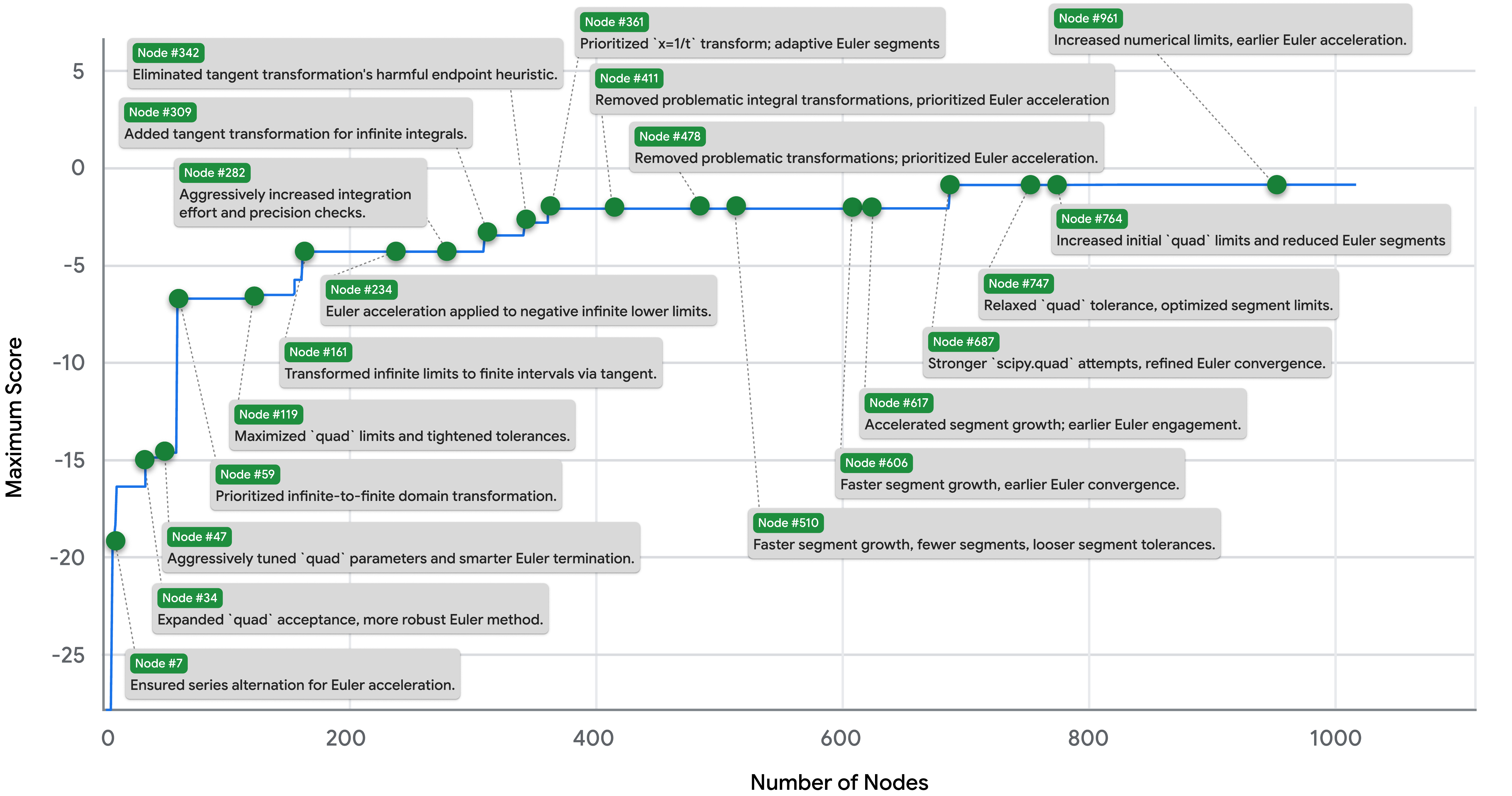}
    \end{subfigure}
    \hfill 
    \begin{subfigure}[b]{0.7\textwidth}
        \centering

        \includegraphics[width=\linewidth]{Figures/tree_visualization_pretty_integrals.pdf}
    \end{subfigure}
    
    \caption{\textbf{Breakthrough plot and solution tree for the numerical integration task.} {\sl Top Figure} Breakthrough plot for the Integral tree search, showing the evolution of the maximum score as a function of the number of nodes. The green dots label places where the score abruptly increases due to an improvement in the code, and the label describes the change in the code that resulted in the score increase. {\sl Bottom Figure} Structure of the tree for this same search. The color range consists of orange (lower scores) to green (higher scores) with the highest score denoted by a diamond node.}
    \phantomsection
    \addcontentsline{toc}{subsection}{\protect\numberline{Fig. S19:}Breakthrough plot and solution tree for the numerical integration task.}
    \label{fig:tree_breakthrough_integrals}
\end{figure}

\begin{figure}
    \centering
    \includegraphics[width=0.75\linewidth]{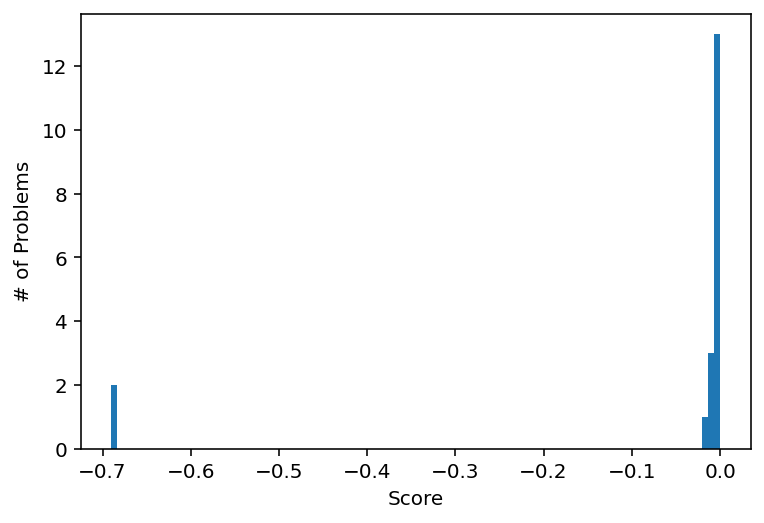}
    \caption{
        \textbf{Scores of the best numerical integration routine applied to the held-out set of 19 integrals.} Zero is a perfect score. 
        The generated function solved 17 of 19 integrals to within 3 percent.
        The standard function, \texttt{scipy.integrate.quad()} failed in all these cases.
    }
    \phantomsection
    \addcontentsline{toc}{subsection}{\protect\numberline{Fig. S20:}Scores of the best numerical integration routine applied to the held-out set of 19 integrals.}
    \label{fig:integrals-scores}
\end{figure}
\clearpage
\newpage

\newpage
\section{Supplementary Tables}
\begin{table}[htbp]
    \centering
        \begin{threeparttable}
            \caption{\textbf{Computational budget and execution costs per search node across evaluated benchmarks.} We report the average number of request and response tokens processed by the language model per node, alongside the average sandbox execution duration and hardware requirements. The overall computational cost scales linearly with the total number of nodes explored during the search.}
            \phantomsection
            \addcontentsline{toc}{subsection}{\protect\numberline{Table S1:}Computational budget and execution costs per search node across evaluated benchmarks.}
                \label{tab:compute_cost}
            \vspace{0.1in}
            \begin{tabular}{l r r r c}
                \toprule
                \textbf{Task} & \textbf{Request Tokens} & \textbf{Response Tokens} & \textbf{Duration (min)} & \textbf{Sandbox Type} \\
                \midrule
                Batch Integration & 16,171 & 4,183 & 8.0 & GPU\tnote{b} \\
                Covid Forecasting & 9,607 & 3,392 & 1.2 & CPU \\
                GIFT-Eval & 15,828 & 9,308 & 53.3 & CPU \\
                Integrals & 224,415\tnote{a} & 6,074 & 1.5 & CPU \\
                Geospatial & 7,186 & 3,172 & 16.4 & GPU \\
                ZAPBench & 16,809 & 8,036 & 192.2 & GPU \\
                \bottomrule
            \end{tabular}
            \begin{tablenotes}
                \small
                \item[a] The request token count for the integrals is artificially inflated due to tokenization of images in the prompt.
                \item[b] The GPUs used in this study are NVIDIA Tesla T4.
            \end{tablenotes}
        \end{threeparttable}

\end{table}

\begin{table}[ht]
\centering
\caption{\textbf{Kaggle Playground Series (Season 3) Competitions included in the experiments.}}
\phantomsection
\addcontentsline{toc}{subsection}{\protect\numberline{Table S2:}Kaggle Playground Series (Season 3) competitions included in the experiments.}
\label{tab:kaggle_competitions}
\begin{tabular}{ll}
\hline
\textbf{Competition Name} & \textbf{Episode} \\ \hline
Regression with a Tabular California Housing Dataset & \href{https://www.kaggle.com/competitions/playground-series-s3e1}{s3e1} \\
Binary Classification with a Tabular Stroke Prediction Dataset & \href{https://www.kaggle.com/competitions/playground-series-s3e2}{s3e2} \\
Binary Classification with a Tabular Employee Attrition Dataset & \href{https://www.kaggle.com/competitions/playground-series-s3e3}{s3e3} \\
Binary Classification with a Tabular Credit Card Fraud Dataset & \href{https://www.kaggle.com/competitions/playground-series-s3e4}{s3e4} \\
Tabular Classification with a Wine Quality Dataset & \href{https://www.kaggle.com/competitions/playground-series-s3e5}{s3e5} \\
Binary Classification with a Tabular Reservation Cancellation Dataset & \href{https://www.kaggle.com/competitions/playground-series-s3e7}{s3e7} \\
Tabular Regression with a Gemstone Price Dataset & \href{https://www.kaggle.com/competitions/playground-series-s3e8}{s3e8} \\
Regression with a Tabular Concrete Strength Dataset & \href{https://www.kaggle.com/competitions/playground-series-s3e9}{s3e9} \\
Binary Classification with a Tabular Pulsar Dataset & \href{https://www.kaggle.com/competitions/playground-series-s3e10}{s3e10} \\
Regression with a Tabular Media Campaign Cost Dataset & \href{https://www.kaggle.com/competitions/playground-series-s3e11}{s3e11} \\
Binary Classification with a Tabular Kidney Stone Prediction Dataset & \href{https://www.kaggle.com/competitions/playground-series-s3e12}{s3e12} \\
Regression with a Wild Blueberry Yield Dataset & \href{https://www.kaggle.com/competitions/playground-series-s3e14}{s3e14} \\
Regression with a Crab Age Dataset & \href{https://www.kaggle.com/competitions/playground-series-s3e16}{s3e16} \\
Binary Classification of Machine Failures & \href{https://www.kaggle.com/competitions/playground-series-s3e17}{s3e17} \\
Forecasting Mini-Course Sales & \href{https://www.kaggle.com/competitions/playground-series-s3e19}{s3e19} \\
Predict CO2 Emissions in Rwanda & \href{https://www.kaggle.com/competitions/playground-series-s3e20}{s3e20} \\ \hline
\end{tabular}
\end{table}

\begin{table}[ht]
\caption{\textbf{Prompt for Kaggle Playground competitions.} The prompt is used for ERA on the Kaggle Playground Benchmark. This example is for Season 3 Episode 17.}
\phantomsection
\addcontentsline{toc}{subsection}{\protect\numberline{Table S3:}Prompt for Kaggle Playground competitions.}
\label{table:playground_normal}
    \centering
    \begin{tabular}{l}
    \begin{tcolorbox}[title={ Prompt for Kaggle Playground Competitions}]
Please write the python code to work on a Kaggle competition. Use any model you
like.

Kaggle competition name: Binary Classification of Machine Failures

The competition is evaluated as follows: Submissions are evaluated on area under the ROC curve between the predicted probability and the observed target.
\begin{verbatim}
Submission File
For each `id` in the test set, you must predict the probability of a
`Machine failure`.  The file should contain a header and have the following
format:

    id,Machine failure
    136429,0.5
    136430,0.1
    136431,0.9
    etc.

Here are a few lines of each of the files:

file_name : sample_submission.csv

file_contents:

id,Machine failure
79996,0
100009,0
etc.

====
file_name : test.csv

file_contents:
etc.
====
file_name : train.csv

file_contents:
etc.
====
Please provide complete code that will generate the submission file in the
format below:

```python
YOUR CODE
```
\end{verbatim}
\end{tcolorbox}
    \end{tabular}
\end{table}

\newpage
\begin{table}[ht]
\caption{\textbf{Expert advice for Kaggle Playground competitions.} The prompt is used for the TS with Expert Advice on the Kaggle Playground Benchmark.}
\phantomsection
\addcontentsline{toc}{subsection}{\protect\numberline{Table S4:}Expert advice for Kaggle Playground competitions.}
\label{table:playground_expert}
    \centering
    \begin{tabular}{l}
    \begin{tcolorbox}[title={Expert advice prompt for Kaggle Playground competitions}]
Here is high level advice: Instead of putting all your effort into a single model, experiment with combining two or more models. Start with simple averaging of predictions and then explore more advanced techniques like stacking.

Try out several different types of models (e.g., gradient boosting machines, linear models, and even simpler models like logistic regression) to see how they perform.

Look for opportunities to go beyond standard preprocessing. Investigate the data for potential leaks, and consider using optimization libraries to find the best way to combine your models' predictions.

While feature engineering is a crucial skill, it's also important to recognize when it might not be the most important factor. Sometimes, the choice of model and ensembling strategy can have a bigger impact. Don't be afraid to try a more "brute-force" approach with powerful models that can handle raw data effectively.

\end{tcolorbox}
    \end{tabular}
\end{table}

\begin{table}[ht]
\caption{\textbf{Boosted decision trees for Kaggle Playground competitions.} The prompt is used for the TS with Boosted Decision Tree on the Kaggle Playground Benchmark.}
\phantomsection
\addcontentsline{toc}{subsection}{\protect\numberline{Table S5:}Boosted decision trees for Kaggle Playground competitions.}
\label{table:playground_bdt}
    \centering
    \begin{tabular}{l}
    \begin{tcolorbox}[title={Expert Advice Prompt for Kaggle Playground Competitions}]
Given the code you are given please rewrite any library code (such as  XGBoost, LightGBM, and CatBoost) by making internal
algorithmic choices that produce performant training code and models that
generalize well in many situations. Things you can try are alternative
representations of data, using different step size algorithms, using the output
of a strong learner as input to the next weak learner.
If the code contains such libraries, please extract the raw code that is being used in the library and rewrite it to improve performance.

\end{tcolorbox}
    \end{tabular}
\end{table}
\clearpage

\begin{table}[htbp]
\caption{\textbf{Example code generated by \ourmethod.} We prompted ERA to generate a solution for single-cell batch integration. We refer to this solution as \texttt{BBKNN (TS)}.}
\phantomsection
\addcontentsline{toc}{subsection}{\protect\numberline{Table S6:}Example code generated by \ourmethod.}
\label{table:sc_bbknn_code}
\begin{tcblisting}{
  listing engine=listings,
  colback=black!5,                               % Box background color
  colframe=black!75,                             % Box frame color
  title={Example code generated by \ourmethod.},
  listing only,                                  % The box contains ONLY the code listing
  listing options={basicstyle=\ttfamily\supertiny,frame=none, backgroundcolor={}}
}
from typing import Any
from sklearn.decomposition import TruncatedSVD
from sklearn.neighbors import NearestNeighbors
from scipy.sparse import lil_matrix, csr_matrix
import numpy as np
import scanpy as sc
import anndata as ad
import heapq # For efficiently getting top K elements from merged lists

# Define parameters for the config.
# These values are chosen to balance computational cost and integration performance
# for datasets with up to ~300k cells and 2k genes.
config = {
    'n_pca_components': 100,     # Number of PCA components. Recommended: 50-200.
                                 # Captures sufficient variance while reducing dimensionality.
    'n_neighbors_per_batch': 10, # Number of neighbors to find within each batch. Recommended: 5-15.
                                 # This defines the local batch context for each cell.
    'total_k_neighbors': 50,     # Total number of nearest neighbors to retain for the final graph. Recommended: 15-100.
                                 # This forms the global batch-integrated graph.
}


def eliminate_batch_effect_fn(
    adata: ad.AnnData, config: dict[str, Any]
) -> ad.AnnData:
  # Create a copy to ensure the original input adata remains unchanged.
  adata_integrated = adata.copy()

  # --- Preprocessing: Normalize, log-transform, scale ---
  # These are standard initial steps for scRNA-seq data.
  # Use adata.X which contains raw counts.
  sc.pp.normalize_total(adata_integrated, target_sum=1e4)
  sc.pp.log1p(adata_integrated)
  sc.pp.scale(adata_integrated, max_value=10) # Clip values to avoid extreme outliers

  # --- Batch Correction: ComBat on the gene expression matrix ---
  # This step applies a more robust linear model-based batch correction
  # directly on the gene expression data before dimensionality reduction.
  # ComBat modifies adata_integrated.X in place.
  sc.pp.combat(adata_integrated, key='batch')

  # --- Dimensionality Reduction: PCA on the ComBat-corrected data ---
  # n_comps cannot exceed min(n_obs - 1, n_vars). Robustly handle small datasets.
  n_pca_components = config.get('n_pca_components', 100)
  actual_n_pca_components = min(n_pca_components, adata_integrated.n_vars, adata_integrated.n_obs - 1)

  # Handle edge cases for PCA and graph construction where data is too small.
  # If PCA cannot be run meaningfully, return a minimal AnnData object to avoid errors.
  if actual_n_pca_components <= 0 or adata_integrated.n_obs <= 1:
      print(f"Warning: Too few observations ({adata_integrated.n_obs}) or dimensions ({adata_integrated.n_vars}) for PCA/graph construction. Returning trivial embedding.")
      # Provide a placeholder embedding and empty graph structure.
      adata_integrated.obsm['X_emb'] = np.zeros((adata_integrated.n_obs, 1))
      adata_integrated.obsp['connectivities'] = csr_matrix((adata_integrated.n_obs, adata_integrated.n_obs))
      adata_integrated.obsp['distances'] = csr_matrix((adata_integrated.n_obs, adata_integrated.n_obs))
      adata_integrated.uns['neighbors'] = {
          'params': {
              'n_neighbors': 0,
              'method': 'degenerate',
              'n_pcs': 0,
              'n_neighbors_per_batch': 0,
              'pca_batch_correction': 'none',
          },
          'connectivities_key': 'connectivities',
          'distances_key': 'distances',
      }
      return adata_integrated

  sc.tl.pca(adata_integrated, n_comps=actual_n_pca_components, svd_solver='arpack')

  # Set the ComBat-corrected PCA embedding as the integrated output embedding.
  # This 'X_emb' will be directly evaluated by metrics like ASW, LISI, PCR.
  adata_integrated.obsm['X_emb'] = adata_integrated.obsm['X_pca']


  # --- Custom Batch-Aware Nearest Neighbors Graph Construction ---
  # This implements the expert advice: find neighbors independently within batches, then merge.
  # This part of the code remains largely the same, but now operates on the
  # ComBat-corrected PCA embedding (adata_integrated.obsm['X_emb']).
  k_batch_neighbors = config.get('n_neighbors_per_batch', 10)
  total_k_neighbors = config.get('total_k_neighbors', 50)

  # A list of dictionaries to store unique neighbors and their minimum distances for each cell.
  # Using dictionaries allows efficient updating if a cell is found as a neighbor from multiple batches.
  merged_neighbors_per_cell = [{} for _ in range(adata_integrated.n_obs)]

  # Group cell indices by batch for efficient querying.
  batches = adata_integrated.obs['batch'].values
  unique_batches = np.unique(batches)
  batch_to_indices = {b: np.where(batches == b)[0] for b in unique_batches}

  # Pre-fit NearestNeighbors models for each batch's data using the corrected PCA embedding.
  # This avoids refitting the model for every query.
  batch_nn_models = {}
  for b_id in unique_batches:
    batch_cell_indices = batch_to_indices[b_id]
    # Ensure there are enough cells to fit a NearestNeighbors model (at least k_batch_neighbors + 1 for self-exclusion, or just > 0 for min k=1)
    if len(batch_cell_indices) > 0:
        # Fit with a k that is at most the batch size to avoid errors if k_batch_neighbors is too high for a small batch.
        k_fit_effective = min(k_batch_neighbors + 1, len(batch_cell_indices)) # +1 to ensure self-loop can be found and excluded
\end{tcblisting}
\end{table}

\begin{table}[htbp]
\begin{tcblisting}{
  listing engine=listings,
  colback=black!5,                               % Box background color
  colframe=black!75,                             % Box frame color
  title={Example code generated by \ourmethod (continued).},
  listing only,                                  % The box contains ONLY the code listing
  listing options={basicstyle=\ttfamily\supertiny,frame=none, backgroundcolor={}}
}
        if k_fit_effective > 0: # Only fit if there are points available
            nn_model = NearestNeighbors(n_neighbors=k_fit_effective, metric='euclidean', algorithm='auto')
            nn_model.fit(adata_integrated.obsm['X_emb'][batch_cell_indices])
            batch_nn_models[b_id] = nn_model

  # Iterate through all possible query batches and target batches to find neighbors.
  for query_batch_id in unique_batches:
      query_global_indices = batch_to_indices[query_batch_id]
      if len(query_global_indices) == 0:
          continue # Skip empty query batches

      query_data = adata_integrated.obsm['X_emb'][query_global_indices]

      for target_batch_id in unique_batches:
          if target_batch_id not in batch_nn_models:
              continue # Skip target batches that were too small to fit an NN model
          
          nn_model = batch_nn_models[target_batch_id]
          target_global_indices = batch_to_indices[target_batch_id]

          # Ensure n_neighbors does not exceed the number of points in the target batch.
          k_for_query = min(k_batch_neighbors, len(target_global_indices) -1) # -1 to avoid finding self as neighbor if batch is query batch
          if k_for_query <= 0: # No valid neighbors can be found in this target batch
            continue

          # Query neighbors for all cells in the current query batch against the target batch's data.
          distances, indices_in_target_batch = nn_model.kneighbors(query_data, n_neighbors=k_for_query, return_distance=True)

          for i_query_local in range(len(query_global_indices)):
              current_cell_global_idx = query_global_indices[i_query_local]
              
              dists_for_cell = distances[i_query_local]
              global_neighbors_for_cell = target_global_indices[indices_in_target_batch[i_query_local]]

              for k_idx in range(len(global_neighbors_for_cell)):
                  neighbor_global_idx = global_neighbors_for_cell[k_idx]
                  dist = dists_for_cell[k_idx]
                  
                  # Exclude self-loops: a cell should not be its own neighbor in graph construction.
                  if neighbor_global_idx == current_cell_global_idx:
                      continue

                  # Store neighbor and its distance. If already present, keep the minimum distance (closest connection).
                  if (neighbor_global_idx not in merged_neighbors_per_cell[current_cell_global_idx] or
                      dist < merged_neighbors_per_cell[current_cell_global_idx][neighbor_global_idx]):
                      merged_neighbors_per_cell[current_cell_global_idx][neighbor_global_idx] = dist

  # Convert collected neighbors and distances into sparse matrices.
  rows = []
  cols = []
  data_distances = []

  for i in range(adata_integrated.n_obs):
    # Retrieve all candidate neighbors for cell 'i', sort by distance, and take the top 'total_k_neighbors'.
    current_cell_candidates = list(merged_neighbors_per_cell[i].items())
    
    if not current_cell_candidates: # If a cell has no valid neighbors after all filtering
        continue

    # Use heapq for efficient selection of the smallest distances.
    selected_neighbors = heapq.nsmallest(total_k_neighbors, current_cell_candidates, key=lambda item: item[1])

    for neighbor_idx, dist in selected_neighbors:
        rows.append(i)
        cols.append(neighbor_idx)
        data_distances.append(dist)

  # Create distance matrix. Handle case with no neighbors found at all for the entire dataset.
  if not rows:
      distances_matrix = csr_matrix((adata_integrated.n_obs, adata_integrated.n_obs))
  else:
      distances_matrix = csr_matrix((data_distances, (rows, cols)), shape=(adata_integrated.n_obs, adata_integrated.n_obs))
  
  # Symmetrize the distance matrix: if A is a neighbor of B, then B is also a neighbor of A,
  # with the distance being the maximum of the two observed distances (ensures undirected graph).
  distances_matrix = distances_matrix.maximum(distances_matrix.T)
  distances_matrix.eliminate_zeros() # Remove any explicit zeros created by max operation

  # Create connectivities matrix (binary representation of connections).
  connectivities_matrix = distances_matrix.copy()
  connectivities_matrix.data[:] = 1.0  # All non-zero entries become 1.0 (connected).
  connectivities_matrix.eliminate_zeros()
  connectivities_matrix = connectivities_matrix.astype(float)

  # Store the custom graph in adata.obsp. These keys are used by scib metrics.
  adata_integrated.obsp['connectivities'] = connectivities_matrix
  adata_integrated.obsp['distances'] = distances_matrix

  # Store parameters in adata.uns['neighbors'] for completeness and scanpy/scib compatibility.
  adata_integrated.uns['neighbors'] = {
      'params': {
          'n_neighbors': total_k_neighbors,
          'method': 'custom_batch_aware_combat_pca', # Reflects the integration strategy
          'metric': 'euclidean',
          'n_pcs': actual_n_pca_components,
          'n_neighbors_per_batch': k_batch_neighbors,
          'pca_batch_correction': 'combat', # Indicates ComBat was applied before PCA
      },
      'connectivities_key': 'connectivities',
      'distances_key': 'distances',
  }

  return adata_integrated
\end{tcblisting}
\end{table}

\clearpage
\newpage

\begin{table}[ht]
\caption{\textbf{Prompt for recombination of baseline method ideas.} The prompt instructs Gemini to identify the main differences in the principles of top-performing solutions, obtained from tree search runs seeded with baseline methods. This generated summary then serves as part of an explicit instruction for tree search to create hybrid strategies.}
\phantomsection
\addcontentsline{toc}{subsection}{\protect\numberline{Table S7:}Prompt for recombination of baseline method ideas.}
\label{table:sc_recombine_summary_prompt}
    \centering
    \begin{tabular}{l}
    \begin{tcolorbox}[title={Prompt for summarizing differences between two baseline methods.}]
Compare these two code solutions to the same problem of integrating single-cell batch effects. Explain the main principles that differ between the codes:\\

CODE 1: [CODE FROM BASELINE 1]\\

CODE 2: [CODE FROM BASELINE 2]
\end{tcolorbox}
    \end{tabular}
\end{table}

\newpage

\begin{table}
    \centering
    \footnotesize 
    \caption{\textbf{Configuration of COVID-19 forecasting data splits.}}
    \phantomsection
    \addcontentsline{toc}{subsection}{\protect\numberline{Table S8:}Configuration of COVID-19 forecasting data splits.}
    \label{tab:covid_splits}
    \begin{tabular*}{\textwidth}{@{\extracolsep{\fill}} p{0.18\textwidth} p{0.28\textwidth} p{0.18\textwidth} c c c}
        \toprule
        \textbf{Strategy} & \textbf{Validation Reference Dates} & \textbf{Test Reference Dates} & \textbf{Horizons} & \textbf{Locations} & \textbf{Metric} \\
        \midrule
        \textbf{Google Retro. 1} & 2024-10-05, 2024-10-12, 2024-10-19, 2024-10-26, 2024-11-02, 2024-11-09 & 2024-11-16, 2024-11-23, 2024-11-30 & 1--4 weeks & 52 US & WIS \\
        \addlinespace
        \textbf{Google Retro. 2} & 2024-10-19, 2024-10-26, 2024-11-02, 2024-11-09, 2024-11-16, 2024-11-23 & 2024-11-30, 2024-12-07, 2024-12-14 & 1--4 weeks & 52 US & WIS \\
        \addlinespace
        \textbf{Google Retro. 3} & 2024-11-02, 2024-11-09, 2024-11-16, 2024-11-23, 2024-11-30, 2024-12-07 & 2024-12-14, 2024-12-21, 2024-12-28 & 1--4 weeks & 52 US & WIS \\
        \addlinespace
        \textbf{Google Retro. 4} & 2024-11-16, 2024-11-23, 2024-11-30, 2024-12-07, 2024-12-14, 2024-12-21 & 2024-12-28, 2025-01-04, 2025-01-11 & 1--4 weeks & 52 US & WIS \\
        \addlinespace
        \textbf{Google Retro. 5} & 2024-11-30, 2024-12-07, 2024-12-14, 2024-12-21, 2024-12-28, 2025-01-04 & 2025-01-11, 2025-01-18, 2025-01-25 & 1--4 weeks & 52 US & WIS \\
        \addlinespace
        \textbf{Google Retro. 6} & 2024-12-14, 2024-12-21, 2024-12-28, 2025-01-04, 2025-01-11, 2025-01-18 & 2025-01-25, 2025-02-01, 2025-02-08 & 1--4 weeks & 52 US & WIS \\
        \addlinespace
        \textbf{Google Retro. 7} & 2024-12-28, 2025-01-04, 2025-01-11, 2025-01-18, 2025-01-25, 2025-02-01 & 2025-02-08, 2025-02-15, 2025-02-22 & 1--4 weeks & 52 US & WIS \\
        \addlinespace
        \textbf{Google Retro. 8} & 2025-01-11, 2025-01-18, 2025-01-25, 2025-02-01, 2025-02-08, 2025-02-15 & 2025-02-22, 2025-03-01, 2025-03-08 & 1--4 weeks & 52 US & WIS \\
        \addlinespace
        \textbf{Google Retro. 9} & 2025-01-25, 2025-02-01, 2025-02-08, 2025-02-15, 2025-02-22, 2025-03-01 & 2025-03-08, 2025-03-15, 2025-03-22 & 1--4 weeks & 52 US & WIS \\
        \addlinespace
        \textbf{Google Retro. 10} & 2025-02-08, 2025-02-15, 2025-02-22, 2025-03-01, 2025-03-08, 2025-03-15 & 2025-03-22, 2025-03-29, 2025-04-05 & 1--4 weeks & 52 US & WIS \\
        \addlinespace
        \textbf{Google Retro. 11} & 2025-02-22, 2025-03-01, 2025-03-08, 2025-03-15, 2025-03-22, 2025-03-29 & 2025-04-05, 2025-04-12, 2025-04-19 & 1--4 weeks & 52 US & WIS \\
        \addlinespace
        \textbf{Google Retro. 12} & 2025-03-08, 2025-03-15, 2025-03-22, 2025-03-29, 2025-04-05, 2025-04-12 & 2025-04-19, 2025-04-26, 2025-05-03 & 1--4 weeks & 52 US & WIS \\
        \addlinespace
        \textbf{Google Retro. 13} & 2025-03-22, 2025-03-29, 2025-04-05, 2025-04-12, 2025-04-19, 2025-04-26 & 2025-05-03, 2025-05-10, 2025-05-17 & 1--4 weeks & 52 US & WIS \\
        \addlinespace
        \textbf{Google Retro. 14} & 2025-04-05, 2025-04-12, 2025-04-19, 2025-04-26, 2025-05-03, 2025-05-10 & 2025-05-17, 2025-05-24, 2025-05-31 & 1--4 weeks & 52 US & WIS \\
        \midrule
        \textbf{All other TS} & 2025-02-22, 2025-03-01, 2025-03-08, 2025-03-15, 2025-03-22, 2025-03-29 & 2025-04-05, 2025-04-12, 2025-04-19 & 1--4 weeks & 52 US & WIS \\
        \bottomrule
    \end{tabular*}
\end{table}

\newpage
\clearpage
\captionsetup{type=table}
\captionof{table}{\textbf{Method descriptions used for replicating COVID-19 models submitted to the CDC's CovidHub.}}
\phantomsection
\addcontentsline{toc}{subsection}{\protect\numberline{Table S9:}Method descriptions used for replicating COVID-19 models submitted to the CDC's CovidHub.}
\label{table:covid-rep-methods}

\begin{tcolorbox}[title={\texttt{CEPH-Rtrend\_covid}}]
    ``Use a renewal equation method based on Bayesian estimation of Rt from hospitalization data. Model forecasts should be obtained by using a renewal equation based on the estimated net reproduction number Rt. Apply a lowpass filter to the time series of weekly hospitalizations, then interpolate it to daily resolution. Then use MCMC Metropolis-Hastings sampling to estimate the posterior distribution of Rt based on the filtered data, considering an informed prior on Rt based on COVID-19 literature. The estimated Rt in the last weeks of available data is used to forecast Rt in the upcoming weeks, with a drift term proportional to the current incidence. Finally, use the renewal equation with the posterior distribution and trend of the estimated Rt in the most recent weeks of hospitalization data."
\end{tcolorbox}

\begin{tcolorbox}[title={\texttt{CMU-TimeSeries}}]
    ``Use an ensemble of AR-based time-series models, involving a basic quantile autoregression fit using lagged values of covid-related hospitalization counts (normalized by population). The data should be smoothed in time. Fit the model jointly across all jurisdictions using the most recently available 21 days of training data. Learn each of the 23 quantiles using a separate quantile regression with nonnegativity and quantile sorting constraints applied post hoc."
\end{tcolorbox}

\begin{tcolorbox}[title={\texttt{CMU-climate\_baseline}}]
    ``Use an ensemble of historically formed quantiles. Using data from 2022 onwards, this climatological model should use samples from the 7 weeks centered around the target week and reference week to form the quantiles for the target week, as one might use climate information to form a meteorological forecast. To get more variation at some potential issue of generalization, one can form quantiles after aggregating across geographic values as well as years (after converting to a rate based case count). This model should use a simple average of the geo-specific quantiles and the geo-aggregated quantiles."
\end{tcolorbox}

\begin{tcolorbox}[title={\texttt{JHU\_CSSE-CSSE\_Ensemble}}]
    ``Use a Multi-Pathogen Optimized Geo-Hierarchical Ensemble Framework (MPOG-Ensemble). Forecast state-level COVID-19 hospitalizations using a combination of time series forecasting methods, organized across three hierarchical levels. At the individual state level, forecasts are generated using Holt-Winters Exponential Smoothing. For regional predictions, which group states based on past 2 years covid-19 activity trends identified through the Louvain method, Long Short-Term Memory (LSTM) models are employed. Additionally, a LSTM model that covers all states is implemented. These three-tiered model outputs are integrated, selecting weights based on their recent performance in terms of Mean Absolute Error (MAE) to produce the final prediction."
\end{tcolorbox}

\begin{tcolorbox}[title={\texttt{OHT\_JHU-nbxd}}]
    ``Use a neural network that encodes the data inputs using a TCN (Bai et al. 2018) and decodes the result into a forecast using N-BEATS (Oreshkin et al. 2000). This is a residual block type architecture that generates point forecasts from univariate time series data. The network accepts a fixed lookback window of time points as input, and has a set number of output nodes corresponding to the length of the forecast horizon. Extend the network with additional residual blocks that output error variance forecasts (evaluated using a likelihood loss function) which allows generating quantile forecasts, assuming a parametric (gamma) error distribution. Additional predictor variables are incorporated using a temporal convolutional network (TCN; Bai et al. 2018). The TCN accepts one input channel for each predictor time series (or static variable), including past values of the target variable, and outputs a single channel with the same length as the lookback window. The TCN output channel is used as the input to the extended N-BEATS network. Each value in the TCN output sequence is a non-linear combination of the predictor variables at that point and all previous points in the lookback window, which preserves the temporal structure of the input. Forecast is the median of an ensemble of such models with varying lookback window sizes and random initializations."
\end{tcolorbox}

\begin{tcolorbox}[title={\texttt{UM-DeepOutbreak}}]
    ``Use a deep neural network model with conformal predictions. The neural network architecture is a sequence-to-sequence model based on recurrent units and self-attention modules. It is trained in a multi-task setting where each region is considered a task. The uncertainty quantification is conducted post hoc with conformal predictions that follows adaptive conformal inference to adapt to distribution shifts. Spatial correlation is not considered."
\end{tcolorbox}

\begin{tcolorbox}[title={\texttt{UMass-ar6\_pooled}}]
    ``Use an autoregressive model with shared coefficients across locations: AR(6) model after fourth root data transform. AR coefficients are shared across all locations. A separate variance parameter is estimated for each location."
\end{tcolorbox}

\begin{tcolorbox}[title={\texttt{UMass-gbqr}}]
    ``Use gradient boosting quantile regression. Do gradient boosting using features summarizing signal activity, properties of the location, information about the timing of forecast creation, and the forecast horizon."
\end{tcolorbox}

\clearpage
\newpage
\begin{table}[ht]
\caption{\textbf{Prompt for replicating COVID-19 models submitted to CovidHub by injecting method descriptions as \texttt{\{method\}} into existing tree search prompt.}}
\phantomsection
\addcontentsline{toc}{subsection}{\protect\numberline{Table S10:}Prompt for replicating COVID-19 models submitted to CovidHub by injecting method descriptions as \texttt{\{method\}} into existing tree search prompt.}
\label{table:covid-rep-prompt}
    \centering
    \begin{tabular}{l}
    \begin{tcolorbox}[title={Prompt for replicating models submitted to CovidHub.}]
    
    Please write the python code to work on a competition.
    
    \texttt{\{method\}} 

    I've already loaded the train / test files and split out the x and y parts.

    Please provide a new definition for the function below, complete with imports,
    that will generalize well. However, do not do any cross-validation in here.
    Your function should expect options to be passed in via the config argument.
    I'll use cross-validation myself to select which of the options in the \texttt{config\_list} generalizes best.

    \texttt{\{method\}}

\begin{verbatim}
from typing import Any  # Don't forget this!
import pandas as pd

def fit_and_predict_fn(
   train_x: pd.DataFrame,
   train_y: pd.Series,
   test_x: pd.DataFrame,
   config: dict[str, Any]) -> pd.Series:
    """Make predictions for test_x by modeling train_x to train_y.
    Do not do any cross-validation in here.
    """
    mean_y = np.mean(train_y)
    return pd.Series([mean_y] * len(test_x), index=test_x.index)

    # These will get scored by code that I supply. You'll get back a summary
    # of the performance of each of them.

config_list = [{}]

\end{verbatim}
    
And format it like this:
\begin{verbatim}
# YOUR CODE
# YOUR config_list
\end{verbatim}
    
\end{tcolorbox}
    \end{tabular}
\end{table}

\newpage
\begin{longtable}{>{\raggedright\arraybackslash}p{3.7cm}Y{2.3cm}p{9.7cm}}
\caption{\textbf{Expert manual inspection of adherence of \ourmethod implementation to COVID-19 modeling methods.}}
    \phantomsection
    \addcontentsline{toc}{subsection}{\protect\numberline{Table S11:}Expert manual inspection of adherence of \ourmethod implementation to COVID-19 modeling methods.}
\label{table:covid_method_adherence}\\
\toprule[1.5pt]
\textbf{Method} & \textbf{Judgment} & \textbf{Notes} \\
\midrule
\endfirsthead

\multicolumn{3}{c}%
{{\bfseries \tablename\ \thetable{} -- continued from previous page}} \\
\toprule[1.5pt]
\textbf{Method} & \textbf{Judgment} & \textbf{Notes} \\
\midrule
\endhead

\midrule
\multicolumn{3}{r}{{\footnotesize Continued on next page}} \\
\endfoot

\bottomrule[1.5pt]
\endlastfoot

CEPH-Rtrend\_covid x CMU-TimeSeries & Follow & 
\\

CEPH-Rtrend\_covid x CMU-climate\_ baseline & Follow & 
\\

CEPH-Rtrend\_covid x JHU\_CSSE- CSSE\_Ensemble & Follow & 
\\

CEPH-Rtrend\_covid x OHT\_JHU-nbxd & Follow & Translates $R_t$ into engineered features (lagged differences, ratios). \\

CEPH-Rtrend\_covid and UM-DeepOutbreak & Follow & Feeds mechanistic-inspired features 
into GRU-based encoder, predicts quantiles via pinball loss. \\

CEPH-Rtrend\_covid x UMass-ar6\_pooled & Follow + Innovate & 
Simulates from normal distribution in transformed space then inverse transforms to derive quantiles. \\

CEPH-Rtrend\_covid x UMass-gbqr & Follow & Implements mechanistic model components as input features to ML model. \\

CMU-TimeSeries x CMU-climate\_baseline & Follow &
AR model with climatological features as predictors.\\

CMU-TimeSeries x JHU\_CSSE-CSSE\_Ensemble & Follow & Hierarchical ensemble of QuantReg AR models with performance-based weighting. \\

CMU-TimeSeries x OHT\_JHU-nbxd & Follow & Ensemble of bagged QuantReg AR models. \\

CMU-TimeSeries x UM-DeepOutbreak & Follow & LightGBM quantile regression models with iterative forecasting + conformal-like calibration. \\

CMU-TimeSeries x UMass-ar6\_pooled & Follow & Ensemble of AR QuantReg models on fourth-root transformed data. \\

CMU-TimeSeries x UMass-gbqr & Follow & LightGBM quantile models on population-normalized data with (un)smoothed lags + direct multi-horizon prediction. \\

CMU-climate\_baseline x JHU\_CSSE-CSSE\_Ensemble & Follow & Hierarchical ensemble of climatological models. \\

CMU-climate\_baseline x OHT\_JHU-nbxd & Follow & Feeds climatological quantiles into LightGBM to learn directly from seasonal baseline. \\

CMU-climate\_baseline x UM-DeepOutbreak & Follow & LightGBM to predict central trend + climatological model for empirical quantile spreads.\\

CMU-climate\_baseline x UMass-ar6\_pooled & Follow & Seasonally-aware method for estimating uncertainty based on empirical quantiles of AR residuals. \\

CMU-climate\_baseline x UMass-gbqr & Follow & Feeds climatological statistics as features into LightGBM. \\

JHU\_CSSE-CSSE\_Ensemble x OHT\_JHU-nbxd & Partially Follow & Hierarchical structure (state, regional, national models) + adaptive MAE-weighting. \\

JHU\_CSSE-CSSE\_Ensemble x UM-DeepOutbreak & Follow + Innovate & Secondary model to predict error magnitudes \& find quantiles of normalized residuals. \\

JHU\_CSSE-CSSE\_Ensemble x UMass-ar6\_pooled & Follow & 
\\

JHU\_CSSE-CSSE\_Ensemble x UMass-gbqr & Follow & Combines predictions from `adaptive' model trained on recent data \& `stable' model trained on longer history. \\

OHT\_JHU-nbxd x UM-DeepOutbreak & Follow & 
\\

OHT\_JHU-nbxd x UMass-ar6\_pooled & Follow & Feature engineering + ensembling + variance-stabilizing transformation \_ recursive forecasting. \\

OHT\_JHU-nbxd x UMass-gbqr & Follow & Uses LightGBM predicts parameters of Gamma distribution. \\

UM-DeepOutbreak x UMass-ar6\_pooled & Follow & 
\\

UM-DeepOutbreak x UMass-gbqr & Follow & 
\\

UMass-ar6\_pooled x UMass-gbqr & Follow & LightGBM quantile regression on fourth-root transformed target.
\\

DEEP-RESEARCH-CSTGT & Follow & 
Simplified static graph + synthetically generated policy feature. \\

DEEP-RESEARCH-MetaEnsembler & Follow & 
Meta-model to predict WIS.\\

DEEP-RESEARCH-FairnessAwareOptimization & Follow & Iterative re-weighting approximates composite fairness loss. \\

DEEP-RESEARCH-RegimeSwitchingDetection & Follow & 
\\

CO-SCIENTIST-STGNN-AgACI & Does not Follow & AR quantile regression model using LightGBM. 
Omits AgACI stage, replaces with simpler post-processing. \\

CO-SCIENTIST-MAPS & Partially Follow & 3-stage ensemble: substitutes core models (GNN, TCN, GPR, MLP) with feature-engineered LightGBM proxies. \\

DEEP-RESEARCH-GenomiWastewater Fusion & Follow & 
Uses mock API calls. \\

DEEP-RESEARCH-AdversarialRecalibration & Follow + Innovate & 
Implements a post-hoc GAN structure. 
Composite loss function combining adversarial + pinball loss. \\

DEEP-RESEARCH-BehavioralSensing & Follow & Simulates external data. \\

DEEP-RESEARCH-HierarchicalBayesian NODE & Follow &
Three-level model: Negative Binomial observation layer, Neural ODE for jurisdiction-level dynamics, global hyperpriors for partial pooling. \\

CO-SCIENTIST-HGPC & Partially Follow & LightGBM quantile regression, uses feature engineering as proxy for complex stages. \\

DEEP-RESEARCH-PIDM & Follow & Implements conditional Denoising Diffusion Probabilistic Model (DDPM) with U-Net backbone, with loss function a weighted composite of standard diffusion loss and a physics-based regularization term derived from an SEIR-H model's outputs. Probabilistic forecasts generated by sampling from the learned reverse process. \\

CO-SCIENTIST-HQE & Partially Follow & Trains multiple base models, feeds their predictions into a meta-learner, then applies a conformal prediction step to adjust final quantiles. Uses multiple LightGBM models instead of suggested Prophet/TBATS for diversity, manually implements conformal prediction instead of using MAPIE. \\

DEEP-RESEARCH-CounterfactualSimulation & Follow + Innovate & Follows Monte Carlo structure: defines uncertain drivers with distributions, simulates N trajectories by applying sampled shocks to base median forecast, calculates empirical quantiles. Introduces Poisson noise on top of scenario-driven forecasts. \\

rep-OHT\_JHU-nbxd & Follow & Implements TCN encoder and N-BEATS decoder architecture, including extension of parallel residual blocks to forecast mean and variance for Gamma distribution. The final forecast is generated as a median of an ensemble with varying lookback windows and initializations. \\

rep-CMU-TimeSeries & Follow & Implements a quantile autoregression model fit jointly across jurisdictions on smoothed, population-normalized data. \\

rep-UMass-ar6\_pooled & Follow & Uses OLS on lagged, fourth-root transformed data to create a shared-coefficient AR model, then calculates separate variance parameters for each location based on residuals. \\

rep-UM-DeepOutbreak & Follow & Implements sequence-to-sequence model using a GRU and self-attention, with location embeddings. Uncertainty quantified post hoc using split conformal prediction on a recent time window. \\

rep-UMass-gbqr & Follow & Uses LightGBM with engineered features (lags for signal activity, location and population for location properties, date components for timing, and the horizon itself). \\

rep-JHU\_CSSE-CSSE\_Ensemble & Follow + Innovate & Implements three-tiered hierarchical ensemble, using Holt-Winters, regional LSTMs with Louvain grouping, and a national LSTM, combined with MAE-based weighting. Uses scaled residuals to create prediction intervals that adapt to the magnitude of the forecast to generate quantile predictions. \\

rep-CMU-climate\_baseline & Follow + Innovate & Averages geo-specific and geo-aggregated quantiles within a centered weekly window. Introduces a configurable 'smoothing\_factor', which regularizes final predictions by pulling them towards zero. \\

rep-CEPH-Rtrend\_covid & Follow & Lowpass filtering, daily interpolation, MCMC for Bayesian Rt estimation, and a renewal equation forecast. The Rt forecast correctly incorporates a sophisticated drift term that is modulated by the current incidence level. \\

retro\_1 & Follow & \\

\end{longtable}
\clearpage
\newpage

\begin{table}[h!]
\centering
\caption{\textbf{Full GIFT-Eval leaderboard (05/18/2025 snapshot).}}
\phantomsection
\addcontentsline{toc}{subsection}{\protect\numberline{Table S12:}Full GIFT-Eval leaderboard (05/18/2025 snapshot).}
\label{tab:leaderboard_snapshot_mase_full}
\begin{tabular*}{\textwidth}{l @{\extracolsep{\fill}} r l}
\toprule
\textbf{Model} & \textbf{MASE} & \textbf{Type} \\
\midrule
\textbf{Per-dataset} & \textbf{0.671} & \textbf{ERA} \\
TTM-R2-Finetuned & 0.679 & fine-tuned \\
timesfm\_2\_0\_500m & 0.680 & pretrained \\
TabPFN-TS & 0.692 & pretrained \\
chronos\_bolt\_base & 0.725 & pretrained \\
\textbf{Unified} & \textbf{0.734} & \textbf{ERA} \\
chronos\_bolt\_small & 0.738 & pretrained \\
PatchTST & 0.762 & deep-learning \\
TEMPO\_ensemble & 0.773 & fine-tuned \\
VisionTS & 0.775 & pretrained \\
Chronos\_large & 0.781 & pretrained \\
Moirai\_large & 0.785 & pretrained \\
Chronos\_base & 0.786 & pretrained \\
Chronos\_small & 0.800 & pretrained \\
Moirai\_base & 0.809 & pretrained \\
TFT & 0.822 & deep-learning \\
N-BEATS & 0.842 & deep-learning \\
Moirai\_small & 0.849 & pretrained \\
TTM-R2-Zeroshot & 0.915 & pretrained \\
DLinear & 0.952 & deep-learning \\
Auto\_Arima & 0.964 & statistical \\
TimesFM & 0.967 & pretrained \\
TTM-R1-Zeroshot & 0.969 & pretrained \\
Auto\_Theta & 0.978 & statistical \\
TIDE & 0.980 & deep-learning \\
Seasonal\_Naive & 1.000 & statistical \\
Timer & 1.019 & pretrained \\
Auto\_ETS & 1.088 & statistical \\
Lag-Llama & 1.102 & pretrained \\
DeepAR & 1.206 & deep-learning \\
Naive & 1.260 & statistical \\
Crossformer & 2.310 & deep-learning \\
\bottomrule
\end{tabular*}
\end{table}
\clearpage
\newpage

\begin{table}[htbp]
\centering
\caption{\textbf{Example configurations from the final \texttt{unified solution} for the GIFT-Eval task.} Each dictionary defines a complete forecasting strategy discovered by the tree search, combining different components of the Iterative Decomposition Model. The validation process selects the best configuration for each dataset.}
\phantomsection
\addcontentsline{toc}{subsection}{\protect\numberline{Table S13:}Example configurations from the final \texttt{unified solution} for the GIFT-Eval task.}
\label{tab:unified_config_list}
\begin{tabular}{@{}p{\linewidth}@{}}
\toprule
\textbf{Unified Solution Example Configurations} \\
\midrule
\begin{tcolorbox}[colback=white, boxrule=0pt, arc=0pt, outer arc=0pt, left=2pt, right=2pt, top=2pt, bottom=2pt]
\small
\begin{verbatim}
config_list = [
  {
      'name': 'seasonal_naive_baseline',
      'description': 'Robust baseline...',
      'components': [{'type': 'base', 'method': 'seasonal_naive_adaptive'}],
      'transform_log': False, 'non_negative': False, 'version': 4,
  },
  {
      'name': 'additive_damped_linear_LogTransform',
      'description': 'General-purpose additive model...',
      'components': [
          {'type': 'base', 'method': 'median_all'},
          {'type': 'trend', 'method': 'polynomial', 'degree': 1, 'damping_factor': 0.90},
          {'type': 'seasonal', 'method': 'average', 'window_multiplier': 5.0},
          {'type': 'residual', 'method': 'median', 'window_size': 18, 'decay_factor': 0.90},
      ],
      'transform_log': True, 'non_negative': True, 'version': 4,
  },
  {
      'name': 'date_features_seasonal',
      'description': 'Robust additive model with key cyclical and datetime features...',
      'components': [
          {'type': 'base', 'method': 'median_all'},
          {'type': 'datetime', 'features': [
              ['dayofweek', 'hour'], 'month', 'is_month_start', 'weekofyear',
              'is_weekend', 'is_quarter_start',
              {'name': '_is_holiday_flag', 
               'country_codes': ['US', 'DE', 'CN', 'GB', 'CA', 'AU']}
          ]},
          {'type': 'seasonal', 'method': 'average', 'window_multiplier': 4.0},
          {'type': 'residual', 'method': 'median', 'window_size': 14, 'decay_factor': 0.92},
      ],
      'transform_log': False, 'non_negative': False, 'version': 4,
  },
  % ... other configurations can be added here ...
]
\end{verbatim}
\end{tcolorbox}
\\
\bottomrule
\end{tabular}
\end{table}

\begin{table}[h]
    \caption{\textbf{Comparison of model performance on the DLRSD benchmark.} The table shows the publication year, architecture, key features, and reported mean Intersection over Union (mIoU) for tree search solutions and the methods from the referenced papers.}
    \phantomsection
    \addcontentsline{toc}{subsection}{\protect\numberline{Table S14:}Comparison of model performance on the DLRSD benchmark.}
    \begin{tabular*}{\textwidth}{@{\extracolsep{\fill}} l c p{0.17\textwidth} p{0.21\textwidth} c}
        \toprule
        \textbf{Method} & \textbf{Year} & \textbf{Architecture Type} & \textbf{Key Features / Techniques} & \textbf{mIoU} \\
        \midrule
        \textbf{Solution 1 (TS)} & 2025 & CNN (UNet++) & `efficientnet-b7' encoder, 8-fold TTA & 0.81 \\
        \textbf{Solution 2 (TS)} & 2025 & Transformer(SegFormer) & `mit-b1' encoder, 4-fold TTA & 0.82 \\
        \textbf{Solution 3 (TS)} & 2025 & CNN (U-Net) & `se\_resnext101\_32x4d' encoder, 7-fold TTA & 0.80 \\
        \midrule
        RE-Net \cite{Zhong2021RENet} & 2021 & CNN (Region-based) & Region Context Learning & 0.762 \\
        FURSformer \cite{Zhang2023FURSformer} & 2023 & CNN+Transformer & Custom fusion module & 0.753 \\
        SCGLU-Net \cite{Atiampo2024SCGLUNet} & 2024 & CNN+Attention & Spatial-Channel-Global-Local block & 0.666 \\
        MA-UNet \cite{Sun2022MAUNet} & 2022 & Attention+U-Net & Residual encoder with simAM & 0.619 \\
        W13 Net \cite{Elgamily2025W13Net} & 2025 & CNN (Lightweight) & Multi-stage encoding-decoding & 0.580 \\
        \bottomrule
    \end{tabular*}
    \label{tab:dlrsd_performance_summary}
\end{table}

\clearpage
\newpage
\begin{table}[ht]
\caption{\textbf{Prompt for Gemini Deep Research to generate ideas to integrate single-cell batch effects.}}
\phantomsection
\addcontentsline{toc}{subsection}{\protect\numberline{Table S15:}Prompt for Gemini Deep Research to generate ideas to integrate single-cell batch effects.}
\label{table:sc_deep_research_prompt}
    \centering
    \begin{tabular}{l}
    \begin{tcolorbox}[title={Prompt for Gemini Deep Research.}]
I am developing new methods for winning single-cell batch integration competitions, as proposed by the Kaggle and extensively researched in the single-cell genomics community.\\

Briefly:
Modelers are asked to develop a function, \verb|eliminate_batch_effect_fn|, that transforms raw gene expression count data from multiple batches into a low-dimensional embedding or feature matrix. This transformed output should effectively remove technical variation (batch effects) while rigorously preserving biological information (e.g., cell type identity). The performance of these methods is evaluated against a suite of metrics that quantify both batch mixing and biological conservation.\\

The key problem is to develop a method that takes an \verb|AnnData| object of raw gene expression counts with batch labels and returns an AnnData object with a batch-integrated low-dimensional embedding in the \verb|.obsm['X_emb']| field. The method must excel across a diverse set of evaluation metrics, including ASW Batch, ASW Label, ARI, NMI, Graph Connectivity, Isolated Labels ASW, Isolated Labels F1, kBET, iLISI, cLISI, PCR, and Cell Cycle Conservation Score, aiming to maximize their average.\\

The following principles should be obeyed when choosing models:

* **Batch Effect Removal**: Prioritize techniques that explicitly model and mitigate batch-specific variations without collapsing biological signal.

* **Biological Conservation**: Ensure the integrated representation retains and accurately reflects genuine biological differences, particularly cell type distinctions, as measured by clustering and silhouette metrics.

* **Scalability and Efficiency**: Given the large dataset sizes (e.g., $329,762$ cells $\times$ $2,000$ genes), models must be computationally efficient and avoid out-of-memory errors.

* **Constraint Adherence**: The implementation must strictly avoid using cell\_type information during integration and should primarily leverage \verb|scanpy|, \verb|sklearn|, \verb|numpy|, \verb|scipy|, \verb|tensorflow|, \verb|torch|, \verb|jax|, or equivalent native implementations rather than specialized single-cell packages.\\

This task aims to develop a SUPERHUMAN METHOD for solving this problem.\\

Please give me 10 highly novel and creative ideas with detailed implementation notes for the set of methods I should explore for solving this task. I aim to create the best method for solving this problem, preferably creating the best ever method.
\end{tcolorbox}
    \end{tabular}
\end{table}

\clearpage
\newpage
\begin{table}[ht]
\caption{\textbf{Prompt for formatting Deep Research ideas into a structure similar to baseline method descriptions.}}
\phantomsection
\addcontentsline{toc}{subsection}{\protect\numberline{Table S16:}Prompt for formatting Deep Research ideas into a structure similar to baseline method descriptions.}
\label{table:sc_deep_research_format}
    \centering
    \begin{tabular}{l}
    \begin{tcolorbox}[title={Prompt for formatting Deep Research ideas.}]
    Structure the given idea into the following format:\\
    
    \textless{}description\textgreater{}\\
    Your description about the method goes here.\\
    \textless{}/description\textgreater{}\\
    \\
    \textless{}steps\textgreater{}\\
    Your list of steps to implement the method goes here.\\
    \textless{}/steps\textgreater{}\\
    \\
    \textless{}notes\textgreater{}\\
    Strengths and weaknesses of the idea goes here.\\
    \textless{}/notes\textgreater{}



\end{tcolorbox}
    \end{tabular}
\end{table}

\clearpage
\newpage
\begin{table}[ht]
\caption{\textbf{Prompt for guiding ERA to generate hybrid strategies.}}
\phantomsection
\addcontentsline{toc}{subsection}{\protect\numberline{Table S17:}Prompt for guiding ERA to generate hybrid strategies.}
\label{table:sc_recombine_ts_prompt}
    \centering
    \begin{tabular}{l}
    \begin{tcolorbox}[title={Prompt for guiding ERA to generate hybrid strategies.}]
We have up until now done experiments with two major types of codes, that are described in detail below. PLEASE CREATE AN ALGORITHM THAT USES THE BEST PARTS OF BOTH STRATEGIES TO CREATE A HYBRID STRATEGY THAT IS TRULY WONDERFUL AND SCORES HIGHER THAN EITHER OF THE INDIVIDUAL STRATEGIES.\\

[Output from \suptab{table:sc_recombine_summary_prompt}.]
\end{tcolorbox}
    \end{tabular}
\end{table}

\clearpage
\begin{table}[ht]
\caption{\textbf{Prompt for summarizing existing batch integration methods.} The prompt instructs the LLM to write a brief description and steps of the method. This generated summary then serves as part of an explicit instruction for tree search to create an optimized version of the existing batch integration method. Example output is shown in \suptab{table:sc_baseline_prompt_output_bbknn}.}
\phantomsection
\addcontentsline{toc}{subsection}{\protect\numberline{Table S18:}Prompt for summarizing existing batch integration methods.}
\label{table:sc_baseline_prompt}
    \centering
    \begin{tabular}{l}
    \begin{tcolorbox}[title={Prompt for summarizing differences between two baseline methods.}]
Given the following paper, please identify the main method being proposed. Then write a very short method description. This method description will be used to reproduce the method. DO NOT mention the algorithm by name.\\

Your output must follow this format:

\textless description\textgreater

Your very short description goes here.

\textless/description\textgreater \\

\textless steps\textgreater

Your short list of steps goes here.

\textless/steps\textgreater

\end{tcolorbox}
    \end{tabular}
\end{table}

\clearpage
\newpage
\begin{table}[ht]
\caption{\textbf{Example of existing method description (BBKNN) given to \ourmethod.} \ourmethod used this short method description to further optimize the method.}
\phantomsection
\addcontentsline{toc}{subsection}{\protect\numberline{Table S19:}Example of existing method description (BBKNN) given to \ourmethod.}
\label{table:sc_baseline_prompt_output_bbknn}
    \centering
    \begin{tabular}{l}
    \begin{tcolorbox}[title={Example of existing method description (BBKNN) given to \ourmethod.}]
    
\textless description\textgreater

This method performs batch correction by modifying the neighborhood graph construction step. For each cell, its nearest neighbors are identified independently within each batch, rather than across the entire combined dataset. The resulting batch-specific neighbor lists for each cell are then merged to create a single, integrated graph. This approach assumes that shared cell types exist across batches and that biological differences are greater than technical batch effects.

\textless/description\textgreater \\

\textless steps\textgreater

\begin{enumerate}
    \item For each cell, iterate through every batch in the dataset.
    \item Find the k-nearest neighbors for the cell from within the current batch, based on a given distance metric (e.g., Euclidean distance in PCA space).
    \item After iterating through all batches, merge the identified neighbor sets for the cell into a single neighborhood.
    \item Repeat for all cells to construct a batch-corrected neighborhood graph.
\end{enumerate}

\textless/steps\textgreater

\end{tcolorbox}
    \end{tabular}
\end{table}

\clearpage
\newpage

\end{appendices}

\end{document}